%% file: main.tex
\documentclass[11pt]{article}
\usepackage[preprint]{acl}
\usepackage{times,latexsym}
\usepackage[T1]{fontenc}
\usepackage[utf8]{inputenc}
\usepackage{microtype,inconsolata}
\usepackage{booktabs,amsmath,amssymb,tabularx,placeins,graphicx}
\title{Can We Believe AI Peer Review?}
\author{Zongrui Yang \quad Haoyuan Li \quad Zhongsheng Wang \\
  Gillian Dobbie \quad Jiamou Liu \\
  University of Auckland \\
  Auckland, New Zealand}
\begin{document}
\maketitle

\begin{abstract}

As large language models become widely used in scientific review and manuscript revision, AI scores are increasingly serving as feedback for assessing paper quality and guiding edits. Yet whether a single AI score reliably reflects the reviewer's judgment, and whether revision gains based on that score are endorsed by other reviewers, remain understudied. Using 200 papers from ICLR 2025 and 2026 and approximately 14,000 AI reviews, we study score reliability, score-distribution estimation, and cross-model agreement on revision gains. Repeated reviews fluctuate even under identical conditions. Finer-grained scoring protocols reduce score variance but also systematically shift mean scores, while the gains assigned to manuscript revisions depend strongly on the reviewer.

\end{abstract}
\section{Introduction}

As large language models become integrated into research workflows, AI-assisted peer review is moving from experimental studies into practical use. Submission volumes at major AI conferences have continued to grow, increasing the pressure on peer review. Conferences including ICLR and AAAI have begun exploring AI assistance in the review process \citep{zou2025leveraging,aaai2025launches}. Meanwhile, researchers are using AI review tools to assess paper quality, obtain revision suggestions, and judge the effects of revisions through changes in scores. Both applications rest on a shared premise: AI judgments are reliable enough to provide meaningful evidence for research decisions.

\begin{figure}[t]
    \centering
    \includegraphics[width=\columnwidth]{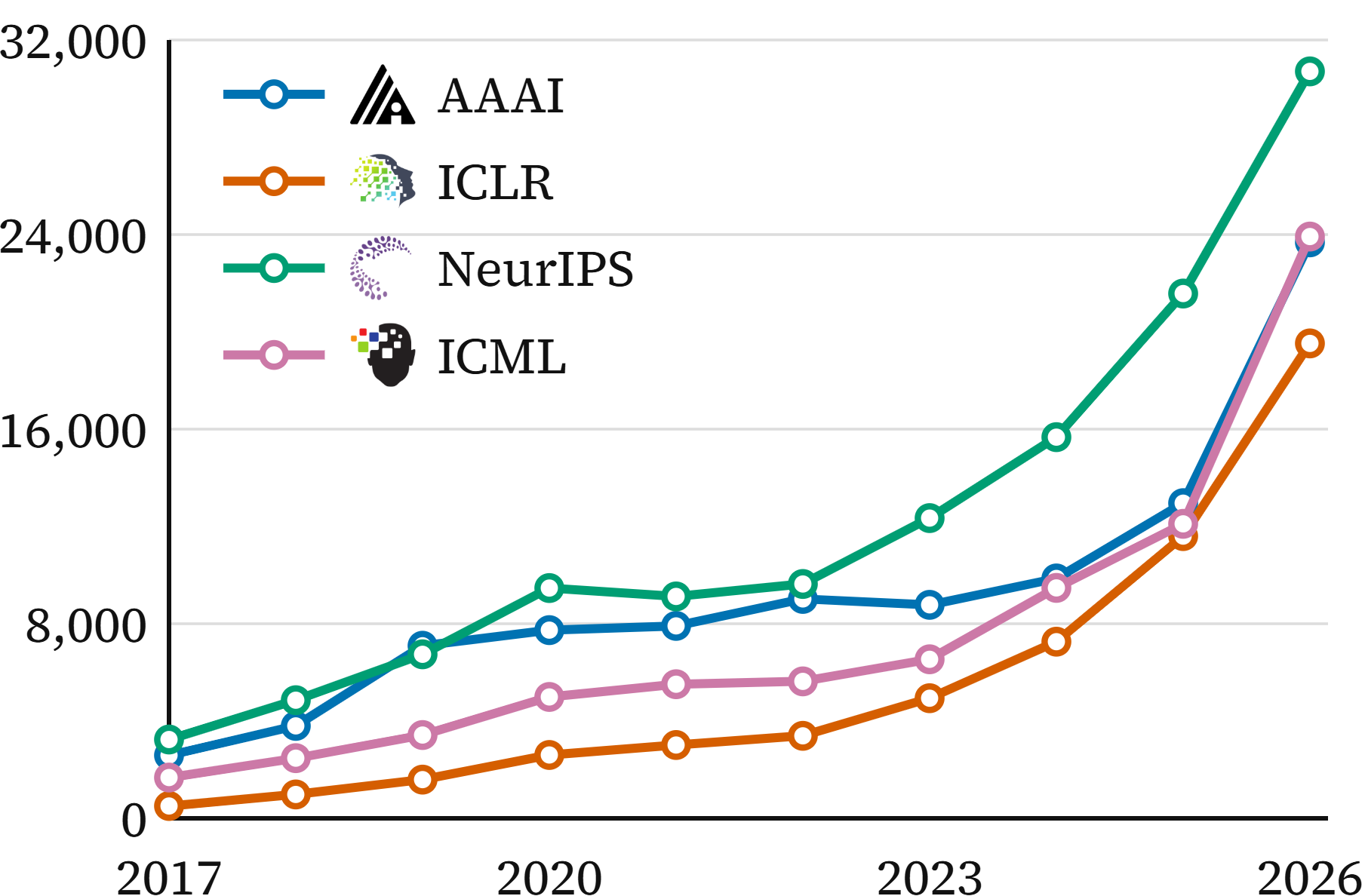}
    \caption{Number of submissions to major AI conferences over time.}
    \label{fig:conference_submissions}
\end{figure}

This premise does not necessarily hold. First, AI scores are stochastic: even when the paper, model, and review protocol remain fixed, repeated reviews can yield different scores, making an individual score or an observed gain an unreliable representation of the model's own assessment. Second, AI judgments depend on the model: reviewers may have different evaluation preferences and, even when they assess original papers similarly, may not endorse the same revisions. Ignoring these issues can lead users to mistake random fluctuations for differences in paper quality and score gains under a particular model for broadly endorsed improvements. How much, then, can we trust AI scores and the gains observed after revision?

Prior work has examined AI reviewers' evaluation capabilities, scoring biases, and sensitivity to paper presentation. ReviewEval~\citep{garg-etal-2025-revieweval}, TitleTrap~\citep{du-2025-titletrap}, and LLM-as-a-Reviewer~\citep{li2026llmreviewer} study different aspects of AI reviewing, while Rhetoric~\citep{li2026rhetoric} further reveals differences in reviewers' responses to revision gains. Practical use of AI scores nevertheless calls for a connected set of answers: How reliable is a single score? How many reviews are needed to estimate a reviewer's expected score accurately? Can score estimation be improved under a limited reasoning budget? Does an observed gain reflect random fluctuation, or an improvement endorsed by a particular reviewer? These questions determine when AI scores can serve as research feedback.

To address these questions, we conduct approximately 14,000 AI reviews of 200 papers from ICLR 2025 and ICLR 2026, systematically studying scoring randomness, estimation accuracy, and cross-model agreement on revision gains. Our main contributions are:

\textbf{Quantifying the reliability of a single AI score.} We use repeated sampling to analyze score fluctuations and their effects on relative paper rankings, then derive sufficient review budgets for specified mean-score error and success-probability targets.

\textbf{Exploring methods to improve scoring reliability.} We examine how score granularity affects variability and systematic shifts, and test whether auxiliary score probabilities improve distribution estimation under a limited reasoning budget.

\textbf{Assessing the credibility of revision gains.} Through repeated scoring and cross-model evaluation, we separate scoring fluctuations from reviewer dependence and test whether different reviewers consistently endorse the same revisions through score gains.

\section{Related Work}

\textbf{AI peer review.} Existing studies primarily assess LLM reviewing ability through the quality of review comments. ReviewEval~\citep{garg-etal-2025-revieweval} examines factuality, analytical depth, and the actionability of suggestions; LLM-as-a-Reviewer~\citep{li2026llmreviewer} compares model and human reviews and tests resistance to prompt injection. These studies help establish whether LLMs can generate high-quality reviews, but devote less attention to how review outcomes inform subsequent decisions. In particular, the extent of score variation when the same paper is reviewed repeatedly, and whether these scores reliably distinguish paper quality before and after revision, remain understudied.

\textbf{Score stability and estimation.} Prior work finds inconsistencies across repeated LLM scoring runs and systematic biases \citep{stureborg2024inconsistent}; output formats and explanation requirements also affect evaluation results \citep{chiang-lee-2023-closer}. To address scoring uncertainty, G-Eval~\citep{liu-etal-2023-g} uses score-token probabilities, while \citet{homan-etal-2026-many} examine how many repeated ratings statistical tests require. However, existing work rarely directly answers how much a single score differs from a mean over repeated scores when the reviewer and reviewing conditions are fixed. We target the expected score and score distribution under fixed reviewing conditions, analyze how repeated sampling affects estimation precision, and evaluate the gains from probability-assisted estimation against independently sampled target scores.

\textbf{AI-assisted revision.} Beyond evaluating papers, LLM scores can be used to select and guide revisions. TitleTrap~\citep{du-2025-titletrap} and Rhetoric~\citep{li2026rhetoric} show that changes to titles and rhetorical presentation can affect AI scores. A score increase may therefore reflect either an improvement in substantive paper quality or a selector's adaptation to particular scoring preferences. Existing work has not sufficiently separated these sources. We hold the candidate edit set fixed and vary the selector and scorer separately, comparing score gains under shared revision content. This tests whether LLM-score-guided revision strategies remain stable across reviewers and how much of the observed gains can be attributed to sampling uncertainty.
\begin{figure}[t]
    \centering
    \includegraphics[width=\columnwidth]{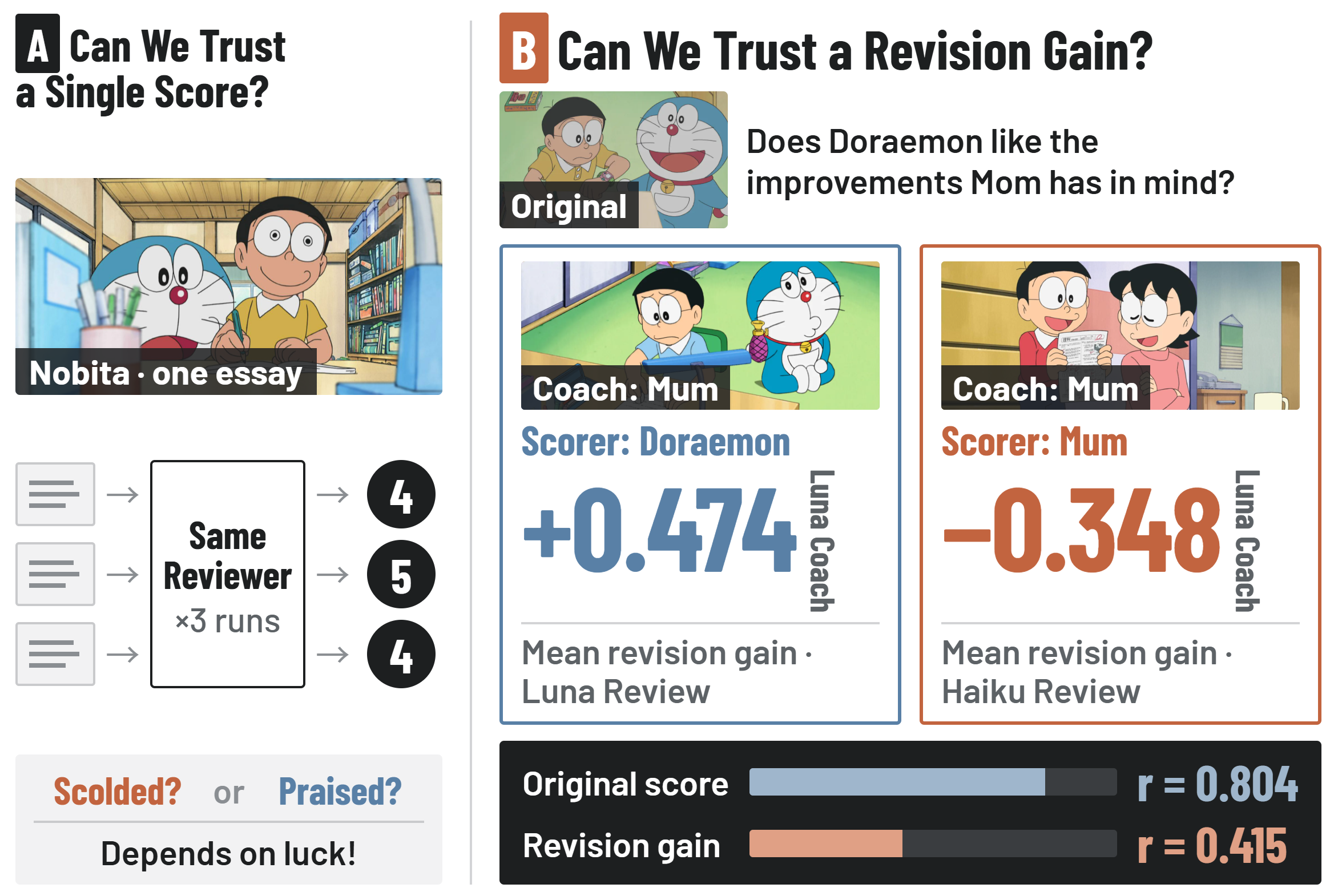}
    \caption{Can we believe AI Peer Review?}
    \label{fig:motivation}
\end{figure}

\section{Problem Formulation and Study Design}

\subsection{What Does a Review Score Estimate?}

\textbf{Score distributions and repeated sampling.} We treat one LLM paper-review procedure as one sample. Given a paper $P_i$ and review configuration $r$ (model, prompt, reasoning settings, and execution environment), an individual score is a random variable $Y_r(P_i)$ with expectation $\mu_{r,i}$ and variance $\sigma_{r,i}^2$. For the same paper, we perform $k$ independent reviews and use their mean $\bar Y_{r,i,k}$ to estimate the expected score under the fixed configuration. Under IID scores within each paper, this mean has variance $\sigma_{r,i}^2/k$.

\textbf{Evaluation validity.} We use the mean human review score $H_i$ as an external reference for paper quality. Correlation and error relative to this reference, together with AUC for acceptance status, serve as indicators of evaluation validity.

\textbf{Scoring reliability.} For a fixed paper and review configuration, scores should exhibit low randomness, and their expectation should be reliably estimable from a finite number of repeated reviews. We report both score stability and relative-ranking stability as indicators of scoring reliability.

\textbf{Evaluation validity for revisions.} For a revision $P_i\rightarrow P_i'$, the reviewer's expected gain is $\Delta_{r,i}=\mu_r(P_i')-\mu_r(P_i)$, estimated by the difference between independent mean scores assigned by the same Scorer to the original and revised manuscripts. Gains from different Scorers are compared on identical revised versions. For fixed original and revised manuscripts, independent scores on the two sides give the gain estimate a variance of $(\sigma_{r,i}^{2}+\sigma_{r,i}^{\prime2})/k$. An observed score increase contains sampling error from both versions. In the crossover experiment, we score each version repeatedly and keep the Scorer of the original-paper baseline consistent, establishing separate controls for within-model fluctuations and between-model response differences.

\subsection{Papers and Review Configurations}

We draw experimental samples from public ICLR 2025 and ICLR 2026 submissions by stratifying jointly on official subject area and mean human score interval, selecting 100 papers per year. The paper lists and configurations are provided in the supplementary material.

For LLM paper review, PDFs are converted into ordered text and images, retaining the main text, figures, tables, and appendices; review inputs are anonymized. A common task asks models to assess novelty, technical correctness, experimental evidence, and limitations, then produce an overall score from 1 to 10 with review comments.

Original integer scoring uses GPT-5.6-Sol, GPT-6-Luna, and GPT-6-Sol, with ten reviews per paper per configuration, totaling 6,000. The granularity experiment compares integer and forced-noninteger protocols for the latter two models. Distribution estimation targets GPT-5.6-Sol reasoning scores. Revision experiments use Luna and Claude Haiku 5.5, with three scores each for original and revised manuscripts; one paper has 224 images, exceeding Claude's input limit, so crossover comparisons use the shared 199 papers. Additional configuration details are provided in the supplementary material.

\section{How Reliable Is a Single AI Review?}

We first characterize score variation and validity for individual papers, then test whether independent reviews reproduce the same paper rankings. We further estimate ranking reliability under finite sampling and recommend review budgets.

\subsection{Repeated Scores and Human References}

Table~\ref{tab:repeatability} summarizes ten-score repeatability and external associations. Across configurations, 30.0\%--56.5\% of papers receive ten identical scores, while 91.5\%--100\% have a score range of at most one point. Fluctuations mainly occur between adjacent categories. GPT-6-Sol has a within-paper SD of 0.272, compared with 0.374 and 0.386 for the other configurations, and a Spearman correlation of 0.409 with human scores. Final-acceptance AUC is 0.7134 for GPT-5.6-Sol, 0.6484 for GPT-6-Luna, and 0.7138 for GPT-6-Sol.

\begin{table*}[t]\centering
\begin{tabularx}{\textwidth}{@{}>{\hsize=1.6970\hsize\raggedright\arraybackslash}X>{\hsize=0.6894\hsize\raggedright\arraybackslash}X>{\hsize=1.1136\hsize\raggedright\arraybackslash}X>{\hsize=1.0076\hsize\raggedright\arraybackslash}X>{\hsize=0.6894\hsize\raggedright\arraybackslash}X>{\hsize=1.1136\hsize\raggedright\arraybackslash}X>{\hsize=0.6894\hsize\raggedright\arraybackslash}X@{}}\toprule
Model & Mean & Identical 10/10 & Range \(\leq\) 1 & SD & Spearman \(\rho\) & AUC\\\midrule
GPT-5.6-Sol & 4.2755 & 30.0\% & 95.0\% & 0.374 & 0.378 & 0.713\\
GPT-6-Luna & 3.4730 & 32.5\% & 91.5\% & 0.386 & 0.346 & 0.648\\
GPT-6-Sol & 4.6635 & 56.5\% & 100.0\% & 0.272 & 0.409 & 0.714\\
\bottomrule\end{tabularx}
\caption{Repeated integer scores and external references. Each configuration covers 200 papers with ten reviews per paper. SD is the square root of average within-paper sample variance. Correlations, AUC, and related metrics use ten-score means.}
\label{tab:repeatability}
\end{table*}

Within-paper SD averages unbiased paper-level variances before taking the square root. Each set of ten scores forms 45 unordered pairs; exact agreement is 72.86\%, 72.36\%, and 85.19\%, respectively. GPT-6-Sol stays within a one-point range for all 200 papers, including 113 with ten identical scores.

\subsection{Do Independent Reviews Reproduce Rankings?}

We form disjoint equal-sized blocks from each paper's ten scores, comparing means of one through five reviews. Enumerating all unordered block partitions gives 45, 630, 2,100, 1,575, and 126 partitions, respectively. Each configuration covers 9,900 within-year paper pairs, averaged over all partitions. Rankings are classified as the same strict order, a tie in at least one block, or reversed strict orders.

\begin{table}[t]
\centering
\small
\begin{tabular}{@{}lcccc@{}}
\toprule
Model & Reviews & Same & Tie & Reverse \\
\midrule
GPT-5.6-Sol & 1 & 62.14 & 36.71 & 1.15 \\
            & 2 & 73.52 & 24.74 & 1.74 \\
            & 3 & 78.92 & 19.08 & 2.00 \\
            & 4 & 82.18 & 15.71 & 2.10 \\
            & 5 & 84.47 & 13.38 & 2.15 \\
\midrule
GPT-6-Luna  & 1 & 63.00 & 35.60 & 1.40 \\
            & 2 & 74.91 & 23.05 & 2.04 \\
            & 3 & 80.40 & 17.30 & 2.30 \\
            & 4 & 83.58 & 14.03 & 2.39 \\
            & 5 & 85.71 & 11.85 & 2.44 \\
\midrule
GPT-6-Sol   & 1 & 57.52 & 42.14 & 0.34 \\
            & 2 & 66.43 & 33.03 & 0.54 \\
            & 3 & 70.24 & 29.11 & 0.65 \\
            & 4 & 72.56 & 26.71 & 0.73 \\
            & 5 & 74.13 & 25.08 & 0.79 \\
\bottomrule
\end{tabular}
\caption{Agreement between disjoint review blocks (\%).
All ten observations participate in the partitions;
Same denotes identical strict rankings,
Tie a tie in at least one block,
and Reverse opposite strict rankings.
Values may not sum to 100 due to rounding.}
\label{tab:ranking}
\end{table}

More reviews mainly improve strict ranking reproducibility by reducing ties. From one to five reviews per block, strict agreement rises from 62.14\% to 84.47\% for GPT-5.6-Sol and from 63.00\% to 85.71\% for Luna. GPT-6-Sol rises from 57.52\% to 74.13\%, with 25.08\% of pairs still tied in at least one five-score block. Its reversal rate is 0.79\%, below 2.15\% and 2.44\% for the other configurations. Finer means establish more strict comparisons, including reversals of pairs previously tied. We next specify a sampling target through absolute mean-score estimation error.

\subsection{How Many Reviews Are Enough?}

Weighting the sampled papers equally, we define the overall success rate of $k$ new reviews as

\[A(k,\epsilon)=\frac{1}{N}\sum_{i=1}^{N}\Pr\!\left(|\bar Y_{i,k}-\mu_i|<\epsilon\right).\]

We assume IID scores within each paper, independence across papers, and future scores independent of the pilot. We require $A(k,\epsilon)\ge0.95$. This means that, for a randomly selected paper, the mean of repeated reviews falls within $\epsilon$ points of the model's expected score with probability at least 95\%. Let $V=N^{-1}\sum_i\sigma_i^2$. Under independent and identically distributed reviews, Chebyshev's inequality gives

\[
A(k,\epsilon) \ge 1-\frac{V}{k\epsilon^2},
\qquad
k \ge \left\lceil \frac{V}{0.05\epsilon^2} \right\rceil.
\]

This budget depends only on the average variance and requires neither normally distributed scores nor a known score distribution. Substituting the average sample variance $\hat V$ from ten pilot scores gives planning budgets of 45, 48, and 24 reviews at a 0.25-point tolerance, and 12, 12, and 6 reviews at a 0.5-point tolerance.

\textbf{Confidence control with finite pilot data.} We also use the known score range $[1,10]$ to construct an upper confidence bound $U$ for $V$. The sample variance of ten scores for each paper equals the average half-squared difference over all score pairs. Treating it as a U-statistic over unordered score pairs, we obtain a one-sided 95\% upper bound depending only on $\hat V$ through random disjoint pairing and a KL-Chernoff bound \citep{hoeffding1963probability}. Each model has an effective pairing count of $200\lfloor10/2\rfloor=1000$; the 9,000 score pairs that share observations are not treated as independent. Specifically, let $B=(10-1)^2/2=40.5$ and $M=1000$, and solve for the root above $\hat V$:

\[M\operatorname{kl}(\hat V/B,U/B)=\log20,\]

where $\operatorname{kl}(p,q)=p\log(p/q)+(1-p)\log((1-p)/(1-q))$. This inversion satisfies $\Pr(V>U)\le0.05$. The key step expresses all overlapping score pairs as the conditional expectation of random disjoint matchings, then uses Jensen's inequality to obtain the moment-generating-function bound for 1000 independent bounded kernels. Substituting $U$ into the same budget formula gives a sufficient budget for an overall success rate of 95\% on future new reviews, with 95\% pilot confidence for each model.

\begin{table*}[t]\centering
\begin{tabularx}{\textwidth}{@{}>{\hsize=1.2069\hsize\raggedright\arraybackslash}X>{\hsize=0.6034\hsize\raggedright\arraybackslash}X>{\hsize=0.6034\hsize\raggedright\arraybackslash}X>{\hsize=1.2931\hsize\raggedright\arraybackslash}X>{\hsize=1.2931\hsize\raggedright\arraybackslash}X@{}}\toprule
Model & $\hat V$ & $U$ & Planning .25 / .5 & Certified .25 / .5\\\midrule
GPT-5.6-Sol & 0.1401 & 0.4114 & 45 / 12 & 132 / 33\\
GPT-6-Luna & 0.1489 & 0.4257 & 48 / 12 & 137 / 35\\
GPT-6-Sol & 0.0741 & 0.2978 & 24 / 6 & 96 / 24\\
\bottomrule\end{tabularx}
\caption{Mean-score precision budgets without empirical score distributions. Planning uses $\hat V$; Certified uses the one-sided 95\% upper bound $U$ for $V$. The paired budgets correspond to error tolerances of 0.25 and 0.5 points. Budgets count new reviews and exclude the ten pilot scores; Certified confidence is computed separately for each model.}
\label{tab:precision}
\end{table*}

The two budgets correspond to different information requirements: planning budgets reflect the scale of observed variance, while confidence-controlled budgets also cover score tails that have not been sufficiently observed. Halving the tolerance approximately quadruples both budgets. In practice, the budget can be selected according to the error tolerance and confidence requirement. The bound's construction, proof, and simultaneous-control results for all three models appear in the supplementary material.

Figure~\ref{fig:precision_budget} plots the budgets for the three configurations at a tolerance of 0.5 points as lower bounds on the success rate. Dashed curves substitute observed variances, and solid curves use their upper confidence bounds, corresponding to the planning scale and confidence-controlled sufficient budgets, respectively. The curves are calculated from the variance relationship without resampling the observed frequencies of the ten scores.

\begin{figure}[t]\centering
\includegraphics[width=\columnwidth]{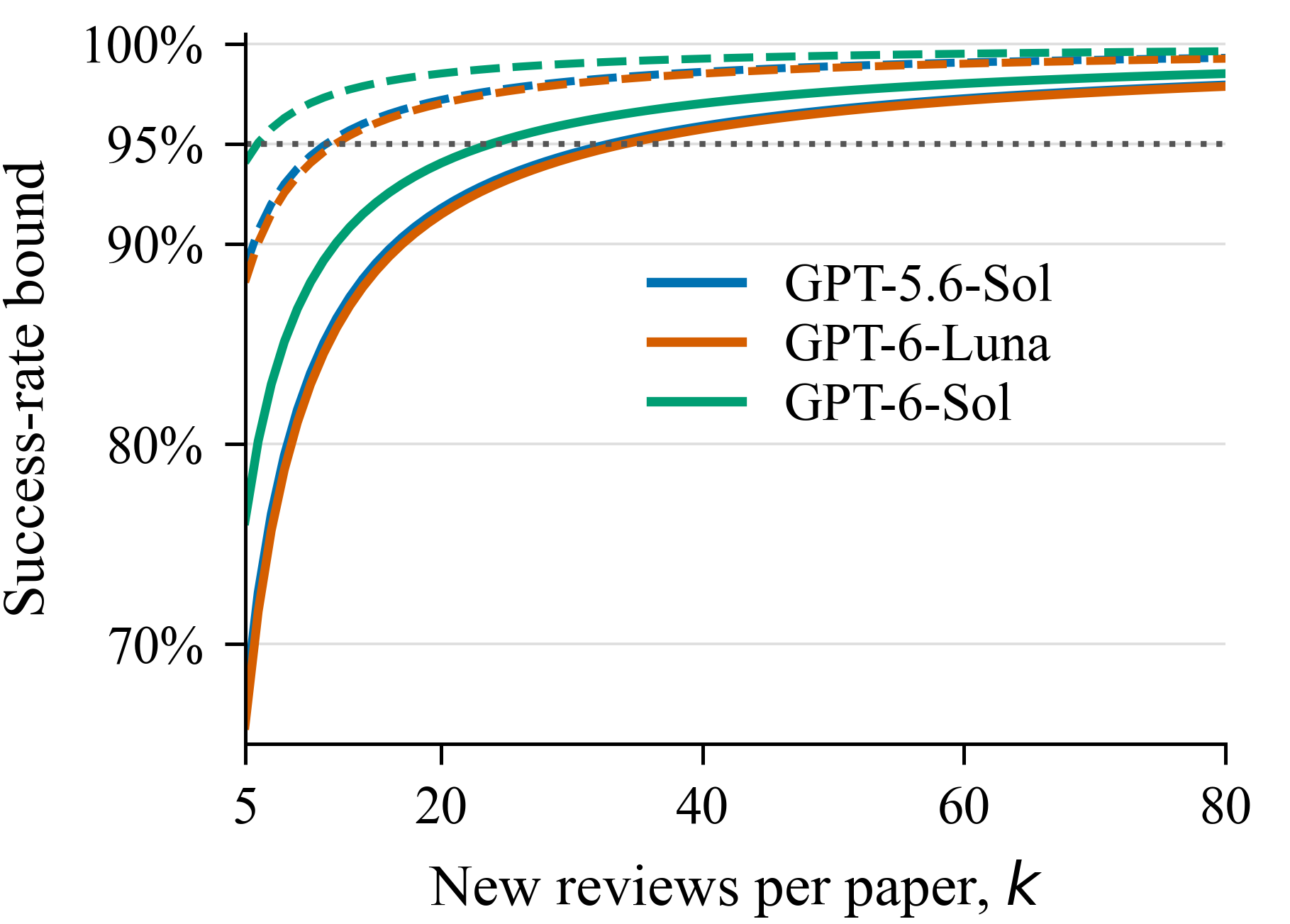}
\caption{Variance-based precision bounds at a half-point tolerance. Dashed curves plug in observed variance for planning; solid curves use the one-sided 95\% variance upper bounds. The dotted horizontal line marks 95\% success.}
\label{fig:precision_budget}
\end{figure}

\section{Can We Improve Scoring Reliability?}

Repeated reviews improve the reliability of scores and paper rankings, but reaching reliable judgments about closely scored papers may require many independent calls. Can AI-score reliability be improved without substantially increasing review costs?

\subsection{Finer Scores Change Both Variance and Location}

If high-quality peer review is viewed as a ranking of paper quality, that ranking need not be inherently discrete. Scores, however, are discrete integers. This motivates testing whether finer output granularity improves model stability. We fix paper inputs and all other review requirements, replacing only the integer protocol with a 1--10 score required to have a nonzero fractional part. Luna and Sol each produce five repeated scores for all 200 papers.

\begin{table*}[t]\centering
\begin{tabularx}{\textwidth}{@{}>{\hsize=1.0000\hsize\raggedright\arraybackslash}X>{\hsize=0.9259\hsize\raggedright\arraybackslash}X>{\hsize=0.6296\hsize\raggedright\arraybackslash}X>{\hsize=1.2222\hsize\raggedright\arraybackslash}X>{\hsize=1.2222\hsize\raggedright\arraybackslash}X@{}}\toprule
Model & Integer / Decimal SD & SD change & Mean shift [95\% CI] & MAE change [95\% CI]\\\midrule
GPT-6-Luna & 0.390 / 0.329 & -15.6\% & +0.693 [0.649, 0.739] & -0.307 [-0.389, -0.223]\\
GPT-6-Sol & 0.286 / 0.249 & -13.2\% & +0.266 [0.233, 0.298] & -0.001 [-0.050, 0.044]\\
\bottomrule\end{tabularx}
\caption{Integer versus forced-decimal scoring. Shift and error changes are Decimal-Integer; brackets give paired paper-bootstrap 95\% intervals. MAE uses public mean human review scores as the reference.}
\label{tab:decimal}
\end{table*}

Score variability decreases for both models, while their means rise by 0.693 and 0.266, respectively. Luna's MAE relative to mean human scores also decreases, whereas Sol's MAE remains largely unchanged. The decimal protocol thus changes both location and variability on the scoring scale. To estimate the original integer reviewer, we next retain its protocol and use additional information to improve estimation.

\subsection{Estimating the Same Reviewer with Fewer Reasoning Calls}

Beyond comparing integer and decimal scoring, we evaluate two simple strategies for estimating the integer score distribution $p_i$ of a fixed GPT-5.6-Sol reasoning configuration with fewer reasoning calls. \textbf{Quick} extracts token probabilities for candidate integer scores from one non-reasoning call, normalizes them, adjusts their concentration using temperature $T$, and calibrates score locations through an affine mapping $s\mapsto a+bs$. Mapped locations are clipped to $[1,10]$, and probability mass at non-integer positions is allocated linearly to the two adjacent score categories, yielding $q_Q$. \textbf{Standard} combines this estimate with the empirical distribution of two target reasoning scores:
$q_S=w\hat p_2+(1-w)q_Q$.

We use the 198 papers with available auxiliary probabilities and evaluate both strategies over 30 random paper-level splits, each containing 99 training and 99 test papers. Quick uses the first valid non-reasoning output for each paper. Its parameters $T,a,b$ are selected by minimizing Brier loss against the training papers' reasoning scores, over a grid of six temperatures, 13 offsets, and three slopes. With these parameters fixed, Standard selects $w$ from eight candidate weights by enumerating all ten ways to divide each training paper's five reasoning scores into two inputs and three independent targets. Evaluation on test papers follows the same input--target separation, with all parameters fixed from the training set.

\textbf{Risk evaluation with independent targets.} For a predicted distribution $q$ and independent target scores $Z_1,\ldots,Z_m$, we compute

\[
\begin{aligned}
\widehat R(q)
&= \|q\|_2^2
  - \frac{2}{m}\sum_j q(Z_j)\\
&\quad
  + \frac{1}{m(m-1)}
    \sum_{j\ne l}\mathbf{1}[Z_j=Z_l].
\end{aligned}
\]

Conditional on $q$, its expectation is $\|q-p_i\|_2^2$. The collision rate between target scores estimates $\|p_i\|_2^2$, so variability from the finite target sample is not directly added to prediction risk. For $n$ direct reasoning samples, risk is $(1-\|p_i\|_2^2)/n$, also estimated using the collision rate of independent scores.

\begin{table*}[t]\centering
\begin{tabularx}{\textwidth}{@{}>{\hsize=1.0000\hsize\raggedright\arraybackslash}X>{\hsize=0.9259\hsize\raggedright\arraybackslash}X>{\hsize=0.6296\hsize\raggedright\arraybackslash}X>{\hsize=1.2222\hsize\raggedright\arraybackslash}X>{\hsize=1.2222\hsize\raggedright\arraybackslash}X@{}}\toprule
Method & R / Q calls & Risk & Risk 95\% CI & Held-out AUC\\\midrule
Quick & 0 / 1 & 0.2427 & [0.2029, 0.2860] & 0.7092\\
Standard & 2 / 1 & 0.0829 & [0.0750, 0.0911] & 0.7194\\
1 Reasoning & 1 / 0 & 0.2434 & [0.2051, 0.2828] & 0.6895\\
2 Reasoning & 2 / 0 & 0.1217 & [0.1025, 0.1414] & 0.7063\\
3 Reasoning & 3 / 0 & 0.0811 & [0.0684, 0.0943] & 0.7124\\
5 Reasoning & 5 / 0 & 0.0487 & [0.0410, 0.0566] & 0.7162\\
\bottomrule\end{tabularx}
\caption{Held-out squared distribution risk. R/Q denote full reasoning reviews and non-reasoning probability calls, respectively. All risks use the shared 198 papers and the same held-out aggregation procedure; 95\% intervals bootstrap papers conditional on the 30 fitted splits. Direct-sampling risk is calculated using the $1/n$ relationship.}
\label{tab:distribution-risk}
\end{table*}

Standard reduces risk by 0.0388 relative to two direct reasoning samples, with a paired 95\% interval of [-0.0524, -0.0255]; its risk difference from three reasoning samples is +0.0018 [-0.0057, 0.0095]. Auxiliary probabilities provide useful information alongside two reasoning samples, reaching a risk level close to three direct reasoning samples. Quick's risk is close to that of a single reasoning sample, and fusion is more effective than replacing the scorer alone. The table reports the call composition separately; actual dollar costs depend on input length and caching.

The effect of fusion on risk can also be expressed directly. If the auxiliary call and the two reasoning scores are conditionally independent, and the empirical reasoning distribution is unbiased for $p_i$, then $\mathbb E\|q_S-p_i\|_2^2=w^2(1-\|p_i\|_2^2)/2+(1-w)^2\mathbb E\|q_Q-p_i\|_2^2$. The reasoning term captures direct-sampling error, while the auxiliary term captures the discrepancy between the calibrated distribution and the target. Selecting the weight on training papers allocates risk between the two terms; debiased test-set evaluation assesses whether that choice remains effective on papers excluded from fitting.

\subsection{Independent-Target Validation}

A separate existing experiment collects ten additional independent reasoning scores for five papers from each year, using subsets of the original five scores for prediction. Quick and Standard are fitted on the other 188 papers. Standard obtains a Brier score of 0.353 and Wasserstein-1 distance of 0.19; three direct reasoning samples obtain 0.379 and 0.18, respectively. The independent targets give a similar estimation pattern to the main experiment. These losses include finite-target variability and are reported separately from the debiased risk in Table~\ref{tab:distribution-risk}; Table~\ref{tab:independent-target} retains all five baselines. This ten-paper evaluation uses new target scores and separates evaluation papers from parameter training.

\begin{table}[t]\centering
\begin{tabularx}{\columnwidth}{@{}Xrr@{}}\toprule
Method & Brier & W1\\\midrule
Quick & 0.497 & 0.36\\
Standard & 0.353 & 0.19\\
1 Reasoning & 0.552 & 0.28\\
3 Reasoning & 0.379 & 0.18\\
5 Reasoning & 0.344 & 0.14\\
\bottomrule\end{tabularx}
\caption{Existing validation with independent target reviews. Five papers per year, each with ten new target scores; all methods use the same ten papers. Quick/Standard are fitted on the other 188 papers, and reasoning baselines enumerate subsets of the original five scores. Brier is the squared loss against individual one-hot targets; W1 is distance to the target empirical CDF, in score points.}
\label{tab:independent-target}
\end{table}

\section{Can We Trust Revision Gains?}

Accurately estimating one reviewer's expected score does not establish whether other reviewers endorse its judgments. This matters particularly for AI-assisted manuscript revision: when a revised paper receives a higher score, can we trust that increase to reflect an effective improvement rather than a particular reviewer's preferences? Through controlled manuscript revision and cross-model reviewing, we study whether different reviewers respond consistently to the same edits and how much of the observed disagreement can be explained by scoring randomness.

\subsection{Holding Revisions Fixed}

With more precise score estimation, we examine another use case: AI scores as revision feedback. The workflow has three roles: the Writer generates candidate text edits from the original manuscript and existing reviews; the Selector decides whether to accept each edit; and the Scorer evaluates the complete manuscript after edits are applied. The Scorer does not receive selection rationales. Revisions are required to preserve facts, experimental data, and non-text materials.

Candidate edits generated by Luna are selected by Luna or Haiku and then scored by both, forming a complete $2\times2$ crossover. Haiku-generated edits add three role combinations, for seven configurations in total. Each Scorer independently scores the original and revised manuscripts three times each, and all comparisons use the same 199 papers. Of Luna's 9,327 candidate edits, 8,459 are accepted by both Selectors, with a mean per-paper acceptance-set Jaccard similarity of 0.919; the final versions still differ for 182 papers. This lets us compare selection effects with fixed candidate edits and scoring effects with fixed final versions.

\subsection{Opposite Gains for the Same Revisions}

\begin{table*}[t]\centering
\begin{tabularx}{\textwidth}{@{}>{\hsize=0.7934\hsize\raggedright\arraybackslash}X>{\hsize=0.8926\hsize\raggedright\arraybackslash}X>{\hsize=0.7934\hsize\raggedright\arraybackslash}X>{\hsize=0.8926\hsize\raggedright\arraybackslash}X>{\hsize=1.2893\hsize\raggedright\arraybackslash}X>{\hsize=1.3388\hsize\raggedright\arraybackslash}X@{}}\toprule
Writer & Selector & Scorer & Mean gain & 95\% CI & + / 0 / -\\\midrule
Luna & Luna & Luna & +0.474 & [0.367, 0.578] & 120 / 54 / 25\\
Luna & Haiku & Luna & +0.472 & [0.362, 0.580] & 129 / 41 / 29\\
Luna & Luna & Haiku & -0.348 & [-0.432, -0.265] & 32 / 63 / 104\\
Luna & Haiku & Haiku & -0.296 & [-0.379, -0.214] & 31 / 63 / 105\\
Haiku & Luna & Luna & +0.302 & [0.238, 0.367] & 108 / 70 / 21\\
Haiku & Haiku & Luna & +0.281 & [0.216, 0.347] & 96 / 84 / 19\\
Haiku & Haiku & Haiku & +0.050 & [0.000, 0.104] & 57 / 102 / 40\\
\bottomrule\end{tabularx}
\caption{Revision gains across writer, selector, and scorer. Each configuration includes 199 papers, with three scores each for the original and revised versions. The 95\% intervals use 10,000 paper-level paired bootstrap replicates. The final column counts papers with positive/zero/negative differences between three-score means.}
\label{tab:revision-gains}
\end{table*}

\begin{figure}[t]\centering
\includegraphics[width=\columnwidth]{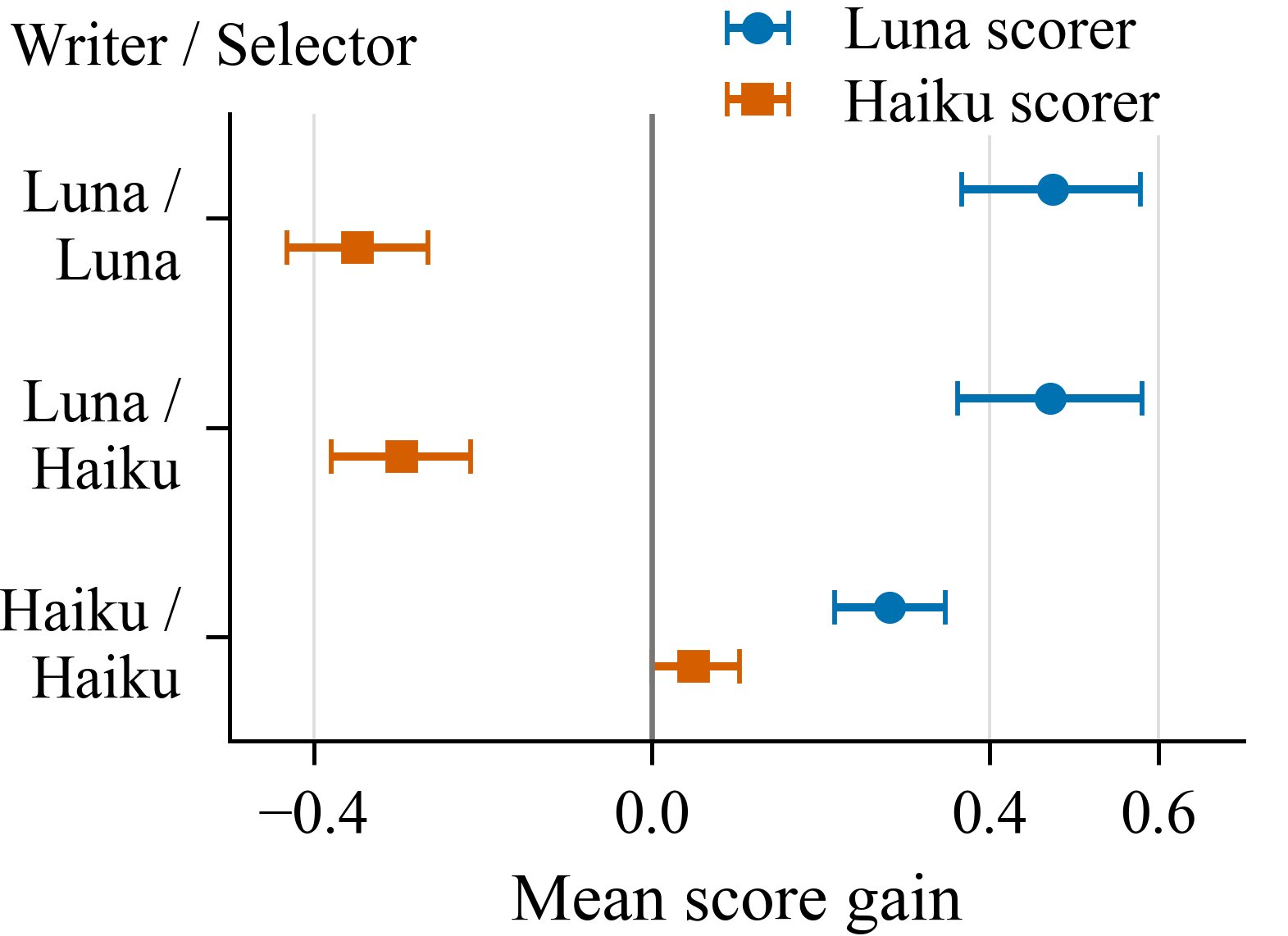}
\caption{Mean revision gains for identical manuscript versions under different scorers. Bars show paper-level bootstrap 95\% intervals.}
\label{fig:revision_gains}
\end{figure}

For identical revisions generated and selected by Luna, Luna assigns a mean gain of +0.474 and Haiku -0.348, a difference of 0.822 [0.717, 0.928]. Replacing the Selector with Haiku still yields +0.472 and -0.296, respectively; the cross-scorer gap changes by only -0.054 [-0.116, 0.010]. Figure~\ref{fig:revision_gains} shows the separation in scorer responses under the same selection condition.

Luna also assigns positive average gains to Haiku-generated revisions, while Haiku assigns its own selected revisions a gain of +0.050. The result is asymmetric: Luna endorses revisions from both Writers; Haiku assigns negative gains to Luna's revisions and near-zero gains to its own. Changing the Selector has a smaller effect on the aggregate results for Luna's candidate edits. Across these two sets of Luna revisions, the final Scorer changes the direction of the gain.

These controls serve different purposes. Holding the final manuscript fixed and changing the Scorer directly measures differing responses to the same textual edits; holding the Writer's candidate edits fixed and changing the Selector tests whether selection changes this gap. The two Selectors accept many of the same edits. Even when final versions differ, Luna and Haiku assign opposite aggregate gains under both selection conditions. Cross-reviewer disagreement can thus be studied without comparing two revisions with different content.

\subsection{Agreement on Originals Does Not Predict Agreement on Gains}

\begin{table}[t]\centering
\begin{tabularx}{\columnwidth}{@{}>{\hsize=1.03\hsize\raggedright\arraybackslash}X>{\hsize=.55\hsize\raggedright\arraybackslash}X>{\hsize=1.42\hsize\raggedright\arraybackslash}X@{}}\toprule
Target & Pearson & Adjusted Pearson [95\% CI]\\\midrule
Original scores & 0.804 & 0.843 [0.783, 0.890]\\
Luna / Luna & 0.415 & 0.521 [0.336, 0.686]\\
Luna / Haiku & 0.407 & 0.518 [0.337, 0.673]\\
Haiku / Haiku & 0.085 & ---\\
\bottomrule\end{tabularx}
\caption{Cross-scorer correlation before and after revision. Adjusted subtracts the scoring-noise variance of three-score means from the observed between-paper variance; intervals bootstrap papers. This diagnostic assumes independent, zero-mean scoring errors across models and manuscript versions. Gain rows: Writer/Selector.}
\label{tab:gain-correlations}
\end{table}

The two Scorers have a Pearson correlation of 0.804 on original-paper means, but their paper-level gain correlations for identical revisions fall to 0.415, 0.407, and 0.085. Gain directions agree exactly for 28.1\%, 30.2\%, and 41.2\% of papers, respectively, including cases in which both assign zero gain.

We estimate the noise contribution from repeated scores: the noise variance of original-paper means is $s_o^2/3$, and that of gains is $(s_o^2+s_v^2)/3$. We subtract these components from the corresponding between-paper variances while retaining the cross-model covariance. Adjusted correlations for Luna revisions are approximately 0.52, still below 0.843 for original papers. Scoring fluctuations explain part of the correlation decline, while opposite aggregate gains for fixed revised versions directly demonstrate differing reviewer responses.

\section{Conclusion}

Using AI scores as research feedback involves two inferences: from individual outputs to expected judgments, and from those judgments to revision gains. Repeated reviews improve mean-score precision, with budgets determined by average variance and error tolerance and further controlled for uncertainty in finite pilot data. Under a fixed scoring protocol, auxiliary probabilities combined with a small number of reasoning scores improve distribution estimation; changing output granularity alters both variability and the scoring center. The revision crossover further shows that even highly correlated original-paper judgments can lead to gains in opposite directions for identical revisions. Score precision clarifies one reviewer's judgment, while cross-reviewer evaluation of shared revisions tests whether this feedback receives consistent endorsement.

\FloatBarrier\clearpage\section*{Limitations}

Sampling budgets target new independent scores under a fixed configuration; changes to model versions, prompts, or service environments change the target. Confidence bounds use the full 1--10 range and are conservative when tail observations are limited. Human scores come from current public records, and PDFs may be updated versions; the two years use different rating categories, and correlations do not directly observe latent quality. Calibration is validated on one API configuration, while the revision crossover includes only two models and omits one role combination. Factual preservation in revisions is required by the editing protocol but has not been validated through blinded human assessment of compliance and quality gains; using existing reviews to generate edits may also affect the independence assumption in noise adjustment. Experiments use public anonymized manuscripts. The findings analyze scoring behavior; automated editing and scoring do not replace authors' responsibility for paper content.

\bibliography{references}
\clearpage
\input{appendix}
\end{document}

%% file: appendix.tex
\appendix

\section{Data and Sampling}

\subsection{Submission Populations and Sample Composition}

The study includes 100 papers each from ICLR 2025 and ICLR 2026. The sampling populations contain 11,520 and 19,474 submissions with publicly available official reviews, respectively, including accepted, rejected, withdrawn, and desk-rejected papers. The paper lists were fixed before downloading PDFs or running AI reviews. AI scores, PDF availability, and overlap with earlier samples played no role in selection.

The 2025 sample was stratified jointly by official primary subject area and mean human-score bin. The 2026 sample uses the second version of the sampling design, which adds submission status to these dimensions.

\begin{table*}[tp]\centering
\normalsize
\setlength{\tabcolsep}{4pt}
\renewcommand{\arraystretch}{1.12}
\renewcommand{\thetable}{S1}
\begin{tabular}{@{}>{\raggedright\arraybackslash}p{248.451pt}>{\raggedleft\arraybackslash}p{95.397pt}>{\raggedleft\arraybackslash}p{95.397pt}@{}}
\toprule
{\textbf{\hspace{0pt}Item}} & {\textbf{\hspace{0pt}ICLR 2025}} & {\textbf{\hspace{0pt}ICLR 2026}} \\
\midrule
{\hspace{0pt}Population with public official reviews} & {\hspace{0pt}11,520} & {\hspace{0pt}19,474} \\
{\hspace{0pt}Study sample} & {\hspace{0pt}100} & {\hspace{0pt}100} \\
{\hspace{0pt}Official primary subject areas} & {\hspace{0pt}21} & {\hspace{0pt}21} \\
{\hspace{0pt}Population mean human score} & {\hspace{0pt}5.1496} & {\hspace{0pt}4.2027} \\
{\hspace{0pt}Sample mean human score} & {\hspace{0pt}5.1983} & {\hspace{0pt}4.2117} \\
{\hspace{0pt}Population acceptance rate} & {\hspace{0pt}32.14\%} & {\hspace{0pt}27.48\%} \\
{\hspace{0pt}Sample acceptance rate} & {\hspace{0pt}36.00\%} & {\hspace{0pt}27.00\%} \\
{\hspace{0pt}Maximum area-share deviation (pp)} & {\hspace{0pt}0.4635} & {\hspace{0pt}0.5573} \\
{\hspace{0pt}Maximum score-bin share deviation (pp)} & {\hspace{0pt}0.9219} & {\hspace{0pt}0.9131} \\
{\hspace{0pt}Acceptance status constrained in sampling} & {\hspace{0pt}No} & {\hspace{0pt}Yes} \\
\bottomrule
\end{tabular}
\caption{Sampling and representativeness in the two years.}
\end{table*}

\subsection{Stratified Quotas and Within-Stratum Sampling}

For 2025, integer quotas for three nested sample sizes of 100, 300, and 1,000 were jointly solved using SciPy \texttt{milp} and HiGHS. For each subject area, score bin, and nonempty area-by-score-bin stratum, the sample count was constrained to the floor or ceiling of its population share multiplied by the sample size, keeping the count discrepancy strictly below one paper. Additional constraints ensured coverage of subject areas and nonempty score bins. Within each stratum, papers were sorted by their exact mean human score, with random ordering of ties, and selected by systematic sampling with a random start. The sample of 1,000 was drawn first, followed by 300 from that sample and then 100 from the 300.

The second 2026 design uses area-by-score-bin-by-status strata. Status categories are Poster, Oral, rejected, withdrawn, and desk rejected. Status quotas were assigned by the largest-remainder method, with the total number of accepted papers fixed at 27. Area, score-bin, and cross-stratum quotas retained the floor/\allowbreak{}ceiling constraints, and within-stratum selection again used systematic sampling after sorting by exact mean score. Section A.3 reports the final status quotas and counts.

The human mean for each paper is the unweighted average of its valid official review scores. Bins are assigned using the exact, unrounded mean. Nonempty bins range from \texttt{[1,2)} to \texttt{[9,10]} in 2025. The 2026 population includes means below 1, with nonempty bins from \texttt{[0,1)} to \texttt{[8,9)}; its \texttt{[9,10]} bin is empty. The complete distributions follow.

\begin{table*}[tp]\centering
\small
\setlength{\tabcolsep}{4pt}
\renewcommand{\arraystretch}{1.12}
\renewcommand{\thetable}{S2}
\begin{tabular}{@{}>{\raggedleft\arraybackslash}p{58.478pt}>{\raggedleft\arraybackslash}p{65.178pt}>{\raggedleft\arraybackslash}p{49.048pt}>{\raggedleft\arraybackslash}p{60.158pt}>{\raggedleft\arraybackslash}p{65.178pt}>{\raggedleft\arraybackslash}p{49.048pt}>{\raggedleft\arraybackslash}p{60.158pt}@{}}
\toprule
{\textbf{\hspace{0pt}Category}} & {\textbf{\hspace{0pt}2025 population}} & {\textbf{\hspace{0pt}2025 sample}} & {\textbf{\hspace{0pt}Deviation (pp)}} & {\textbf{\hspace{0pt}2026 population}} & {\textbf{\hspace{0pt}2026 sample}} & {\textbf{\hspace{0pt}Deviation (pp)}} \\
\midrule
{\hspace{0pt}[0, 1)} & {\hspace{0pt}0} & {\hspace{0pt}0} & {\hspace{0pt}+0.000} & {\hspace{0pt}54} & {\hspace{0pt}1} & {\hspace{0pt}+0.723} \\
{\hspace{0pt}[1, 2)} & {\hspace{0pt}25} & {\hspace{0pt}1} & {\hspace{0pt}+0.783} & {\hspace{0pt}236} & {\hspace{0pt}1} & {\hspace{0pt}-0.212} \\
{\hspace{0pt}[2, 3)} & {\hspace{0pt}243} & {\hspace{0pt}2} & {\hspace{0pt}-0.109} & {\hspace{0pt}1946} & {\hspace{0pt}10} & {\hspace{0pt}+0.007} \\
{\hspace{0pt}[3, 4)} & {\hspace{0pt}1643} & {\hspace{0pt}14} & {\hspace{0pt}-0.262} & {\hspace{0pt}4458} & {\hspace{0pt}23} & {\hspace{0pt}+0.108} \\
{\hspace{0pt}[4, 5)} & {\hspace{0pt}2928} & {\hspace{0pt}25} & {\hspace{0pt}-0.417} & {\hspace{0pt}6927} & {\hspace{0pt}35} & {\hspace{0pt}-0.571} \\
{\hspace{0pt}[5, 6)} & {\hspace{0pt}3142} & {\hspace{0pt}27} & {\hspace{0pt}-0.274} & {\hspace{0pt}4063} & {\hspace{0pt}21} & {\hspace{0pt}+0.136} \\
{\hspace{0pt}[6, 7)} & {\hspace{0pt}2585} & {\hspace{0pt}22} & {\hspace{0pt}-0.439} & {\hspace{0pt}1541} & {\hspace{0pt}7} & {\hspace{0pt}-0.913} \\
{\hspace{0pt}[7, 8)} & {\hspace{0pt}817} & {\hspace{0pt}7} & {\hspace{0pt}-0.092} & {\hspace{0pt}228} & {\hspace{0pt}1} & {\hspace{0pt}-0.171} \\
{\hspace{0pt}[8, 9)} & {\hspace{0pt}128} & {\hspace{0pt}1} & {\hspace{0pt}-0.111} & {\hspace{0pt}21} & {\hspace{0pt}1} & {\hspace{0pt}+0.892} \\
{\hspace{0pt}[9, 10]} & {\hspace{0pt}9} & {\hspace{0pt}1} & {\hspace{0pt}+0.922} & {\hspace{0pt}0} & {\hspace{0pt}0} & {\hspace{0pt}+0.000} \\
\bottomrule
\end{tabular}
\caption{Population and sample distributions of mean human-score bins. Deviations are sample shares minus population shares, in percentage points; empty bins are shown as zero.}
\end{table*}

\begin{figure*}[tp]\centering
\includegraphics[width=\textwidth]{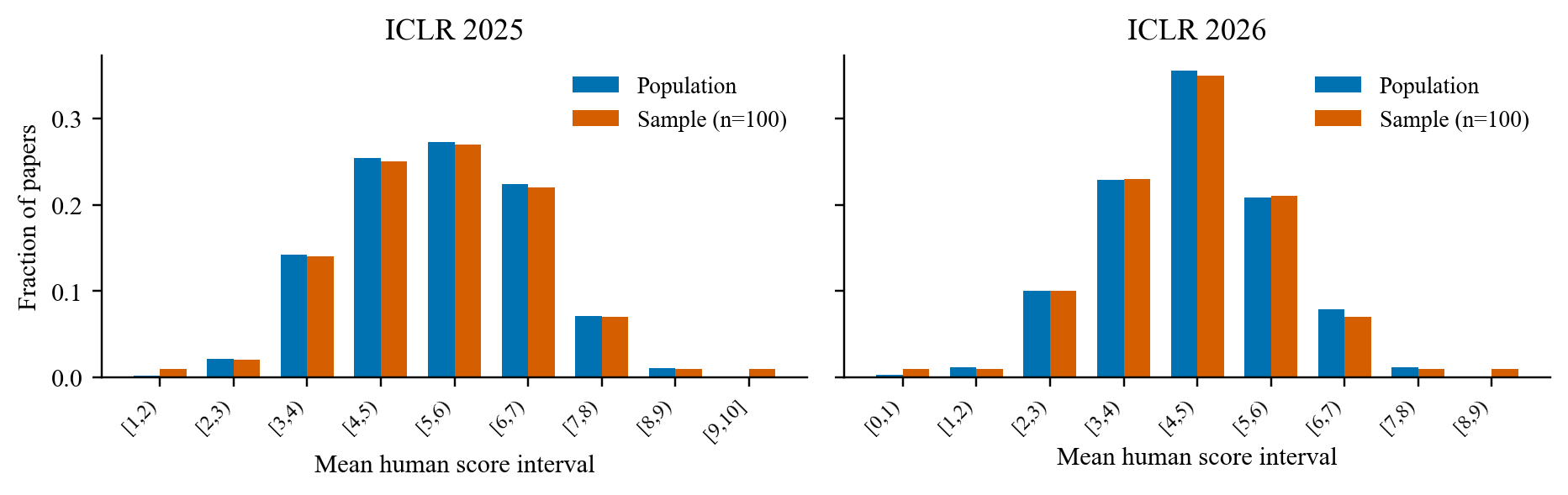}
\renewcommand{\thefigure}{A1}
\caption{Score-bin shares in each year's population and sample of 100 papers. The plotted bins follow the nonempty range for each year.}
\end{figure*}

\subsection{Subject-Area and Submission-Status Distributions}

All 21 official primary subject areas are represented. The table retains the original area names without merging small areas.

\begin{table*}[tp]\centering
\small
\setlength{\tabcolsep}{4pt}
\renewcommand{\arraystretch}{1.12}
\renewcommand{\thetable}{S3}
\begin{tabular}{@{}>{\raggedright\arraybackslash}p{210.432pt}>{\raggedleft\arraybackslash}p{22.432pt}>{\raggedleft\arraybackslash}p{32.432pt}>{\raggedleft\arraybackslash}p{43.542pt}>{\raggedleft\arraybackslash}p{22.432pt}>{\raggedleft\arraybackslash}p{32.432pt}>{\raggedleft\arraybackslash}p{43.542pt}@{}}
\toprule
{\textbf{\hspace{0pt}Category}} & {\textbf{\hspace{0pt}2025 pop.}} & {\textbf{\hspace{0pt}2025 sample}} & {\textbf{\hspace{0pt}Deviation (pp)}} & {\textbf{\hspace{0pt}2026 pop.}} & {\textbf{\hspace{0pt}2026 sample}} & {\textbf{\hspace{0pt}Deviation (pp)}} \\
\midrule
{\hspace{0pt}alignment, fairness, safety, privacy, and societal considerations} & {\hspace{0pt}1039} & {\hspace{0pt}9} & {\hspace{0pt}-0.019} & {\hspace{0pt}1667} & {\hspace{0pt}9} & {\hspace{0pt}+0.440} \\
{\hspace{0pt}applications to computer vision, audio, language, and other modalities} & {\hspace{0pt}1359} & {\hspace{0pt}12} & {\hspace{0pt}+0.203} & {\hspace{0pt}2687} & {\hspace{0pt}14} & {\hspace{0pt}+0.202} \\
{\hspace{0pt}applications to neuroscience \& cognitive science} & {\hspace{0pt}214} & {\hspace{0pt}2} & {\hspace{0pt}+0.142} & {\hspace{0pt}372} & {\hspace{0pt}2} & {\hspace{0pt}+0.090} \\
{\hspace{0pt}applications to physical sciences (physics, chemistry, biology, etc.)} & {\hspace{0pt}540} & {\hspace{0pt}5} & {\hspace{0pt}+0.313} & {\hspace{0pt}846} & {\hspace{0pt}4} & {\hspace{0pt}-0.344} \\
{\hspace{0pt}applications to robotics, autonomy, planning} & {\hspace{0pt}242} & {\hspace{0pt}2} & {\hspace{0pt}-0.101} & {\hspace{0pt}589} & {\hspace{0pt}3} & {\hspace{0pt}-0.025} \\
{\hspace{0pt}causal reasoning} & {\hspace{0pt}116} & {\hspace{0pt}1} & {\hspace{0pt}-0.007} & {\hspace{0pt}189} & {\hspace{0pt}1} & {\hspace{0pt}+0.029} \\
{\hspace{0pt}datasets and benchmarks} & {\hspace{0pt}753} & {\hspace{0pt}7} & {\hspace{0pt}+0.464} & {\hspace{0pt}1551} & {\hspace{0pt}8} & {\hspace{0pt}+0.036} \\
{\hspace{0pt}foundation or frontier models, including LLMs} & {\hspace{0pt}1281} & {\hspace{0pt}11} & {\hspace{0pt}-0.120} & {\hspace{0pt}2623} & {\hspace{0pt}13} & {\hspace{0pt}-0.469} \\
{\hspace{0pt}generative models} & {\hspace{0pt}1045} & {\hspace{0pt}9} & {\hspace{0pt}-0.071} & {\hspace{0pt}1646} & {\hspace{0pt}9} & {\hspace{0pt}+0.548} \\
{\hspace{0pt}infrastructure, software libraries, hardware, systems, etc.} & {\hspace{0pt}86} & {\hspace{0pt}1} & {\hspace{0pt}+0.253} & {\hspace{0pt}176} & {\hspace{0pt}1} & {\hspace{0pt}+0.096} \\
{\hspace{0pt}interpretability and explainable AI} & {\hspace{0pt}505} & {\hspace{0pt}4} & {\hspace{0pt}-0.384} & {\hspace{0pt}776} & {\hspace{0pt}4} & {\hspace{0pt}+0.015} \\
\bottomrule
\end{tabular}
\caption{Distributions across the 21 primary subject areas, using the official labels for each year.}
\end{table*}
\begin{table*}[tp]\centering
\small
\setlength{\tabcolsep}{4pt}
\renewcommand{\arraystretch}{1.12}
\renewcommand{\thetable}{S3}
\begin{tabular}{@{}>{\raggedright\arraybackslash}p{210.432pt}>{\raggedleft\arraybackslash}p{22.432pt}>{\raggedleft\arraybackslash}p{32.432pt}>{\raggedleft\arraybackslash}p{43.542pt}>{\raggedleft\arraybackslash}p{22.432pt}>{\raggedleft\arraybackslash}p{32.432pt}>{\raggedleft\arraybackslash}p{43.542pt}@{}}
\toprule
{\textbf{\hspace{0pt}Category}} & {\textbf{\hspace{0pt}2025 pop.}} & {\textbf{\hspace{0pt}2025 sample}} & {\textbf{\hspace{0pt}Deviation (pp)}} & {\textbf{\hspace{0pt}2026 pop.}} & {\textbf{\hspace{0pt}2026 sample}} & {\textbf{\hspace{0pt}Deviation (pp)}} \\
\midrule
{\hspace{0pt}learning on graphs and other geometries \& topologies} & {\hspace{0pt}428} & {\hspace{0pt}4} & {\hspace{0pt}+0.285} & {\hspace{0pt}528} & {\hspace{0pt}3} & {\hspace{0pt}+0.289} \\
{\hspace{0pt}learning on time series and dynamical systems} & {\hspace{0pt}265} & {\hspace{0pt}2} & {\hspace{0pt}-0.300} & {\hspace{0pt}457} & {\hspace{0pt}2} & {\hspace{0pt}-0.347} \\
{\hspace{0pt}learning theory} & {\hspace{0pt}358} & {\hspace{0pt}3} & {\hspace{0pt}-0.108} & {\hspace{0pt}498} & {\hspace{0pt}2} & {\hspace{0pt}-0.557} \\
{\hspace{0pt}neurosymbolic \& hybrid AI systems (physics-informed, logic \& formal reasoning, etc.)} & {\hspace{0pt}105} & {\hspace{0pt}1} & {\hspace{0pt}+0.089} & {\hspace{0pt}222} & {\hspace{0pt}1} & {\hspace{0pt}-0.140} \\
{\hspace{0pt}optimization} & {\hspace{0pt}509} & {\hspace{0pt}4} & {\hspace{0pt}-0.418} & {\hspace{0pt}725} & {\hspace{0pt}4} & {\hspace{0pt}+0.277} \\
{\hspace{0pt}other topics in machine learning (i.e., none of the above)} & {\hspace{0pt}551} & {\hspace{0pt}5} & {\hspace{0pt}+0.217} & {\hspace{0pt}795} & {\hspace{0pt}4} & {\hspace{0pt}-0.082} \\
{\hspace{0pt}probabilistic methods (Bayesian methods, variational inference, sampling, UQ, etc.)} & {\hspace{0pt}235} & {\hspace{0pt}2} & {\hspace{0pt}-0.040} & {\hspace{0pt}371} & {\hspace{0pt}2} & {\hspace{0pt}+0.095} \\
{\hspace{0pt}reinforcement learning} & {\hspace{0pt}666} & {\hspace{0pt}6} & {\hspace{0pt}+0.219} & {\hspace{0pt}1050} & {\hspace{0pt}5} & {\hspace{0pt}-0.392} \\
{\hspace{0pt}transfer learning, meta learning, and lifelong learning} & {\hspace{0pt}370} & {\hspace{0pt}3} & {\hspace{0pt}-0.212} & {\hspace{0pt}560} & {\hspace{0pt}3} & {\hspace{0pt}+0.124} \\
{\hspace{0pt}unsupervised, self-supervised, semi-supervised, and supervised representation learning} & {\hspace{0pt}853} & {\hspace{0pt}7} & {\hspace{0pt}-0.405} & {\hspace{0pt}1146} & {\hspace{0pt}6} & {\hspace{0pt}+0.115} \\
\bottomrule
\end{tabular}
\caption{Distributions across the 21 primary subject areas, using the official labels for each year. (continued)}
\end{table*}

\begin{table*}[tp]\centering
\small
\setlength{\tabcolsep}{4pt}
\renewcommand{\arraystretch}{1.12}
\renewcommand{\thetable}{S4}
\begin{tabular}{@{}>{\raggedright\arraybackslash}p{88.958pt}>{\raggedleft\arraybackslash}p{60.098pt}>{\raggedleft\arraybackslash}p{43.968pt}>{\raggedleft\arraybackslash}p{55.078pt}>{\raggedleft\arraybackslash}p{60.098pt}>{\raggedleft\arraybackslash}p{43.968pt}>{\raggedleft\arraybackslash}p{55.078pt}@{}}
\toprule
{\textbf{\hspace{0pt}Category}} & {\textbf{\hspace{0pt}2025 population}} & {\textbf{\hspace{0pt}2025 sample}} & {\textbf{\hspace{0pt}Deviation (pp)}} & {\textbf{\hspace{0pt}2026 population}} & {\textbf{\hspace{0pt}2026 sample}} & {\textbf{\hspace{0pt}Deviation (pp)}} \\
\midrule
{\hspace{0pt}accepted\_oral} & {\hspace{0pt}213} & {\hspace{0pt}3} & {\hspace{0pt}+1.151} & {\hspace{0pt}224} & {\hspace{0pt}1} & {\hspace{0pt}-0.150} \\
{\hspace{0pt}accepted\_poster} & {\hspace{0pt}3110} & {\hspace{0pt}30} & {\hspace{0pt}+3.003} & {\hspace{0pt}5127} & {\hspace{0pt}26} & {\hspace{0pt}-0.327} \\
{\hspace{0pt}accepted\_spotlight} & {\hspace{0pt}380} & {\hspace{0pt}3} & {\hspace{0pt}-0.299} & {\hspace{0pt}0} & {\hspace{0pt}0} & {\hspace{0pt}+0.000} \\
{\hspace{0pt}desk\_rejected} & {\hspace{0pt}8} & {\hspace{0pt}0} & {\hspace{0pt}-0.069} & {\hspace{0pt}685} & {\hspace{0pt}4} & {\hspace{0pt}+0.482} \\
{\hspace{0pt}rejected} & {\hspace{0pt}4908} & {\hspace{0pt}41} & {\hspace{0pt}-1.604} & {\hspace{0pt}8352} & {\hspace{0pt}43} & {\hspace{0pt}+0.112} \\
{\hspace{0pt}withdrawn} & {\hspace{0pt}2901} & {\hspace{0pt}23} & {\hspace{0pt}-2.182} & {\hspace{0pt}5086} & {\hspace{0pt}26} & {\hspace{0pt}-0.117} \\
\bottomrule
\end{tabular}
\caption{Submission-status distributions. The 2025 categories include Spotlight, which is absent from the 2026 status classification used here.}
\end{table*}

The 2025 sample contains 36 accepted, 41 rejected, and 23 withdrawn papers. The 2026 sample contains 27 accepted, 43 rejected, 26 withdrawn, and 4 desk-rejected papers. Together, these comprise 63 accepted, 84 rejected, 49 withdrawn, and 4 desk-rejected papers.

Quota rounding allows rare categories to enter a small sample. For example, only nine papers in the 2025 population have a mean human score of at least 9, and one is retained in the sample of 100: a sample share of 1\% versus a population share of approximately 0.078\%. Coverage of subject areas and score bins does not imply coverage of every small cross-stratum; some receive a rounded quota of zero. Results give equal weight to the selected papers, without reweighting to population shares.

\subsection{Human Scores, PDF Versions, and Acceptance Labels}

Human scores are taken from the public records in the frozen data and may include post-rebuttal updates. Paper PDFs may also be updated versions. Human score options are 1, 3, 5, 6, 8, and 10 for 2025, and 0, 2, 4, 6, 8, and 10 for 2026. Section C therefore also reports associations and errors separately by year.

Acceptance labels are derived from final public status: \texttt{accepted\_\allowbreak{}*} is labeled 1, and rejected, withdrawn, and desk-rejected papers are labeled 0. The primary analysis uses all 200 papers, with 63 positives and 137 negatives. Excluding withdrawals leaves 151 papers: 63 positives and 88 negatives. The full and withdrawal-excluded samples contain 100 and 77 papers in 2025, and 100 and 74 in 2026. Positive-negative pairs with equal predicted scores receive a contribution of 0.5 to AUC.

\subsection{Shared Samples and Paper Lists Across Experiments}

\begin{table*}[tp]\centering
\normalsize
\setlength{\tabcolsep}{4pt}
\renewcommand{\arraystretch}{1.12}
\renewcommand{\thetable}{S5}
\begin{tabular}{@{}>{\raggedright\arraybackslash}p{134.264pt}>{\raggedleft\arraybackslash}p{35.616pt}>{\raggedright\arraybackslash}p{125.182pt}>{\raggedright\arraybackslash}p{136.182pt}@{}}
\toprule
{\textbf{\hspace{0pt}Experiment}} & {\textbf{\hspace{0pt}Shared papers}} & {\textbf{\hspace{0pt}Scores or inputs per paper}} & {\textbf{\hspace{0pt}Sample details}} \\
\midrule
{\hspace{0pt}Repeated integer scores on originals} & {\hspace{0pt}200} & {\hspace{0pt}10 per model, three models} & {\hspace{0pt}Historical and additional runs combined: 6,000 scores} \\
{\hspace{0pt}Integer/\allowbreak{}forced-decimal comparison} & {\hspace{0pt}200} & {\hspace{0pt}Luna and Sol: 5 per protocol} & {\hspace{0pt}Integer controls: first five under same configuration; 1,000 decimal scores each} \\
{\hspace{0pt}Probability-assisted distribution estimation} & {\hspace{0pt}198} & {\hspace{0pt}5 target reasoning scores plus auxiliary probabilities} & {\hspace{0pt}Missing probabilities: \texttt{2025-\allowbreak{}paper\_\allowbreak{}056} and \texttt{2026-\allowbreak{}paper\_\allowbreak{}053}} \\
{\hspace{0pt}Independent-target validation} & {\hspace{0pt}10} & {\hspace{0pt}5 existing scores for prediction; 10 additional target scores} & {\hspace{0pt}5 per year; parameters fitted on the other 188} \\
{\hspace{0pt}Revision crossover} & {\hspace{0pt}199} & {\hspace{0pt}3 each for original and each revision} & {\hspace{0pt}Claude excludes \texttt{2026-\allowbreak{}paper\_\allowbreak{}067}, which has 224 images} \\
\bottomrule
\end{tabular}
\caption{Shared samples and scoring inputs across experiments.}
\end{table*}

The two papers missing auxiliary probabilities differ from the paper affected by Claude's image limit. The full calibration sample contains 198 papers, with 63 accepted and 135 nonaccepted. Excluding its 48 withdrawn papers leaves 150 papers, with 63 accepted and 87 nonaccepted. The revision experiment uses a fixed shared set of 199 papers: 100 from 2025 and 99 from 2026.

Scoring models receive anonymized manuscript content. Human scores, acceptance status, and other models' results remain in the analysis metadata.

\section{Review and Revision Protocols}

\subsection{Models and Review Settings}

\begin{table*}[tp]\centering
\normalsize
\setlength{\tabcolsep}{4pt}
\renewcommand{\arraystretch}{1.12}
\renewcommand{\thetable}{S6}
\begin{tabular}{@{}>{\raggedright\arraybackslash}p{142.871pt}>{\raggedright\arraybackslash}p{89.776pt}>{\raggedright\arraybackslash}p{50.895pt}>{\raggedright\arraybackslash}p{147.701pt}@{}}
\toprule
{\textbf{\hspace{0pt}Purpose}} & {\textbf{\hspace{0pt}Model identifier}} & {\textbf{\hspace{0pt}Reasoning setting}} & {\textbf{\hspace{0pt}Calls per paper}} \\
\midrule
{\hspace{0pt}Repeated integer scores on originals} & {\hspace{0pt}\texttt{gpt-\allowbreak{}5.6-\allowbreak{}sol}} & {\hspace{0pt}medium} & {\hspace{0pt}10} \\
{\hspace{0pt}Repeated integer scores on originals} & {\hspace{0pt}\texttt{gpt-\allowbreak{}6-\allowbreak{}luna}} & {\hspace{0pt}low} & {\hspace{0pt}10} \\
{\hspace{0pt}Repeated integer scores on originals} & {\hspace{0pt}\texttt{gpt-\allowbreak{}6-\allowbreak{}sol}} & {\hspace{0pt}low} & {\hspace{0pt}10} \\
\addlinespace[3pt]
{\hspace{0pt}Calibration target reasoning scores} & {\hspace{0pt}\texttt{gpt-\allowbreak{}5.6-\allowbreak{}sol}} & {\hspace{0pt}medium} & {\hspace{0pt}5} \\
\addlinespace[3pt]
{\hspace{0pt}Quick auxiliary probability calls} & {\hspace{0pt}\texttt{gpt-\allowbreak{}5.6-\allowbreak{}sol}} & {\hspace{0pt}none} & {\hspace{0pt}1 for prediction} \\
\addlinespace[3pt]
{\hspace{0pt}Forced-decimal scoring} & {\hspace{0pt}\texttt{gpt-\allowbreak{}6-\allowbreak{}luna}, \texttt{gpt-\allowbreak{}6-\allowbreak{}sol}} & {\hspace{0pt}low} & {\hspace{0pt}5} \\
\addlinespace[3pt]
{\hspace{0pt}Luna revision scorer} & {\hspace{0pt}\texttt{gpt-\allowbreak{}6-\allowbreak{}luna}} & {\hspace{0pt}low} & {\hspace{0pt}3 each for original and revised} \\
\addlinespace[3pt]
{\hspace{0pt}Haiku revision scorer} & {\hspace{0pt}\texttt{claude-\allowbreak{}haiku-\allowbreak{}5-\allowbreak{}5}} & {\hspace{0pt}low} & {\hspace{0pt}3 each for original and revised} \\
\addlinespace[3pt]
{\hspace{0pt}Writer} & {\hspace{0pt}Luna or Haiku} & {\hspace{0pt}medium} & {\hspace{0pt}1 valid candidate manuscript per paper} \\
\addlinespace[3pt]
{\hspace{0pt}Selector} & {\hspace{0pt}Luna or Haiku} & {\hspace{0pt}high} & {\hspace{0pt}1 valid decision set per paper} \\
\bottomrule
\end{tabular}
\caption{Models and reasoning settings.}
\end{table*}

Each experiment fixes its review configuration. Sections C and D use ten original-manuscript scores per paper, whereas distribution estimation in Section E uses a separately stored set of five target reasoning scores per paper.

Auxiliary probability calls use \texttt{reasoning.effort="none"}, \texttt{top\_\allowbreak{}logprobs=20}, and \texttt{max\_\allowbreak{}output\_\allowbreak{}tokens=4096}, and request output-token log probabilities. Target reasoning scores use the same manuscript input and scoring schema, with \texttt{reasoning.effort="medium"} and an output budget of 16,000 tokens. Temperature is not explicitly set in the requests.

Each review runs in an independent session. Models can access only the preloaded manuscript text and images; tools, web search, memory, and prior context are disabled. Analyses use valid results that pass format and execution checks.

\subsection{Manuscript Inputs and Image Processing}

Each original manuscript is represented by \texttt{text.json}, \texttt{index.json}, and \texttt{figures/\allowbreak{}}. Text records contain a stable \texttt{id}, PDF page number, record type, and text. The \texttt{sequence} in \texttt{index.json} specifies the interleaving of text and images. The input loader follows this sequence to assemble the full text, table text, references, and appendices. Images are loaded after relative-path and SHA-256 checks. Adjacent text records may be merged without changing their order.

All scorers receive preloaded text and images. Haiku processes images in the same order, resizing the longest side to 1,568 pixels and re-encoding them before submission, with a limit of 100 images per call. Paper \texttt{2026-\allowbreak{}paper\_\allowbreak{}067} contains 224 images and exceeds this limit, so the revision crossover experiment uses the remaining 199 papers.

Across the 200 input packages, PDF page markers total 4,923 pages and image-index entries total 2,584. The median, 90th percentile, and maximum page counts are 23, 38, and 69; the corresponding image counts are 9, 23, and 224. There are 133,967 text records containing 15,324,311 characters. Page counts come from PDF page markers in the packages, and image counts from index entries rather than individual figure or table panels.

\subsection{Integer-Scoring Task}

The scorer acts as an ICLR reviewer, reads the complete anonymized manuscript with interleaved images, evaluates novelty, technical correctness, experimental evidence, and limitations, and returns an integer overall score from 1 to 10. The anchors are 1 for strong reject, 3 for reject, 5 for marginally below the acceptance threshold, 6 for marginally above it, 8 for accept, and 10 for strong accept. Other intermediate integers are permitted.

The output contains an overall score, confidence from 1 to 5, a brief summary, up to three strengths, up to three concerns, and a scoring rationale. Key judgments cite PDF page numbers. The scorer does not receive human scores, final decisions, or other reviews. Confidence is not used to weight scores.

\subsection{Forced-Decimal Protocol}

The decimal experiment keeps the inputs, models, and other requirements unchanged, replacing the integer requirement with a mandatory nonzero fractional part. The score field changes from \texttt{integer} to \texttt{number}, while its range remains 1–10. Local validation additionally checks that the score is noninteger. All 1,000 valid scores from each model satisfy this requirement.

An earlier pilot that merely allowed decimals was discarded and is excluded from all analyses. The full forced-decimal experiment reused the 20-paper pilot run under the same protocol, then incorporated the remaining papers to obtain five scores each for 200 papers per model. The integer controls were verified paper by paper and round by round against the first five scores in the ten-review original-manuscript experiment, matching the number of repetitions.

\subsection{Inputs to the Writer, Selector, and Scorer}

\begin{table*}[tp]\centering
\normalsize
\setlength{\tabcolsep}{4pt}
\renewcommand{\arraystretch}{1.12}
\renewcommand{\thetable}{S7}
\begin{tabular}{@{}>{\raggedright\arraybackslash}p{44.290pt}>{\raggedright\arraybackslash}p{216.243pt}>{\raggedright\arraybackslash}p{178.710pt}@{}}
\toprule
{\textbf{\hspace{0pt}Role}} & {\textbf{\hspace{0pt}Input}} & {\textbf{\hspace{0pt}Output}} \\
\midrule
{\hspace{0pt}Writer} & {\hspace{0pt}Full original, images, one specified existing first review} & {\hspace{0pt}List of replacement text records \texttt{id,text}} \\
{\hspace{0pt}Selector} & {\hspace{0pt}Full original, images, all candidate diffs} & {\hspace{0pt}\texttt{id,accept,comment} for each diff} \\
{\hspace{0pt}Scorer} & {\hspace{0pt}Full selected revision and images} & {\hspace{0pt}Same scoring JSON as for originals} \\
\bottomrule
\end{tabular}
\caption{Information inputs and outputs for the three revision roles.}
\end{table*}

The Luna Writer uses the first existing GPT-5.6-Sol review as feedback; source records identify this configuration for all 200 papers. The Haiku Writer uses Haiku's first review of the original manuscript. The two Writer conditions thus follow their respective feedback pipelines. With the Writer fixed, both Selectors read the same candidate diffs. With the final manuscript fixed, both Scorers read the same text version. Scorers receive no candidate diffs, selection rationales, other scorers' scores, or comparative prompt identifying the manuscript as revised.

The Writer returns text records to replace. The Selector gives an acceptance decision and a reason for each candidate edit, with exactly one decision per candidate. The program validates record IDs, decision counts, and output format before generating the final manuscript.

The Writer is instructed to preserve facts, numerical values, data, evidential scope, record order, page numbers, and nontext fields. Leading instructions and page markers must remain unchanged. Returned text is merged into a candidate manuscript, from which block-level diffs are generated. Each Selector comment is limited to 32 characters; truncations of longer Claude comments are recorded. The final manuscript is copied from the original records, replacing only the text of accepted edits with candidate text. Section F.5 reports structural checks.

\section{Extended Score Reliability Results}

\subsection{Metrics and Aggregation Units}

Let the ten scores for paper $i$ be $Y_{i1},\ldots,Y_{i,10}$, with mean $\bar Y_i$. Its unbiased within-paper variance is

\[
S_i^2=\frac{1}{9}\sum_{t=1}^{10}(Y_{it}-\bar Y_i)^2.
\]

The mean within-paper variance is $\hat V=N^{-1}\sum_iS_i^2$, and the within-paper SD reported here is $\sqrt{\hat V}$. The SD of all scores additionally includes differences in mean scores between papers. The identical-10/\allowbreak{}10 rate is the fraction of papers with ten identical observations, and the range is the maximum minus the minimum score. Pairwise exact agreement is first averaged over the $\binom{10}{2}=45$ unordered score pairs within each paper, then across papers.

\begin{table*}[tp]\centering
\normalsize
\setlength{\tabcolsep}{4pt}
\renewcommand{\arraystretch}{1.12}
\renewcommand{\thetable}{S8}
\begin{tabular}{@{}>{\raggedright\arraybackslash}p{92.076pt}>{\raggedleft\arraybackslash}p{58.757pt}>{\raggedleft\arraybackslash}p{52.047pt}>{\raggedleft\arraybackslash}p{65.797pt}>{\raggedleft\arraybackslash}p{77.105pt}>{\raggedleft\arraybackslash}p{69.460pt}@{}}
\toprule
{\textbf{\hspace{0pt}Model}} & {\textbf{\hspace{0pt}Year}} & {\textbf{\hspace{0pt}N}} & {\textbf{\hspace{0pt}Mean}} & {\textbf{\hspace{0pt}Identical 10/\allowbreak{}10}} & {\textbf{\hspace{0pt}Range $\leq$1}} \\
\midrule
{\hspace{0pt}GPT-5.6-Sol} & {\hspace{0pt}All} & {\hspace{0pt}200} & {\hspace{0pt}4.2755} & {\hspace{0pt}30.0\%} & {\hspace{0pt}95.0\%} \\
{\hspace{0pt}GPT-5.6-Sol} & {\hspace{0pt}2025} & {\hspace{0pt}100} & {\hspace{0pt}4.4550} & {\hspace{0pt}24.0\%} & {\hspace{0pt}96.0\%} \\
{\hspace{0pt}GPT-5.6-Sol} & {\hspace{0pt}2026} & {\hspace{0pt}100} & {\hspace{0pt}4.0960} & {\hspace{0pt}36.0\%} & {\hspace{0pt}94.0\%} \\
\addlinespace[3pt]
{\hspace{0pt}GPT-6-Luna} & {\hspace{0pt}All} & {\hspace{0pt}200} & {\hspace{0pt}3.4730} & {\hspace{0pt}32.5\%} & {\hspace{0pt}91.5\%} \\
{\hspace{0pt}GPT-6-Luna} & {\hspace{0pt}2025} & {\hspace{0pt}100} & {\hspace{0pt}3.8050} & {\hspace{0pt}25.0\%} & {\hspace{0pt}93.0\%} \\
{\hspace{0pt}GPT-6-Luna} & {\hspace{0pt}2026} & {\hspace{0pt}100} & {\hspace{0pt}3.1410} & {\hspace{0pt}40.0\%} & {\hspace{0pt}90.0\%} \\
\addlinespace[3pt]
{\hspace{0pt}GPT-6-Sol} & {\hspace{0pt}All} & {\hspace{0pt}200} & {\hspace{0pt}4.6635} & {\hspace{0pt}56.5\%} & {\hspace{0pt}100.0\%} \\
{\hspace{0pt}GPT-6-Sol} & {\hspace{0pt}2025} & {\hspace{0pt}100} & {\hspace{0pt}4.8550} & {\hspace{0pt}61.0\%} & {\hspace{0pt}100.0\%} \\
{\hspace{0pt}GPT-6-Sol} & {\hspace{0pt}2026} & {\hspace{0pt}100} & {\hspace{0pt}4.4720} & {\hspace{0pt}52.0\%} & {\hspace{0pt}100.0\%} \\
\bottomrule
\end{tabular}
\par\medskip
\begin{tabular}{@{}>{\raggedright\arraybackslash}p{80.568pt}>{\raggedleft\arraybackslash}p{47.250pt}>{\raggedleft\arraybackslash}p{40.539pt}>{\raggedleft\arraybackslash}p{73.518pt}>{\raggedleft\arraybackslash}p{85.155pt}>{\raggedleft\arraybackslash}p{88.213pt}@{}}
\toprule
{\textbf{\hspace{0pt}Model}} & {\textbf{\hspace{0pt}Year}} & {\textbf{\hspace{0pt}N}} & {\textbf{\hspace{0pt}Pairwise agreement}} & {\textbf{\hspace{0pt}Mean within-paper variance}} & {\textbf{\hspace{0pt}Within-paper SD}} \\
\midrule
{\hspace{0pt}GPT-5.6-Sol} & {\hspace{0pt}All} & {\hspace{0pt}200} & {\hspace{0pt}72.86\%} & {\hspace{0pt}0.140056} & {\hspace{0pt}0.3742} \\
{\hspace{0pt}GPT-5.6-Sol} & {\hspace{0pt}2025} & {\hspace{0pt}100} & {\hspace{0pt}71.04\%} & {\hspace{0pt}0.146111} & {\hspace{0pt}0.3822} \\
{\hspace{0pt}GPT-5.6-Sol} & {\hspace{0pt}2026} & {\hspace{0pt}100} & {\hspace{0pt}74.67\%} & {\hspace{0pt}0.134000} & {\hspace{0pt}0.3661} \\
\addlinespace[3pt]
{\hspace{0pt}GPT-6-Luna} & {\hspace{0pt}All} & {\hspace{0pt}200} & {\hspace{0pt}72.36\%} & {\hspace{0pt}0.148889} & {\hspace{0pt}0.3859} \\
{\hspace{0pt}GPT-6-Luna} & {\hspace{0pt}2025} & {\hspace{0pt}100} & {\hspace{0pt}69.60\%} & {\hspace{0pt}0.160333} & {\hspace{0pt}0.4004} \\
{\hspace{0pt}GPT-6-Luna} & {\hspace{0pt}2026} & {\hspace{0pt}100} & {\hspace{0pt}75.11\%} & {\hspace{0pt}0.137444} & {\hspace{0pt}0.3707} \\
\addlinespace[3pt]
{\hspace{0pt}GPT-6-Sol} & {\hspace{0pt}All} & {\hspace{0pt}200} & {\hspace{0pt}85.19\%} & {\hspace{0pt}0.074056} & {\hspace{0pt}0.2721} \\
{\hspace{0pt}GPT-6-Sol} & {\hspace{0pt}2025} & {\hspace{0pt}100} & {\hspace{0pt}86.33\%} & {\hspace{0pt}0.068333} & {\hspace{0pt}0.2614} \\
{\hspace{0pt}GPT-6-Sol} & {\hspace{0pt}2026} & {\hspace{0pt}100} & {\hspace{0pt}84.04\%} & {\hspace{0pt}0.079778} & {\hspace{0pt}0.2824} \\
\bottomrule
\end{tabular}
\caption{Complete repeatability results for ten original-manuscript reviews. Each model has 200 papers and 2,000 scores overall, with 100 papers per year.}
\end{table*}

All GPT-6-Sol papers have a score range of at most one point, and 113 have ten identical scores. The corresponding identical-score counts are 60 for GPT-5.6-Sol and 65 for Luna. Year-specific results show the same configuration differences: Sol's within-paper SD is 0.2614 in 2025 and 0.2824 in 2026.

\subsection{Association with Human Scores and Final Acceptance}

Pearson correlation, Spearman correlation, MAE, and RMSE compare each paper's ten-review mean with its public mean human score. AUC uses the same ten-review mean to predict the acceptance labels defined in Section A.4.

\begin{table*}[tp]\centering
\small
\setlength{\tabcolsep}{4pt}
\renewcommand{\arraystretch}{1.12}
\renewcommand{\thetable}{S9}
\begin{tabular}{@{}>{\raggedright\arraybackslash}p{66.123pt}>{\raggedleft\arraybackslash}p{35.833pt}>{\raggedleft\arraybackslash}p{49.173pt}>{\raggedleft\arraybackslash}p{58.623pt}>{\raggedleft\arraybackslash}p{42.233pt}>{\raggedleft\arraybackslash}p{43.623pt}>{\raggedleft\arraybackslash}p{42.233pt}>{\raggedleft\arraybackslash}p{61.403pt}@{}}
\toprule
{\textbf{\hspace{0pt}Model}} & {\textbf{\hspace{0pt}Year}} & {\textbf{\hspace{0pt}Pearson}} & {\textbf{\hspace{0pt}Spearman}} & {\textbf{\hspace{0pt}MAE}} & {\textbf{\hspace{0pt}RMSE}} & {\textbf{\hspace{0pt}AUC: all}} & {\textbf{\hspace{0pt}AUC: excl. withdrawn}} \\
\midrule
{\hspace{0pt}GPT-5.6-Sol} & {\hspace{0pt}All} & {\hspace{0pt}0.4498} & {\hspace{0pt}0.3777} & {\hspace{0pt}1.1034} & {\hspace{0pt}1.4003} & {\hspace{0pt}0.7134} & {\hspace{0pt}0.7128} \\
{\hspace{0pt}GPT-5.6-Sol} & {\hspace{0pt}2025} & {\hspace{0pt}0.5782} & {\hspace{0pt}0.5206} & {\hspace{0pt}1.0878} & {\hspace{0pt}1.3859} & {\hspace{0pt}0.7812} & {\hspace{0pt}0.7558} \\
{\hspace{0pt}GPT-5.6-Sol} & {\hspace{0pt}2026} & {\hspace{0pt}0.2344} & {\hspace{0pt}0.2031} & {\hspace{0pt}1.1190} & {\hspace{0pt}1.4145} & {\hspace{0pt}0.6185} & {\hspace{0pt}0.6438} \\
\addlinespace[3pt]
{\hspace{0pt}GPT-6-Luna} & {\hspace{0pt}All} & {\hspace{0pt}0.4234} & {\hspace{0pt}0.3457} & {\hspace{0pt}1.4597} & {\hspace{0pt}1.8345} & {\hspace{0pt}0.6484} & {\hspace{0pt}0.6438} \\
{\hspace{0pt}GPT-6-Luna} & {\hspace{0pt}2025} & {\hspace{0pt}0.4654} & {\hspace{0pt}0.4017} & {\hspace{0pt}1.5527} & {\hspace{0pt}1.8996} & {\hspace{0pt}0.6756} & {\hspace{0pt}0.6416} \\
{\hspace{0pt}GPT-6-Luna} & {\hspace{0pt}2026} & {\hspace{0pt}0.2190} & {\hspace{0pt}0.1572} & {\hspace{0pt}1.3667} & {\hspace{0pt}1.7672} & {\hspace{0pt}0.6010} & {\hspace{0pt}0.6186} \\
\addlinespace[3pt]
{\hspace{0pt}GPT-6-Sol} & {\hspace{0pt}All} & {\hspace{0pt}0.4629} & {\hspace{0pt}0.4086} & {\hspace{0pt}1.0150} & {\hspace{0pt}1.2581} & {\hspace{0pt}0.7138} & {\hspace{0pt}0.7103} \\
{\hspace{0pt}GPT-6-Sol} & {\hspace{0pt}2025} & {\hspace{0pt}0.5656} & {\hspace{0pt}0.5457} & {\hspace{0pt}0.9450} & {\hspace{0pt}1.1729} & {\hspace{0pt}0.7852} & {\hspace{0pt}0.7639} \\
{\hspace{0pt}GPT-6-Sol} & {\hspace{0pt}2026} & {\hspace{0pt}0.2641} & {\hspace{0pt}0.2345} & {\hspace{0pt}1.0850} & {\hspace{0pt}1.3378} & {\hspace{0pt}0.6185} & {\hspace{0pt}0.6367} \\
\bottomrule
\end{tabular}
\caption{Associations with human scores and acceptance discrimination by year. Full-sample and withdrawal-excluded AUCs use their respective denominators; other metrics use all papers in the specified year.}
\end{table*}

Spearman correlations range from 0.4017 to 0.5457 in 2025 and from 0.1572 to 0.2345 in 2026. Full-sample AUCs range from 0.6756 to 0.7852 in 2025 and from 0.6010 to 0.6185 in 2026. These year-specific results use the scores and labels for the corresponding year.

\subsection{Score Concentration and Paper-Level Variance Distributions}

\begin{table*}[tp]\centering
\small
\setlength{\tabcolsep}{4pt}
\renewcommand{\arraystretch}{1.12}
\renewcommand{\thetable}{S10}
\begin{tabular}{@{}>{\raggedright\arraybackslash}p{69.468pt}>{\raggedleft\arraybackslash}p{28.078pt}>{\raggedleft\arraybackslash}p{33.078pt}>{\raggedleft\arraybackslash}p{33.078pt}>{\raggedleft\arraybackslash}p{33.078pt}>{\raggedleft\arraybackslash}p{38.078pt}>{\raggedleft\arraybackslash}p{33.078pt}>{\raggedleft\arraybackslash}p{28.078pt}>{\raggedleft\arraybackslash}p{28.078pt}>{\raggedleft\arraybackslash}p{23.078pt}>{\raggedleft\arraybackslash}p{28.078pt}@{}}
\toprule
{\textbf{\hspace{0pt}Model}} & {\textbf{\hspace{0pt}1}} & {\textbf{\hspace{0pt}2}} & {\textbf{\hspace{0pt}3}} & {\textbf{\hspace{0pt}4}} & {\textbf{\hspace{0pt}5}} & {\textbf{\hspace{0pt}6}} & {\textbf{\hspace{0pt}7}} & {\textbf{\hspace{0pt}8}} & {\textbf{\hspace{0pt}9}} & {\textbf{\hspace{0pt}10}} \\
\midrule
{\hspace{0pt}GPT-5.6-Sol} & {\hspace{0pt}20} & {\hspace{0pt}64} & {\hspace{0pt}493} & {\hspace{0pt}463} & {\hspace{0pt}751} & {\hspace{0pt}163} & {\hspace{0pt}36} & {\hspace{0pt}3} & {\hspace{0pt}7} & {\hspace{0pt}0} \\
{\hspace{0pt}GPT-6-Luna} & {\hspace{0pt}33} & {\hspace{0pt}405} & {\hspace{0pt}644} & {\hspace{0pt}487} & {\hspace{0pt}383} & {\hspace{0pt}38} & {\hspace{0pt}0} & {\hspace{0pt}10} & {\hspace{0pt}0} & {\hspace{0pt}0} \\
{\hspace{0pt}GPT-6-Sol} & {\hspace{0pt}0} & {\hspace{0pt}19} & {\hspace{0pt}252} & {\hspace{0pt}415} & {\hspace{0pt}1042} & {\hspace{0pt}251} & {\hspace{0pt}11} & {\hspace{0pt}10} & {\hspace{0pt}0} & {\hspace{0pt}0} \\
\bottomrule
\end{tabular}
\par\medskip
\begin{tabular}{@{}>{\raggedright\arraybackslash}p{160.121pt}>{\raggedleft\arraybackslash}p{139.561pt}>{\raggedleft\arraybackslash}p{139.561pt}@{}}
\toprule
{\textbf{\hspace{0pt}Model}} & {\textbf{\hspace{0pt}Share at 5–7}} & {\textbf{\hspace{0pt}Share at 5–6}} \\
\midrule
{\hspace{0pt}GPT-5.6-Sol} & {\hspace{0pt}47.50\%} & {\hspace{0pt}45.70\%} \\
{\hspace{0pt}GPT-6-Luna} & {\hspace{0pt}21.05\%} & {\hspace{0pt}21.05\%} \\
{\hspace{0pt}GPT-6-Sol} & {\hspace{0pt}65.20\%} & {\hspace{0pt}64.65\%} \\
\bottomrule
\end{tabular}
\caption{Counts at scores 1–10 and score concentration. Each row contains all 2,000 scores for one model. The intervals 5–7 and 5–6 include both endpoints.}
\end{table*}

Luna's scores are concentrated at 2–4. Sol assigns a score of 5 in 1,042 cases, or 52.1\% of its 2,000 scores. Scores from 5 to 7 account for 65.20\% of Sol's outputs, most of them at 5 or 6. The range of repeated scores and numerical separation between papers therefore describe different aspects of scoring.

\begin{table*}[tp]\centering
\small
\setlength{\tabcolsep}{4pt}
\renewcommand{\arraystretch}{1.12}
\renewcommand{\thetable}{S11}
\begin{tabular}{@{}>{\raggedright\arraybackslash}p{73.298pt}>{\raggedleft\arraybackslash}p{64.688pt}>{\raggedleft\arraybackslash}p{49.408pt}>{\raggedleft\arraybackslash}p{54.688pt}>{\raggedleft\arraybackslash}p{49.408pt}>{\raggedleft\arraybackslash}p{49.408pt}>{\raggedleft\arraybackslash}p{66.348pt}@{}}
\toprule
{\textbf{\hspace{0pt}Model}} & {\textbf{\hspace{0pt}Minimum variance}} & {\textbf{\hspace{0pt}Q25}} & {\textbf{\hspace{0pt}Median}} & {\textbf{\hspace{0pt}Q75}} & {\textbf{\hspace{0pt}Q90}} & {\textbf{\hspace{0pt}Maximum}} \\
\midrule
{\hspace{0pt}GPT-5.6-Sol} & {\hspace{0pt}0.0000} & {\hspace{0pt}0.0000} & {\hspace{0pt}0.1389} & {\hspace{0pt}0.2333} & {\hspace{0pt}0.2667} & {\hspace{0pt}0.4889} \\
{\hspace{0pt}GPT-6-Luna} & {\hspace{0pt}0.0000} & {\hspace{0pt}0.0000} & {\hspace{0pt}0.1778} & {\hspace{0pt}0.2417} & {\hspace{0pt}0.2778} & {\hspace{0pt}0.6667} \\
{\hspace{0pt}GPT-6-Sol} & {\hspace{0pt}0.0000} & {\hspace{0pt}0.0000} & {\hspace{0pt}0.0000} & {\hspace{0pt}0.1000} & {\hspace{0pt}0.2333} & {\hspace{0pt}0.2778} \\
\bottomrule
\end{tabular}
\par\medskip
\begin{tabular}{@{}>{\raggedright\arraybackslash}p{103.985pt}>{\raggedleft\arraybackslash}p{79.815pt}>{\raggedleft\arraybackslash}p{79.815pt}>{\raggedleft\arraybackslash}p{79.815pt}>{\raggedleft\arraybackslash}p{79.815pt}@{}}
\toprule
{\textbf{\hspace{0pt}Model}} & {\textbf{\hspace{0pt}Range 0}} & {\textbf{\hspace{0pt}Range 1}} & {\textbf{\hspace{0pt}Range 2}} & {\textbf{\hspace{0pt}Range 3}} \\
\midrule
{\hspace{0pt}GPT-5.6-Sol} & {\hspace{0pt}60} & {\hspace{0pt}130} & {\hspace{0pt}10} & {\hspace{0pt}0} \\
{\hspace{0pt}GPT-6-Luna} & {\hspace{0pt}65} & {\hspace{0pt}118} & {\hspace{0pt}17} & {\hspace{0pt}0} \\
{\hspace{0pt}GPT-6-Sol} & {\hspace{0pt}113} & {\hspace{0pt}87} & {\hspace{0pt}0} & {\hspace{0pt}0} \\
\bottomrule
\end{tabular}
\caption{Quantiles of unbiased paper-level sample variance and counts by score range. Quantiles use NumPy's default linear interpolation; ranges use all ten scores.}
\end{table*}

\begin{figure*}[tp]\centering
\includegraphics[width=\textwidth]{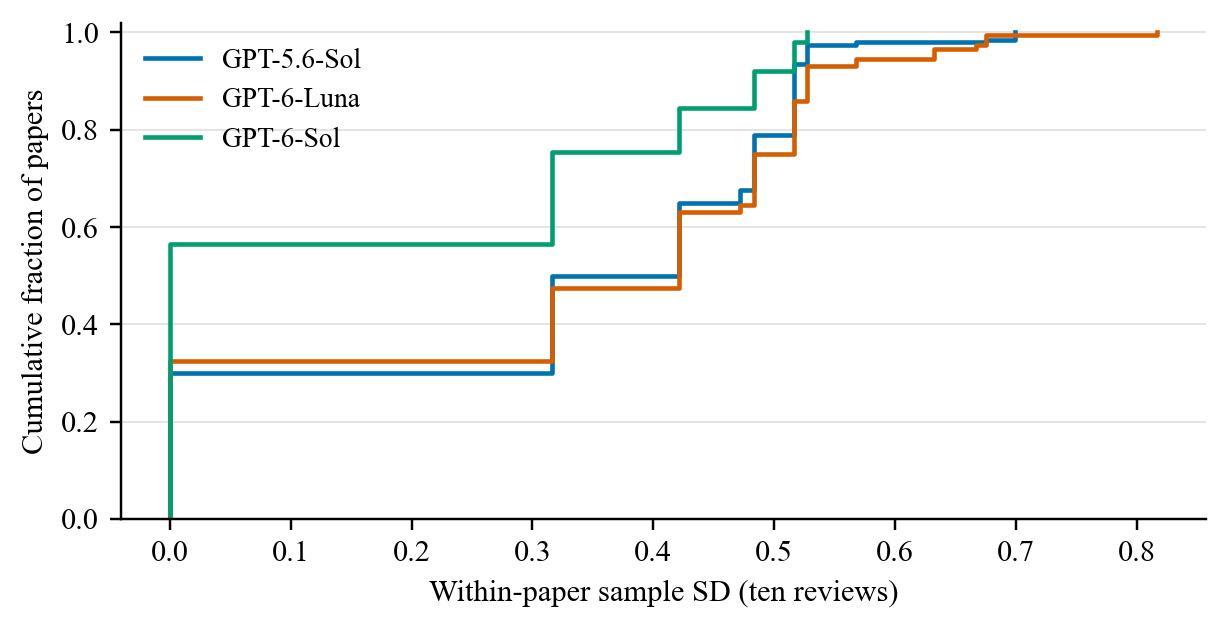}
\renewcommand{\thefigure}{C1}
\caption{Empirical cumulative distributions of paper-level sample SD for the three models, including all 200 papers.}
\end{figure*}

\subsection{First Five Versus Last Five Reviews}

The ten reviews combine historical scores with later runs under the same configuration. The table separates rounds 1–5 from rounds 6–10, computes the difference between batch means within each paper, and bootstraps papers.

\begin{table*}[tp]\centering
\normalsize
\setlength{\tabcolsep}{4pt}
\renewcommand{\arraystretch}{1.12}
\renewcommand{\thetable}{S12}
\begin{tabular}{@{}>{\raggedright\arraybackslash}p{78.269pt}>{\raggedleft\arraybackslash}p{65.113pt}>{\raggedleft\arraybackslash}p{63.287pt}>{\raggedleft\arraybackslash}p{144.137pt}>{\raggedleft\arraybackslash}p{72.439pt}@{}}
\toprule
{\textbf{\hspace{0pt}Model}} & {\textbf{\hspace{0pt}First-five mean}} & {\textbf{\hspace{0pt}Last-five mean}} & {\textbf{\hspace{0pt}Last$-$first and 95\% CI}} & {\textbf{\hspace{0pt}Mean absolute difference: ten versus first five}} \\
\midrule
{\hspace{0pt}GPT-5.6-Sol} & {\hspace{0pt}4.2790} & {\hspace{0pt}4.2720} & {\hspace{0pt}-0.0070 [-0.0410, +0.0270]} & {\hspace{0pt}0.0835} \\
{\hspace{0pt}GPT-6-Luna} & {\hspace{0pt}3.4590} & {\hspace{0pt}3.4870} & {\hspace{0pt}+0.0280 [-0.0090, +0.0650]} & {\hspace{0pt}0.0890} \\
{\hspace{0pt}GPT-6-Sol} & {\hspace{0pt}4.6530} & {\hspace{0pt}4.6740} & {\hspace{0pt}+0.0210 [-0.0020, +0.0440]} & {\hspace{0pt}0.0465} \\
\bottomrule
\end{tabular}
\caption{Mean-score comparison between the two batches of five reviews. Intervals use 10,000 paper-level bootstrap resamples. Mean absolute differences compare the ten-review mean with the first-five mean.}
\end{table*}

For all three models, the interval for the overall difference between batches includes zero. This analysis records shifts between batch positions; the precision budgets in Section D concern the expectation of future independent scores under a fixed configuration.

\subsection{Ranking Reproducibility Across All Ten Reviews}

For each paper, we construct two nonoverlapping blocks of $k$ scores from its ten reviews. Treating the pair of blocks as unordered gives

\[
\begin{gathered}L_k=\frac12\binom{10}{k}\binom{10-k}{k},\\ 
 k=1,\ldots,5.\end{gathered}
\]

The resulting numbers of partitions are 45, 630, 2,100, 1,575, and 126. For each partition, block means are compared between papers from the same year, then averaged over partitions and paper pairs. The two years yield $2\binom{100}{2}=9,900$ distinct paper pairs.

Same denotes the same strict ordering in both blocks; Tie denotes a tie in at least one block; and Reverse denotes opposite strict orderings. These categories are mutually exclusive and exhaustive.

\begin{table*}[tp]\centering
\normalsize
\setlength{\tabcolsep}{4pt}
\renewcommand{\arraystretch}{1.12}
\renewcommand{\thetable}{S13}
\begin{tabular}{@{}>{\raggedright\arraybackslash}p{84.646pt}>{\raggedleft\arraybackslash}p{66.606pt}>{\raggedleft\arraybackslash}p{74.559pt}>{\raggedleft\arraybackslash}p{62.030pt}>{\raggedleft\arraybackslash}p{62.030pt}>{\raggedleft\arraybackslash}p{65.374pt}@{}}
\toprule
{\textbf{\hspace{0pt}Model}} & {\textbf{\hspace{0pt}Reviews per block}} & {\textbf{\hspace{0pt}Partitions}} & {\textbf{\hspace{0pt}Same}} & {\textbf{\hspace{0pt}Tie}} & {\textbf{\hspace{0pt}Reverse}} \\
\midrule
{\hspace{0pt}GPT-5.6-Sol} & {\hspace{0pt}1} & {\hspace{0pt}45} & {\hspace{0pt}62.14\%} & {\hspace{0pt}36.71\%} & {\hspace{0pt}1.15\%} \\
{\hspace{0pt}GPT-5.6-Sol} & {\hspace{0pt}2} & {\hspace{0pt}630} & {\hspace{0pt}73.52\%} & {\hspace{0pt}24.74\%} & {\hspace{0pt}1.74\%} \\
{\hspace{0pt}GPT-5.6-Sol} & {\hspace{0pt}3} & {\hspace{0pt}2100} & {\hspace{0pt}78.92\%} & {\hspace{0pt}19.08\%} & {\hspace{0pt}2.00\%} \\
{\hspace{0pt}GPT-5.6-Sol} & {\hspace{0pt}4} & {\hspace{0pt}1575} & {\hspace{0pt}82.18\%} & {\hspace{0pt}15.71\%} & {\hspace{0pt}2.10\%} \\
{\hspace{0pt}GPT-5.6-Sol} & {\hspace{0pt}5} & {\hspace{0pt}126} & {\hspace{0pt}84.47\%} & {\hspace{0pt}13.38\%} & {\hspace{0pt}2.15\%} \\
\addlinespace[3pt]
{\hspace{0pt}GPT-6-Luna} & {\hspace{0pt}1} & {\hspace{0pt}45} & {\hspace{0pt}63.00\%} & {\hspace{0pt}35.60\%} & {\hspace{0pt}1.40\%} \\
{\hspace{0pt}GPT-6-Luna} & {\hspace{0pt}2} & {\hspace{0pt}630} & {\hspace{0pt}74.91\%} & {\hspace{0pt}23.05\%} & {\hspace{0pt}2.04\%} \\
{\hspace{0pt}GPT-6-Luna} & {\hspace{0pt}3} & {\hspace{0pt}2100} & {\hspace{0pt}80.40\%} & {\hspace{0pt}17.30\%} & {\hspace{0pt}2.30\%} \\
{\hspace{0pt}GPT-6-Luna} & {\hspace{0pt}4} & {\hspace{0pt}1575} & {\hspace{0pt}83.58\%} & {\hspace{0pt}14.03\%} & {\hspace{0pt}2.39\%} \\
{\hspace{0pt}GPT-6-Luna} & {\hspace{0pt}5} & {\hspace{0pt}126} & {\hspace{0pt}85.71\%} & {\hspace{0pt}11.85\%} & {\hspace{0pt}2.44\%} \\
\addlinespace[3pt]
{\hspace{0pt}GPT-6-Sol} & {\hspace{0pt}1} & {\hspace{0pt}45} & {\hspace{0pt}57.52\%} & {\hspace{0pt}42.14\%} & {\hspace{0pt}0.34\%} \\
{\hspace{0pt}GPT-6-Sol} & {\hspace{0pt}2} & {\hspace{0pt}630} & {\hspace{0pt}66.43\%} & {\hspace{0pt}33.03\%} & {\hspace{0pt}0.54\%} \\
{\hspace{0pt}GPT-6-Sol} & {\hspace{0pt}3} & {\hspace{0pt}2100} & {\hspace{0pt}70.24\%} & {\hspace{0pt}29.11\%} & {\hspace{0pt}0.65\%} \\
{\hspace{0pt}GPT-6-Sol} & {\hspace{0pt}4} & {\hspace{0pt}1575} & {\hspace{0pt}72.56\%} & {\hspace{0pt}26.71\%} & {\hspace{0pt}0.73\%} \\
{\hspace{0pt}GPT-6-Sol} & {\hspace{0pt}5} & {\hspace{0pt}126} & {\hspace{0pt}74.13\%} & {\hspace{0pt}25.08\%} & {\hspace{0pt}0.79\%} \\
\bottomrule
\end{tabular}
\caption{Complete ranking reproducibility results for 1–5 reviews per block. Percentages are computed before rounding; displayed values may differ slightly from a sum of 100\%.}
\end{table*}

\subsection{Close-Pair Analysis with Independent Grouping Scores}

To examine close paper pairs, we group pairs by the absolute difference between their first-five means, reserving the last five scores for ranking comparisons. Among the last five, single-review blocks enumerate all 10 partitions, and two-review blocks enumerate all 15 partitions. Grouping and ranking comparisons thus use separate scores.

The first-five mean differences are grouped as 0, $(0,0.4]$, $(0.4,1]$, $(1,2]$, and greater than 2 points. These differences are finite-sample estimates. Same, Tie, and Reverse below measure reproducibility between two subsequent independent review blocks, without treating the grouping estimate as the correct ordering.

\begin{table*}[tp]\centering
\normalsize
\setlength{\tabcolsep}{4pt}
\renewcommand{\arraystretch}{1.12}
\renewcommand{\thetable}{S14}
\begin{tabular}{@{}>{\raggedright\arraybackslash}p{83.634pt}>{\raggedleft\arraybackslash}p{74.141pt}>{\raggedleft\arraybackslash}p{55.210pt}>{\raggedleft\arraybackslash}p{67.420pt}>{\raggedleft\arraybackslash}p{67.420pt}>{\raggedleft\arraybackslash}p{67.420pt}@{}}
\toprule
{\textbf{\hspace{0pt}Model}} & {\textbf{\hspace{0pt}First-five mean difference}} & {\textbf{\hspace{0pt}Paper pairs}} & {\textbf{\hspace{0pt}1-review Same}} & {\textbf{\hspace{0pt}1-review Tie}} & {\textbf{\hspace{0pt}1-review Reverse}} \\
\midrule
{\hspace{0pt}GPT-5.6-Sol} & {\hspace{0pt}0} & {\hspace{0pt}824} & {\hspace{0pt}6.57\%} & {\hspace{0pt}91.40\%} & {\hspace{0pt}2.04\%} \\
{\hspace{0pt}GPT-5.6-Sol} & {\hspace{0pt}(0, 0.4]} & {\hspace{0pt}1920} & {\hspace{0pt}17.33\%} & {\hspace{0pt}79.49\%} & {\hspace{0pt}3.17\%} \\
{\hspace{0pt}GPT-5.6-Sol} & {\hspace{0pt}(0.4, 1]} & {\hspace{0pt}2364} & {\hspace{0pt}52.94\%} & {\hspace{0pt}45.63\%} & {\hspace{0pt}1.43\%} \\
{\hspace{0pt}GPT-5.6-Sol} & {\hspace{0pt}(1, 2]} & {\hspace{0pt}3251} & {\hspace{0pt}91.51\%} & {\hspace{0pt}8.36\%} & {\hspace{0pt}0.13\%} \\
{\hspace{0pt}GPT-5.6-Sol} & {\hspace{0pt}>2} & {\hspace{0pt}1541} & {\hspace{0pt}99.97\%} & {\hspace{0pt}0.03\%} & {\hspace{0pt}0.00\%} \\
\addlinespace[3pt]
{\hspace{0pt}GPT-6-Luna} & {\hspace{0pt}0} & {\hspace{0pt}775} & {\hspace{0pt}7.65\%} & {\hspace{0pt}89.63\%} & {\hspace{0pt}2.72\%} \\
{\hspace{0pt}GPT-6-Luna} & {\hspace{0pt}(0, 0.4]} & {\hspace{0pt}1832} & {\hspace{0pt}22.25\%} & {\hspace{0pt}74.84\%} & {\hspace{0pt}2.91\%} \\
{\hspace{0pt}GPT-6-Luna} & {\hspace{0pt}(0.4, 1]} & {\hspace{0pt}2828} & {\hspace{0pt}57.86\%} & {\hspace{0pt}40.80\%} & {\hspace{0pt}1.34\%} \\
{\hspace{0pt}GPT-6-Luna} & {\hspace{0pt}(1, 2]} & {\hspace{0pt}2926} & {\hspace{0pt}91.17\%} & {\hspace{0pt}8.74\%} & {\hspace{0pt}0.10\%} \\
{\hspace{0pt}GPT-6-Luna} & {\hspace{0pt}>2} & {\hspace{0pt}1539} & {\hspace{0pt}99.80\%} & {\hspace{0pt}0.20\%} & {\hspace{0pt}0.00\%} \\
\addlinespace[3pt]
{\hspace{0pt}GPT-6-Sol} & {\hspace{0pt}0} & {\hspace{0pt}2062} & {\hspace{0pt}1.13\%} & {\hspace{0pt}98.67\%} & {\hspace{0pt}0.20\%} \\
{\hspace{0pt}GPT-6-Sol} & {\hspace{0pt}(0, 0.4]} & {\hspace{0pt}1425} & {\hspace{0pt}13.92\%} & {\hspace{0pt}85.01\%} & {\hspace{0pt}1.07\%} \\
{\hspace{0pt}GPT-6-Sol} & {\hspace{0pt}(0.4, 1]} & {\hspace{0pt}3047} & {\hspace{0pt}74.25\%} & {\hspace{0pt}25.46\%} & {\hspace{0pt}0.29\%} \\
{\hspace{0pt}GPT-6-Sol} & {\hspace{0pt}(1, 2]} & {\hspace{0pt}2688} & {\hspace{0pt}97.34\%} & {\hspace{0pt}2.66\%} & {\hspace{0pt}0.00\%} \\
{\hspace{0pt}GPT-6-Sol} & {\hspace{0pt}>2} & {\hspace{0pt}678} & {\hspace{0pt}100.00\%} & {\hspace{0pt}0.00\%} & {\hspace{0pt}0.00\%} \\
\bottomrule
\end{tabular}
\par\medskip
\begin{tabular}{@{}>{\raggedright\arraybackslash}p{83.634pt}>{\raggedleft\arraybackslash}p{74.141pt}>{\raggedleft\arraybackslash}p{55.210pt}>{\raggedleft\arraybackslash}p{67.420pt}>{\raggedleft\arraybackslash}p{67.420pt}>{\raggedleft\arraybackslash}p{67.420pt}@{}}
\toprule
{\textbf{\hspace{0pt}Model}} & {\textbf{\hspace{0pt}First-five mean difference}} & {\textbf{\hspace{0pt}Paper pairs}} & {\textbf{\hspace{0pt}2-review Same}} & {\textbf{\hspace{0pt}2-review Tie}} & {\textbf{\hspace{0pt}2-review Reverse}} \\
\midrule
{\hspace{0pt}GPT-5.6-Sol} & {\hspace{0pt}0} & {\hspace{0pt}824} & {\hspace{0pt}14.07\%} & {\hspace{0pt}82.16\%} & {\hspace{0pt}3.77\%} \\
{\hspace{0pt}GPT-5.6-Sol} & {\hspace{0pt}(0, 0.4]} & {\hspace{0pt}1920} & {\hspace{0pt}33.44\%} & {\hspace{0pt}61.59\%} & {\hspace{0pt}4.98\%} \\
{\hspace{0pt}GPT-5.6-Sol} & {\hspace{0pt}(0.4, 1]} & {\hspace{0pt}2364} & {\hspace{0pt}74.64\%} & {\hspace{0pt}23.58\%} & {\hspace{0pt}1.78\%} \\
{\hspace{0pt}GPT-5.6-Sol} & {\hspace{0pt}(1, 2]} & {\hspace{0pt}3251} & {\hspace{0pt}98.15\%} & {\hspace{0pt}1.81\%} & {\hspace{0pt}0.04\%} \\
{\hspace{0pt}GPT-5.6-Sol} & {\hspace{0pt}>2} & {\hspace{0pt}1541} & {\hspace{0pt}100.00\%} & {\hspace{0pt}0.00\%} & {\hspace{0pt}0.00\%} \\
\addlinespace[3pt]
{\hspace{0pt}GPT-6-Luna} & {\hspace{0pt}0} & {\hspace{0pt}775} & {\hspace{0pt}16.70\%} & {\hspace{0pt}78.43\%} & {\hspace{0pt}4.88\%} \\
{\hspace{0pt}GPT-6-Luna} & {\hspace{0pt}(0, 0.4]} & {\hspace{0pt}1832} & {\hspace{0pt}39.53\%} & {\hspace{0pt}55.73\%} & {\hspace{0pt}4.73\%} \\
{\hspace{0pt}GPT-6-Luna} & {\hspace{0pt}(0.4, 1]} & {\hspace{0pt}2828} & {\hspace{0pt}77.43\%} & {\hspace{0pt}20.80\%} & {\hspace{0pt}1.77\%} \\
{\hspace{0pt}GPT-6-Luna} & {\hspace{0pt}(1, 2]} & {\hspace{0pt}2926} & {\hspace{0pt}97.95\%} & {\hspace{0pt}1.98\%} & {\hspace{0pt}0.08\%} \\
{\hspace{0pt}GPT-6-Luna} & {\hspace{0pt}>2} & {\hspace{0pt}1539} & {\hspace{0pt}99.99\%} & {\hspace{0pt}0.01\%} & {\hspace{0pt}0.00\%} \\
\addlinespace[3pt]
{\hspace{0pt}GPT-6-Sol} & {\hspace{0pt}0} & {\hspace{0pt}2062} & {\hspace{0pt}2.86\%} & {\hspace{0pt}96.81\%} & {\hspace{0pt}0.33\%} \\
{\hspace{0pt}GPT-6-Sol} & {\hspace{0pt}(0, 0.4]} & {\hspace{0pt}1425} & {\hspace{0pt}29.74\%} & {\hspace{0pt}68.45\%} & {\hspace{0pt}1.82\%} \\
{\hspace{0pt}GPT-6-Sol} & {\hspace{0pt}(0.4, 1]} & {\hspace{0pt}3047} & {\hspace{0pt}90.13\%} & {\hspace{0pt}9.56\%} & {\hspace{0pt}0.31\%} \\
{\hspace{0pt}GPT-6-Sol} & {\hspace{0pt}(1, 2]} & {\hspace{0pt}2688} & {\hspace{0pt}99.58\%} & {\hspace{0pt}0.42\%} & {\hspace{0pt}0.00\%} \\
{\hspace{0pt}GPT-6-Sol} & {\hspace{0pt}>2} & {\hspace{0pt}678} & {\hspace{0pt}100.00\%} & {\hspace{0pt}0.00\%} & {\hspace{0pt}0.00\%} \\
\bottomrule
\end{tabular}
\caption{Ranking reproducibility after grouping by independent score differences. Rounds 1–5 define groups; rounds 6–10 provide comparisons. Each model includes all 9,900 within-year paper pairs.}
\end{table*}

For pairs whose first-five mean difference lies in $(0,0.4]$, subsequent single-review comparisons produce ties in 74.84\%–85.01\% of cases; with two-review means, this remains 55.73\%–68.45\%. For the $(1,2]$ group, strict same-order rates with two-review means reach 97.95\%–99.58\%. Improvements averaged across all pairs therefore combine contributions from different score-gap regions.

\begin{figure*}[tp]\centering
\includegraphics[width=\textwidth]{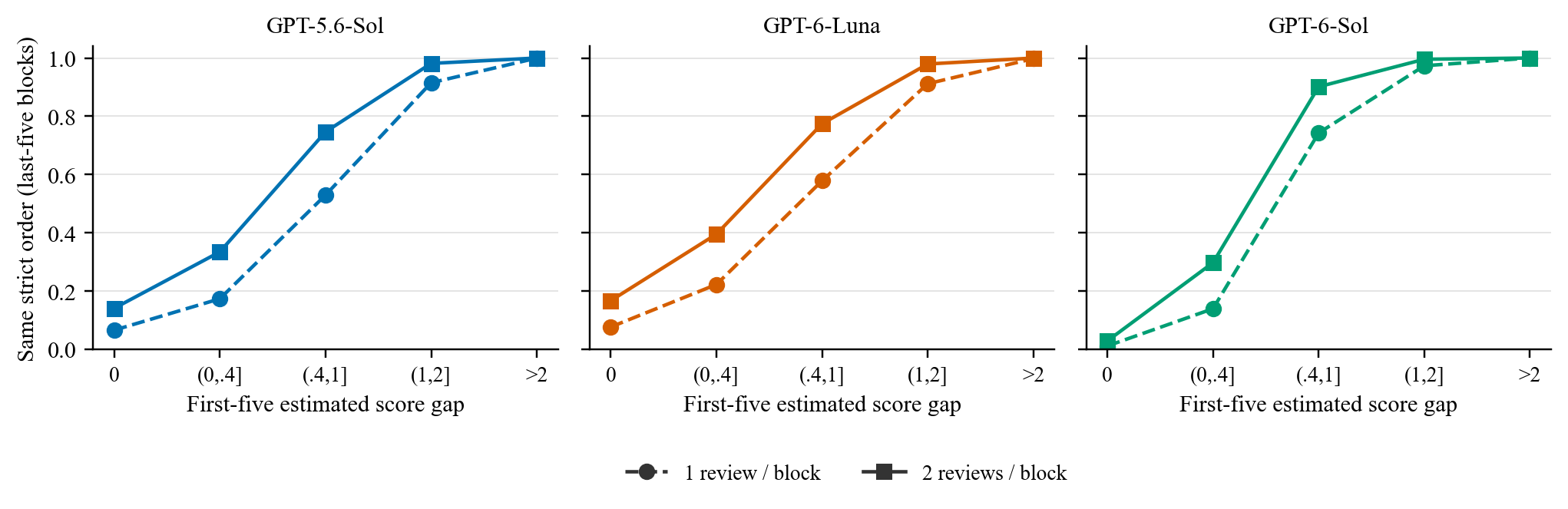}
\renewcommand{\thefigure}{C2}
\caption{Strict same-order rates from the last five reviews. The first five define the horizontal-axis groups. Solid lines use two reviews per block, and dashed lines use one. Groups are defined separately for each model; full denominators appear in the table above.}
\end{figure*}

Each paper participates in multiple pairs, creating dependence across pairs. We enumerate all partitions exactly and report descriptive proportions, without treating the 9,900 paper pairs as independent observations for binomial intervals.

\subsection{Semantic Consistency in the Existing Review Corpus}

The existing text-analysis corpus contains five original-manuscript reviews each from Luna and Sol and three from Haiku, for a shared set of 199 papers. It excludes the later rounds 6–10. We use every available round in this fixed corpus: five reviews yield 10 within-model pairs, three yield 3, and cross-model comparisons yield 25 or 15 pairs.

Each strength and concern is encoded with \texttt{sentence-\allowbreak{}transformers/\allowbreak{}all-\allowbreak{}mpnet-\allowbreak{}base-\allowbreak{}v2}. For two lists of the same polarity, we compute the pairwise cosine-similarity matrix and use the Hungarian algorithm for one-to-one matching. The sum of matched cosines is divided by the longer list's length, with unmatched items contributing zero:

\[
S(A,B)=\frac{\max_{\mathcal M}\sum_{(a,b)\in\mathcal M}\cos(e_a,e_b)}{\max(|A|,|B|)}.
\]

All strength and concern lists in the corpus are nonempty. Review-pair values are first averaged within each paper and condition, then across papers. The table covers 15,522 review pairs. Confidence intervals retain the original analysis's 5,000 paper-level bootstrap resamples.

\begin{table*}[tp]\centering
\normalsize
\setlength{\tabcolsep}{4pt}
\renewcommand{\arraystretch}{1.12}
\renewcommand{\thetable}{S15}
\begin{tabular}{@{}>{\raggedright\arraybackslash}p{108.697pt}>{\raggedright\arraybackslash}p{40.915pt}>{\raggedleft\arraybackslash}p{39.056pt}>{\raggedleft\arraybackslash}p{37.230pt}>{\raggedleft\arraybackslash}p{94.672pt}>{\raggedleft\arraybackslash}p{94.672pt}@{}}
\toprule
{\textbf{\hspace{0pt}Comparison}} & {\textbf{\hspace{0pt}Rounds}} & {\textbf{\hspace{0pt}Review pairs per paper}} & {\textbf{\hspace{0pt}Papers}} & {\textbf{\hspace{0pt}Strengths and 95\% CI}} & {\textbf{\hspace{0pt}Concerns and 95\% CI}} \\
\midrule
{\hspace{0pt}Luna repeated reviews} & {\hspace{0pt}5} & {\hspace{0pt}10} & {\hspace{0pt}199} & {\hspace{0pt}0.544 [0.534, 0.553]} & {\hspace{0pt}0.618 [0.611, 0.625]} \\
{\hspace{0pt}Sol repeated reviews} & {\hspace{0pt}5} & {\hspace{0pt}10} & {\hspace{0pt}199} & {\hspace{0pt}0.545 [0.532, 0.558]} & {\hspace{0pt}0.639 [0.630, 0.647]} \\
{\hspace{0pt}Haiku repeated reviews} & {\hspace{0pt}3} & {\hspace{0pt}3} & {\hspace{0pt}199} & {\hspace{0pt}0.569 [0.558, 0.580]} & {\hspace{0pt}0.607 [0.599, 0.615]} \\
{\hspace{0pt}Luna–Sol} & {\hspace{0pt}5$\times$5} & {\hspace{0pt}25} & {\hspace{0pt}199} & {\hspace{0pt}0.459 [0.451, 0.467]} & {\hspace{0pt}0.545 [0.539, 0.552]} \\
{\hspace{0pt}Luna–Haiku} & {\hspace{0pt}5$\times$3} & {\hspace{0pt}15} & {\hspace{0pt}199} & {\hspace{0pt}0.464 [0.457, 0.472]} & {\hspace{0pt}0.532 [0.526, 0.538]} \\
{\hspace{0pt}Sol–Haiku} & {\hspace{0pt}5$\times$3} & {\hspace{0pt}15} & {\hspace{0pt}199} & {\hspace{0pt}0.419 [0.410, 0.428]} & {\hspace{0pt}0.502 [0.495, 0.508]} \\
\bottomrule
\end{tabular}
\caption{Semantic proximity in the fixed five-/\allowbreak{}three-review original-manuscript corpus. Values are mean cosines after list matching, not percentages of viewpoint agreement.}
\end{table*}

Repeated reviews from the same model have closer opinion lists on average, and concern lists are closer than strength lists. An existing assistant inspection of 40 matched opinion pairs found shared cores, partial overlaps, and complementary observations; it does not constitute a human blind-evaluation gold standard. General-purpose NLI also misclassified some compound technical statements. We therefore retain the reproducible continuous semantic measure rather than converting automatic entailment thresholds into rates of substantive agreement.

\section{Review Budgets and Mean-Score Precision}

\subsection{Average Success Probability Across Papers}

We give equal weight to a fixed set of $N=200$ papers. For paper $i$, let $\mu_i$ and $\sigma_i^2$ denote the mean and variance of scores under a fixed configuration. Let $\bar Y_{i,k}$ be the mean of $k$ future independent scores, and define

\[
\begin{gathered}A(k,\epsilon)=\frac1N\sum_{i=1}^{N}\Pr\bigl(|\bar Y_{i,k}-\mu_i|<\epsilon\bigr),
\\ 
 V=\frac1N\sum_{i=1}^{N}\sigma_i^2.\end{gathered}
\]

The target is $A(k,\epsilon)\ge1-\beta$. This is the overall probability that the mean of $k$ scores estimates the model's expected score to within $\epsilon$ for a uniformly selected sample paper. The main paper sets $\beta=0.05$ and considers $\epsilon=0.25$ and $0.5$.

Under independent, identically distributed scoring within each paper, $\operatorname{Var}(\bar Y_{i,k})=\sigma_i^2/k$. Applying Chebyshev's inequality to the failure event $|\bar Y_{i,k}-\mu_i|\ge\epsilon$ gives

\[
\begin{gathered}A(k,\epsilon)\ge1-\frac{V}{k\epsilon^2},
\\ 
 k\ge\left\lceil\frac{V}{\beta\epsilon^2}\right\rceil.\end{gathered}
\]

This expression gives a sufficient budget when the score scale is known and the mean and variance exist. It requires neither a between-paper score gap nor a normality assumption or an empirical histogram as ground truth.

\subsection{Planning Budgets from Observed Variance}

With $n$ pilot scores per paper, estimate within-paper variance by $\hat V=N^{-1}\sum_iS_i^2$. Substituting $\hat V$ into the preceding expression gives the Planning budget. With ten pilot scores, the three configurations have $\hat V$ values of 0.140056, 0.148889, and 0.074056. Planning budgets are 45, 48, and 24 at a tolerance of 0.25 points, and 12, 12, and 6 at 0.5 points.

Planning expresses the budget implied by the observed variance. The next section also accounts for uncertainty in variance estimation from a finite pilot sample.

\subsection{An Upper Bound on Mean Variance from Nonoverlapping Pairs}

Scores lie in $[a,b]=[1,10]$. Define the symmetric kernel

\[
\begin{gathered}h(y,y')=\frac{(y-y')^2}{2},\\ 
 0\le h\le B=\frac{(b-a)^2}{2}=40.5.\end{gathered}
\]

For two independent, identically distributed scores, $\mathbb E h(Y,Y')=\sigma_i^2$. The unbiased sample variance for each paper also equals the average kernel value over all score pairs:

\[
S_i^2=\binom n2^{-1}\sum_{r<s}h(Y_{ir},Y_{is}).
\]

Let $K=\lfloor n/2\rfloor$. Randomly permute each paper's $n$ scores and form $K$ nonoverlapping pairs from the first $2K$ scores. Let $T_i$ be their mean kernel value and $T=N^{-1}\sum_iT_i$ its average across papers. Conditional on the observed scores, every unordered pair has the same probability of selection, so

\[
\begin{gathered}\mathbb E_{\pi}[T_i\mid Y_{i1},\ldots,Y_{in}]=S_i^2,
\\ 
 \mathbb E_{\pi}[T\mid Y]=\hat V.\end{gathered}
\]

For fixed permutations, there are $M=NK$ nonoverlapping kernel values. Assuming calls are independent across papers, these values are independent, and their expectations average to $V$. For any real $\lambda$, Jensen's inequality gives

\[
\begin{aligned}&\mathbb E\exp\!\left(\lambda M\frac{\hat V}{B}\right)\\ &\le \mathbb E\exp\!\left(\lambda M\frac{T}{B}\right).\end{aligned}
\]

For $X\in[0,1]$, the chord bound for the exponential function gives

\[
e^{\lambda X}\le1+X(e^\lambda-1).
\]

The kernel values may have different expectations. Multiplying the moment-generating functions of the independent variables and applying concavity of the logarithm yields a bound depending only on the overall mean $V/B$:

\[
\begin{aligned}&\mathbb E\exp\!\left(\lambda M\frac{\hat V}{B}\right)\\ &\le \left[1+\frac VB(e^\lambda-1)\right]^M.\end{aligned}
\]

For $v<V$, taking $\lambda<0$ and optimizing the Chernoff parameter gives

\[
\Pr(\hat V\le v)\le\exp\!\left[-M\operatorname{kl}\!\left(\frac vB,\frac VB\right)\right],
\]

where

\[
\operatorname{kl}(p,q)=p\log\frac pq+(1-p)\log\frac{1-p}{1-q}.
\]

Given the observed $\hat V$, let $U_\alpha$ be the root above $\hat V$ of

\[
M\operatorname{kl}\!\left(\frac{\hat V}{B},\frac{U_\alpha}{B}\right)=\log\frac1\alpha.
\]

Inverting the lower-tail bound yields

\[
\Pr(V>U_\alpha)\le\alpha.
\]

The score range also implies $V\le(b-a)^2/4=20.25$, so the final upper bound can be capped at 20.25. The derivation uses a Chernoff argument for bounded variables and random grouping of a U-statistic; see \citet{hoeffding1963probability}.

With ten pilot reviews, all 45 pairs per paper, or 9,000 pairs overall, contribute to the unbiased estimate $\hat V$, while the concentration bound uses $M=200\times5=1,000$. These quantities serve variance estimation and confidence control, respectively; the overlapping 9,000 pairs are not treated as independent kernel values.

\subsection{The Two 95\% Levels}

Choose $k(U_\alpha)=\lceil U_\alpha/(\beta\epsilon^2)\rceil$ from the pilot data. For separately collected, independent future scores,

\[
\Pr_{\mathrm{pilot}}\{A(k(U_\alpha),\epsilon)\ge1-\beta\}\ge1-\alpha.
\]

The main paper uses $\alpha=\beta=0.05$. The first 95\% is the confidence level for the variance upper bound constructed from the pilot data; the second is the overall success probability of future mean-score estimation. The budget $k$ counts new independent reviews and excludes the existing pilot reviews.

Averaging over both pilot-data and future-score randomness gives a joint success lower bound of $(1-\alpha)(1-\beta)$. This equals 0.9025 when $\alpha=\beta=0.05$. To require a joint lower bound of at least 0.95 in advance, one can choose $\alpha=0.01,\beta=0.04$, giving $0.99\times0.96=0.9504$. Section D.5 reports the corresponding budgets.

\subsection{Simultaneous Control Across Models and Different Tolerances}

Assigning failure probability $\alpha/3$ to each model's variance bound gives simultaneous coverage of at least $1-\alpha$ for all three bounds by Bonferroni's inequality. Each model's overall future-score success probability remains $1-\beta$.

\begin{table*}[tp]\centering
\small
\setlength{\tabcolsep}{4pt}
\renewcommand{\arraystretch}{1.12}
\renewcommand{\thetable}{S16}
\begin{tabular}{@{}>{\raggedright\arraybackslash}p{55.943pt}>{\raggedright\arraybackslash}p{212.553pt}>{\raggedleft\arraybackslash}p{43.433pt}>{\raggedleft\arraybackslash}p{24.553pt}>{\raggedleft\arraybackslash}p{27.753pt}>{\raggedleft\arraybackslash}p{22.753pt}>{\raggedleft\arraybackslash}p{20.253pt}@{}}
\toprule
{\textbf{\hspace{0pt}Model}} & {\textbf{\hspace{0pt}Confidence/\allowbreak{}success convention}} & {\textbf{\hspace{0pt}Variance upper bound}} & {\textbf{\hspace{0pt}$\epsilon$=.1}} & {\textbf{\hspace{0pt}$\epsilon$=.25}} & {\textbf{\hspace{0pt}$\epsilon$=.5}} & {\textbf{\hspace{0pt}$\epsilon$=1}} \\
\midrule
{\hspace{0pt}GPT-5.6-Sol} & {\hspace{0pt}Per-model 95\% confidence /\allowbreak{} 95\% success} & {\hspace{0pt}0.411370} & {\hspace{0pt}823} & {\hspace{0pt}132} & {\hspace{0pt}33} & {\hspace{0pt}9} \\
{\hspace{0pt}GPT-5.6-Sol} & {\hspace{0pt}Simultaneous 95\% confidence /\allowbreak{} 95\% success per model} & {\hspace{0pt}0.475736} & {\hspace{0pt}952} & {\hspace{0pt}153} & {\hspace{0pt}39} & {\hspace{0pt}10} \\
{\hspace{0pt}GPT-5.6-Sol} & {\hspace{0pt}Joint pilot/\allowbreak{}future success $\geq$95\%} & {\hspace{0pt}0.504356} & {\hspace{0pt}1261} & {\hspace{0pt}202} & {\hspace{0pt}51} & {\hspace{0pt}13} \\
\addlinespace[3pt]
{\hspace{0pt}GPT-6-Luna} & {\hspace{0pt}Per-model 95\% confidence /\allowbreak{} 95\% success} & {\hspace{0pt}0.425662} & {\hspace{0pt}852} & {\hspace{0pt}137} & {\hspace{0pt}35} & {\hspace{0pt}9} \\
{\hspace{0pt}GPT-6-Luna} & {\hspace{0pt}Simultaneous 95\% confidence /\allowbreak{} 95\% success per model} & {\hspace{0pt}0.490875} & {\hspace{0pt}982} & {\hspace{0pt}158} & {\hspace{0pt}40} & {\hspace{0pt}10} \\
{\hspace{0pt}GPT-6-Luna} & {\hspace{0pt}Joint pilot/\allowbreak{}future success $\geq$95\%} & {\hspace{0pt}0.519842} & {\hspace{0pt}1300} & {\hspace{0pt}208} & {\hspace{0pt}52} & {\hspace{0pt}13} \\
\addlinespace[3pt]
{\hspace{0pt}GPT-6-Sol} & {\hspace{0pt}Per-model 95\% confidence /\allowbreak{} 95\% success} & {\hspace{0pt}0.297823} & {\hspace{0pt}596} & {\hspace{0pt}96} & {\hspace{0pt}24} & {\hspace{0pt}6} \\
{\hspace{0pt}GPT-6-Sol} & {\hspace{0pt}Simultaneous 95\% confidence /\allowbreak{} 95\% success per model} & {\hspace{0pt}0.354954} & {\hspace{0pt}710} & {\hspace{0pt}114} & {\hspace{0pt}29} & {\hspace{0pt}8} \\
{\hspace{0pt}GPT-6-Sol} & {\hspace{0pt}Joint pilot/\allowbreak{}future success $\geq$95\%} & {\hspace{0pt}0.380626} & {\hspace{0pt}952} & {\hspace{0pt}153} & {\hspace{0pt}39} & {\hspace{0pt}10} \\
\bottomrule
\end{tabular}
\caption{Sufficient budgets with confidence control from ten pilot reviews. Per-model control uses $\alpha$=.05 and $\beta$=.05; simultaneous control across three models uses $\alpha$=.05/\allowbreak{}3 per model and $\beta$=.05; joint control uses $\alpha$=.01 and $\beta$=.04.}
\end{table*}

\begin{table*}[tp]\centering
\normalsize
\setlength{\tabcolsep}{4pt}
\renewcommand{\arraystretch}{1.12}
\renewcommand{\thetable}{S17}
\begin{tabular}{@{}>{\raggedright\arraybackslash}p{94.184pt}>{\raggedleft\arraybackslash}p{73.097pt}>{\raggedleft\arraybackslash}p{84.702pt}>{\raggedleft\arraybackslash}p{79.840pt}>{\raggedleft\arraybackslash}p{91.423pt}@{}}
\toprule
{\textbf{\hspace{0pt}Model}} & {\textbf{\hspace{0pt}Success target}} & {\textbf{\hspace{0pt}Tolerance}} & {\textbf{\hspace{0pt}Planning}} & {\textbf{\hspace{0pt}Certified (per-model 95\% confidence)}} \\
\midrule
{\hspace{0pt}GPT-5.6-Sol} & {\hspace{0pt}90\%} & {\hspace{0pt}0.25} & {\hspace{0pt}23} & {\hspace{0pt}66} \\
{\hspace{0pt}GPT-5.6-Sol} & {\hspace{0pt}90\%} & {\hspace{0pt}0.5} & {\hspace{0pt}6} & {\hspace{0pt}17} \\
{\hspace{0pt}GPT-5.6-Sol} & {\hspace{0pt}95\%} & {\hspace{0pt}0.25} & {\hspace{0pt}45} & {\hspace{0pt}132} \\
{\hspace{0pt}GPT-5.6-Sol} & {\hspace{0pt}95\%} & {\hspace{0pt}0.5} & {\hspace{0pt}12} & {\hspace{0pt}33} \\
{\hspace{0pt}GPT-5.6-Sol} & {\hspace{0pt}99\%} & {\hspace{0pt}0.25} & {\hspace{0pt}225} & {\hspace{0pt}659} \\
{\hspace{0pt}GPT-5.6-Sol} & {\hspace{0pt}99\%} & {\hspace{0pt}0.5} & {\hspace{0pt}57} & {\hspace{0pt}165} \\
\addlinespace[3pt]
{\hspace{0pt}GPT-6-Luna} & {\hspace{0pt}90\%} & {\hspace{0pt}0.25} & {\hspace{0pt}24} & {\hspace{0pt}69} \\
{\hspace{0pt}GPT-6-Luna} & {\hspace{0pt}90\%} & {\hspace{0pt}0.5} & {\hspace{0pt}6} & {\hspace{0pt}18} \\
{\hspace{0pt}GPT-6-Luna} & {\hspace{0pt}95\%} & {\hspace{0pt}0.25} & {\hspace{0pt}48} & {\hspace{0pt}137} \\
{\hspace{0pt}GPT-6-Luna} & {\hspace{0pt}95\%} & {\hspace{0pt}0.5} & {\hspace{0pt}12} & {\hspace{0pt}35} \\
{\hspace{0pt}GPT-6-Luna} & {\hspace{0pt}99\%} & {\hspace{0pt}0.25} & {\hspace{0pt}239} & {\hspace{0pt}682} \\
{\hspace{0pt}GPT-6-Luna} & {\hspace{0pt}99\%} & {\hspace{0pt}0.5} & {\hspace{0pt}60} & {\hspace{0pt}171} \\
\addlinespace[3pt]
{\hspace{0pt}GPT-6-Sol} & {\hspace{0pt}90\%} & {\hspace{0pt}0.25} & {\hspace{0pt}12} & {\hspace{0pt}48} \\
{\hspace{0pt}GPT-6-Sol} & {\hspace{0pt}90\%} & {\hspace{0pt}0.5} & {\hspace{0pt}3} & {\hspace{0pt}12} \\
{\hspace{0pt}GPT-6-Sol} & {\hspace{0pt}95\%} & {\hspace{0pt}0.25} & {\hspace{0pt}24} & {\hspace{0pt}96} \\
{\hspace{0pt}GPT-6-Sol} & {\hspace{0pt}95\%} & {\hspace{0pt}0.5} & {\hspace{0pt}6} & {\hspace{0pt}24} \\
{\hspace{0pt}GPT-6-Sol} & {\hspace{0pt}99\%} & {\hspace{0pt}0.25} & {\hspace{0pt}119} & {\hspace{0pt}477} \\
{\hspace{0pt}GPT-6-Sol} & {\hspace{0pt}99\%} & {\hspace{0pt}0.5} & {\hspace{0pt}30} & {\hspace{0pt}120} \\
\bottomrule
\end{tabular}
\caption{Planning and Certified budgets for different future success targets. Certified uses a per-model 95\% variance upper bound throughout, while the success target varies separately.}
\end{table*}

Halving the tolerance approximately quadruples the variance-based budget, and a higher success target also increases the budget. All tables round budgets upward to integer review counts.

\subsection{Five Versus Ten Pilot Reviews}

\begin{table*}[tp]\centering
\normalsize
\setlength{\tabcolsep}{4pt}
\renewcommand{\arraystretch}{1.12}
\renewcommand{\thetable}{S18}
\begin{tabular}{@{}>{\raggedright\arraybackslash}p{84.589pt}>{\raggedleft\arraybackslash}p{50.060pt}>{\raggedleft\arraybackslash}p{70.828pt}>{\raggedleft\arraybackslash}p{69.310pt}>{\raggedleft\arraybackslash}p{70.245pt}>{\raggedleft\arraybackslash}p{70.212pt}@{}}
\toprule
{\textbf{\hspace{0pt}Model}} & {\textbf{\hspace{0pt}Pilot n}} & {\textbf{\hspace{0pt}Variance}} & {\textbf{\hspace{0pt}95\% variance upper bound}} & {\textbf{\hspace{0pt}Planning $\epsilon$=.5}} & {\textbf{\hspace{0pt}Certified $\epsilon$=.5}} \\
\midrule
{\hspace{0pt}GPT-5.6-Sol} & {\hspace{0pt}5} & {\hspace{0pt}0.139000} & {\hspace{0pt}0.654330} & {\hspace{0pt}12} & {\hspace{0pt}53} \\
{\hspace{0pt}GPT-5.6-Sol} & {\hspace{0pt}10} & {\hspace{0pt}0.140056} & {\hspace{0pt}0.411370} & {\hspace{0pt}12} & {\hspace{0pt}33} \\
\addlinespace[3pt]
{\hspace{0pt}GPT-6-Luna} & {\hspace{0pt}5} & {\hspace{0pt}0.152000} & {\hspace{0pt}0.679444} & {\hspace{0pt}13} & {\hspace{0pt}55} \\
{\hspace{0pt}GPT-6-Luna} & {\hspace{0pt}10} & {\hspace{0pt}0.148889} & {\hspace{0pt}0.425662} & {\hspace{0pt}12} & {\hspace{0pt}35} \\
\addlinespace[3pt]
{\hspace{0pt}GPT-6-Sol} & {\hspace{0pt}5} & {\hspace{0pt}0.082000} & {\hspace{0pt}0.536812} & {\hspace{0pt}7} & {\hspace{0pt}43} \\
{\hspace{0pt}GPT-6-Sol} & {\hspace{0pt}10} & {\hspace{0pt}0.074056} & {\hspace{0pt}0.297823} & {\hspace{0pt}6} & {\hspace{0pt}24} \\
\bottomrule
\end{tabular}
\caption{Five versus ten pilot reviews under the same configuration. The five-review pilot uses the first five reviews. The future success target is 95\%, with a tolerance of 0.5 points.}
\end{table*}

Ten pilot reviews increase the number of independent kernel values available to the bound from 400 to 1,000. Observed variances change little, but the upper bounds tighten: Certified budgets at a half-point tolerance decrease from 53, 55, and 43 to 33, 35, and 24.

\subsection{A Reference Bound Using Only the Score Range}

Without pilot variance estimates, the paper-specific Hoeffding bound gives

\[
\Pr(|\bar Y_{i,k}-\mu_i|\ge\epsilon)\le2\exp\left(-\frac{2k\epsilon^2}{81}\right).
\]

Thus $k\ge\lceil81\log(2/\beta)/(2\epsilon^2)\rceil$ suffices to control each paper's failure probability and hence the average failure probability. With $\beta=.05$, tolerances of 0.25 and 0.5 points require 2,391 and 598 reviews, respectively. Our primary analysis uses pilot variances and their upper bounds to exploit the observed dispersion of scores.

\section{Score Granularity and Distribution Estimation}

\subsection{Integer Versus Forced-Decimal Scoring}

We compare the two protocols within papers, with five scores per protocol: 1,000 observations per model and protocol. Integer scores are exactly the first five from the ten-review original-manuscript experiment. The table also reports year-specific results and both acceptance-label conventions.

\begin{table*}[tp]\centering
\small
\setlength{\tabcolsep}{4pt}
\renewcommand{\arraystretch}{1.12}
\renewcommand{\thetable}{S19}
\begin{tabular}{@{}>{\raggedright\arraybackslash}p{77.349pt}>{\raggedleft\arraybackslash}p{47.349pt}>{\raggedright\arraybackslash}p{88.179pt}>{\raggedleft\arraybackslash}p{53.749pt}>{\raggedleft\arraybackslash}p{84.589pt}>{\raggedleft\arraybackslash}p{64.029pt}@{}}
\toprule
{\textbf{\hspace{0pt}Model}} & {\textbf{\hspace{0pt}Year}} & {\textbf{\hspace{0pt}Protocol}} & {\textbf{\hspace{0pt}Mean}} & {\textbf{\hspace{0pt}Within-paper SD}} & {\textbf{\hspace{0pt}Identical 5/\allowbreak{}5}} \\
\midrule
{\hspace{0pt}GPT-6-Luna} & {\hspace{0pt}All} & {\hspace{0pt}Integer} & {\hspace{0pt}3.4590} & {\hspace{0pt}0.3899} & {\hspace{0pt}45.5\%} \\
{\hspace{0pt}GPT-6-Luna} & {\hspace{0pt}All} & {\hspace{0pt}Forced decimal} & {\hspace{0pt}4.1524} & {\hspace{0pt}0.3290} & {\hspace{0pt}3.5\%} \\
{\hspace{0pt}GPT-6-Luna} & {\hspace{0pt}2025} & {\hspace{0pt}Integer} & {\hspace{0pt}3.8080} & {\hspace{0pt}0.3975} & {\hspace{0pt}41.0\%} \\
{\hspace{0pt}GPT-6-Luna} & {\hspace{0pt}2025} & {\hspace{0pt}Forced decimal} & {\hspace{0pt}4.4850} & {\hspace{0pt}0.3184} & {\hspace{0pt}5.0\%} \\
{\hspace{0pt}GPT-6-Luna} & {\hspace{0pt}2026} & {\hspace{0pt}Integer} & {\hspace{0pt}3.1100} & {\hspace{0pt}0.3821} & {\hspace{0pt}50.0\%} \\
{\hspace{0pt}GPT-6-Luna} & {\hspace{0pt}2026} & {\hspace{0pt}Forced decimal} & {\hspace{0pt}3.8198} & {\hspace{0pt}0.3392} & {\hspace{0pt}2.0\%} \\
\addlinespace[3pt]
{\hspace{0pt}GPT-6-Sol} & {\hspace{0pt}All} & {\hspace{0pt}Integer} & {\hspace{0pt}4.6530} & {\hspace{0pt}0.2864} & {\hspace{0pt}66.0\%} \\
{\hspace{0pt}GPT-6-Sol} & {\hspace{0pt}All} & {\hspace{0pt}Forced decimal} & {\hspace{0pt}4.9190} & {\hspace{0pt}0.2485} & {\hspace{0pt}17.0\%} \\
{\hspace{0pt}GPT-6-Sol} & {\hspace{0pt}2025} & {\hspace{0pt}Integer} & {\hspace{0pt}4.8380} & {\hspace{0pt}0.2793} & {\hspace{0pt}69.0\%} \\
{\hspace{0pt}GPT-6-Sol} & {\hspace{0pt}2025} & {\hspace{0pt}Forced decimal} & {\hspace{0pt}5.0720} & {\hspace{0pt}0.2587} & {\hspace{0pt}16.0\%} \\
{\hspace{0pt}GPT-6-Sol} & {\hspace{0pt}2026} & {\hspace{0pt}Integer} & {\hspace{0pt}4.4680} & {\hspace{0pt}0.2933} & {\hspace{0pt}63.0\%} \\
{\hspace{0pt}GPT-6-Sol} & {\hspace{0pt}2026} & {\hspace{0pt}Forced decimal} & {\hspace{0pt}4.7660} & {\hspace{0pt}0.2380} & {\hspace{0pt}18.0\%} \\
\bottomrule
\end{tabular}
\par\medskip
\begin{tabular}{@{}>{\raggedright\arraybackslash}p{62.433pt}>{\raggedleft\arraybackslash}p{32.433pt}>{\raggedright\arraybackslash}p{73.263pt}>{\raggedleft\arraybackslash}p{55.223pt}>{\raggedleft\arraybackslash}p{38.833pt}>{\raggedleft\arraybackslash}p{40.223pt}>{\raggedleft\arraybackslash}p{38.833pt}>{\raggedleft\arraybackslash}p{58.003pt}@{}}
\toprule
{\textbf{\hspace{0pt}Model}} & {\textbf{\hspace{0pt}Year}} & {\textbf{\hspace{0pt}Protocol}} & {\textbf{\hspace{0pt}Spearman}} & {\textbf{\hspace{0pt}MAE}} & {\textbf{\hspace{0pt}RMSE}} & {\textbf{\hspace{0pt}AUC: all}} & {\textbf{\hspace{0pt}AUC: excl. withdrawn}} \\
\midrule
{\hspace{0pt}GPT-6-Luna} & {\hspace{0pt}All} & {\hspace{0pt}Integer} & {\hspace{0pt}0.3367} & {\hspace{0pt}1.4687} & {\hspace{0pt}1.8500} & {\hspace{0pt}0.6427} & {\hspace{0pt}0.6399} \\
{\hspace{0pt}GPT-6-Luna} & {\hspace{0pt}All} & {\hspace{0pt}Forced decimal} & {\hspace{0pt}0.3687} & {\hspace{0pt}1.1620} & {\hspace{0pt}1.4507} & {\hspace{0pt}0.6675} & {\hspace{0pt}0.6668} \\
{\hspace{0pt}GPT-6-Luna} & {\hspace{0pt}2025} & {\hspace{0pt}Integer} & {\hspace{0pt}0.3974} & {\hspace{0pt}1.5417} & {\hspace{0pt}1.9033} & {\hspace{0pt}0.6795} & {\hspace{0pt}0.6504} \\
{\hspace{0pt}GPT-6-Luna} & {\hspace{0pt}2025} & {\hspace{0pt}Forced decimal} & {\hspace{0pt}0.4008} & {\hspace{0pt}1.2029} & {\hspace{0pt}1.4874} & {\hspace{0pt}0.6940} & {\hspace{0pt}0.6612} \\
{\hspace{0pt}GPT-6-Luna} & {\hspace{0pt}2026} & {\hspace{0pt}Integer} & {\hspace{0pt}0.1368} & {\hspace{0pt}1.3957} & {\hspace{0pt}1.7951} & {\hspace{0pt}0.5850} & {\hspace{0pt}0.6060} \\
{\hspace{0pt}GPT-6-Luna} & {\hspace{0pt}2026} & {\hspace{0pt}Forced decimal} & {\hspace{0pt}0.2120} & {\hspace{0pt}1.1212} & {\hspace{0pt}1.4131} & {\hspace{0pt}0.6190} & {\hspace{0pt}0.6485} \\
\addlinespace[3pt]
{\hspace{0pt}GPT-6-Sol} & {\hspace{0pt}All} & {\hspace{0pt}Integer} & {\hspace{0pt}0.4096} & {\hspace{0pt}1.0207} & {\hspace{0pt}1.2612} & {\hspace{0pt}0.7069} & {\hspace{0pt}0.7028} \\
{\hspace{0pt}GPT-6-Sol} & {\hspace{0pt}All} & {\hspace{0pt}Forced decimal} & {\hspace{0pt}0.3998} & {\hspace{0pt}1.0195} & {\hspace{0pt}1.2572} & {\hspace{0pt}0.7201} & {\hspace{0pt}0.7250} \\
{\hspace{0pt}GPT-6-Sol} & {\hspace{0pt}2025} & {\hspace{0pt}Integer} & {\hspace{0pt}0.5407} & {\hspace{0pt}0.9503} & {\hspace{0pt}1.1774} & {\hspace{0pt}0.7743} & {\hspace{0pt}0.7483} \\
{\hspace{0pt}GPT-6-Sol} & {\hspace{0pt}2025} & {\hspace{0pt}Forced decimal} & {\hspace{0pt}0.5221} & {\hspace{0pt}0.8911} & {\hspace{0pt}1.1103} & {\hspace{0pt}0.7713} & {\hspace{0pt}0.7449} \\
{\hspace{0pt}GPT-6-Sol} & {\hspace{0pt}2026} & {\hspace{0pt}Integer} & {\hspace{0pt}0.2290} & {\hspace{0pt}1.0910} & {\hspace{0pt}1.3397} & {\hspace{0pt}0.6136} & {\hspace{0pt}0.6348} \\
{\hspace{0pt}GPT-6-Sol} & {\hspace{0pt}2026} & {\hspace{0pt}Forced decimal} & {\hspace{0pt}0.2492} & {\hspace{0pt}1.1478} & {\hspace{0pt}1.3887} & {\hspace{0pt}0.6525} & {\hspace{0pt}0.7013} \\
\bottomrule
\end{tabular}
\caption{Complete metrics for integer and forced-decimal protocols. Means, human-score associations, and errors use all papers in the specified year; AUCs use the full and withdrawal-excluded samples separately.}
\end{table*}

\begin{table*}[tp]\centering
\normalsize
\setlength{\tabcolsep}{4pt}
\renewcommand{\arraystretch}{1.12}
\renewcommand{\thetable}{S20}
\begin{tabular}{@{}>{\raggedright\arraybackslash}p{90.582pt}>{\raggedleft\arraybackslash}p{57.582pt}>{\raggedleft\arraybackslash}p{142.810pt}>{\raggedleft\arraybackslash}p{140.269pt}@{}}
\toprule
{\textbf{\hspace{0pt}Model}} & {\textbf{\hspace{0pt}Year}} & {\textbf{\hspace{0pt}Mean change and 95\% CI}} & {\textbf{\hspace{0pt}MAE change and 95\% CI}} \\
\midrule
{\hspace{0pt}GPT-6-Luna} & {\hspace{0pt}All} & {\hspace{0pt}+0.693 [+0.649, +0.739]} & {\hspace{0pt}-0.307 [-0.389, -0.223]} \\
{\hspace{0pt}GPT-6-Luna} & {\hspace{0pt}2025} & {\hspace{0pt}+0.677 [+0.613, +0.744]} & {\hspace{0pt}-0.339 [-0.453, -0.221]} \\
{\hspace{0pt}GPT-6-Luna} & {\hspace{0pt}2026} & {\hspace{0pt}+0.710 [+0.648, +0.771]} & {\hspace{0pt}-0.274 [-0.400, -0.145]} \\
\addlinespace[3pt]
{\hspace{0pt}GPT-6-Sol} & {\hspace{0pt}All} & {\hspace{0pt}+0.266 [+0.233, +0.298]} & {\hspace{0pt}-0.001 [-0.050, +0.044]} \\
{\hspace{0pt}GPT-6-Sol} & {\hspace{0pt}2025} & {\hspace{0pt}+0.234 [+0.187, +0.279]} & {\hspace{0pt}-0.059 [-0.119, +0.001]} \\
{\hspace{0pt}GPT-6-Sol} & {\hspace{0pt}2026} & {\hspace{0pt}+0.298 [+0.250, +0.345]} & {\hspace{0pt}+0.057 [-0.011, +0.125]} \\
\bottomrule
\end{tabular}
\caption{Changes in mean score and MAE, decimal minus integer. Overall rows retain the main paper's original intervals; year-specific rows use paired paper-level bootstrap resampling.}
\end{table*}

The forced-decimal protocol raises Luna's mean score by 0.693 points and Sol's by 0.266. Their within-paper SDs decrease from 0.390 to 0.329 and from 0.286 to 0.249, respectively. Luna's MAE decreases in both years. Sol's MAE changes in opposite directions across years, with both intervals including zero.

Output granularity affects exact agreement and score range under the integer protocol. If scores switch only between adjacent integers $s$ and $s+1$, with probabilities $1-\pi$ and $\pi$, the variance is $\pi(1-\pi)$ and the probability of disagreement between two reviews is $2\pi(1-\pi)$. This two-point model shows how switching between adjacent levels can produce frequent disagreement within a small range. The decimal experiment directly measures the variance and score-location changes caused by a protocol change, without assuming that lower variance implies a more accurate estimate of an unchanged scoring target.

\subsection{Construction of Quick and Standard}

The target is the score distribution $p_i(s)$ under the fixed GPT-5.6-Sol medium integer configuration, for $s\in\{1,\ldots,10\}$. An auxiliary call uses a nonreasoning configuration and extracts the visible probabilities of valid candidate score tokens. Restricting candidates to 1–10 and renormalizing yields $r_i(s)$. The analysis does not assign probability mass to candidates absent from the response.

Quick first adjusts probability concentration:

\[
\widetilde r_i(s;T)=\frac{\max(r_i(s),10^{-9})^{1/T}}{\sum_{u\in\mathcal S_i}\max(r_i(u),10^{-9})^{1/T}},
\]

Here $\mathcal S_i$ is the valid candidate set returned by the call. It then maps score locations to $x_s=\operatorname{clip}(a+bs,1,10)$ and linearly distributes the mass at each location between adjacent integers. Integer locations receive all their mass; otherwise, the weights at $\lfloor x_s\rfloor$ and $\lceil x_s\rceil$ are $\lceil x_s\rceil-x_s$ and $x_s-\lfloor x_s\rfloor$, respectively. Denote the resulting distribution by $q_{Q,i}$.

Standard combines Quick with two reasoning scores from the target configuration:

\[
q_{S,i}=w\hat p_{i,2}+(1-w)q_{Q,i}.
\]

A Quick prediction uses one auxiliary call. Standard uses that same auxiliary call and two reasoning reviews.

\subsection{Parameter Search and Train–Test Splits}

The 198 papers are sorted by anonymous ID, and 30 random 99/\allowbreak{}99 train–test splits are generated without additional stratification by year. The search grid is

\[
\begin{gathered}T\in\{1,2,3,5,8,12\},\\ a\in\{-2.0,-1.8,\ldots,0.4\},\\ b\in\{0.9,1.0,1.1\},\end{gathered}
\]

\[
w\in\{0.3,0.4,\ldots,1.0\}.
\]

Quick has 234 parameter combinations. First, $(T,a,b)$ is selected using Brier loss against five target scores per training paper. Next, all ten partitions of five scores into two inputs and three targets are enumerated, and the mean training loss selects one of eight candidate weights. Test papers do not participate in parameter selection, and Standard's input and target scores do not overlap.

The parameter combinations selected across the 30 splits are as follows.

\begin{table*}[tp]\centering
\normalsize
\setlength{\tabcolsep}{4pt}
\renewcommand{\arraystretch}{1.12}
\renewcommand{\thetable}{S21}
\begin{tabular}{@{}>{\raggedleft\arraybackslash}p{79.208pt}>{\raggedleft\arraybackslash}p{85.621pt}>{\raggedleft\arraybackslash}p{81.958pt}>{\raggedleft\arraybackslash}p{81.958pt}>{\raggedleft\arraybackslash}p{94.498pt}@{}}
\toprule
{\textbf{\hspace{0pt}T}} & {\textbf{\hspace{0pt}a}} & {\textbf{\hspace{0pt}b}} & {\textbf{\hspace{0pt}w}} & {\textbf{\hspace{0pt}Splits}} \\
\midrule
{\hspace{0pt}12} & {\hspace{0pt}0.2} & {\hspace{0pt}0.9} & {\hspace{0pt}0.7} & {\hspace{0pt}7} \\
\addlinespace[3pt]
{\hspace{0pt}5} & {\hspace{0pt}0.2} & {\hspace{0pt}0.9} & {\hspace{0pt}0.7} & {\hspace{0pt}5} \\
\addlinespace[3pt]
{\hspace{0pt}8} & {\hspace{0pt}-0.2} & {\hspace{0pt}1.0} & {\hspace{0pt}0.7} & {\hspace{0pt}3} \\
{\hspace{0pt}8} & {\hspace{0pt}0.2} & {\hspace{0pt}0.9} & {\hspace{0pt}0.6} & {\hspace{0pt}3} \\
{\hspace{0pt}8} & {\hspace{0pt}0.2} & {\hspace{0pt}0.9} & {\hspace{0pt}0.7} & {\hspace{0pt}2} \\
\addlinespace[3pt]
{\hspace{0pt}12} & {\hspace{0pt}0.2} & {\hspace{0pt}0.9} & {\hspace{0pt}0.6} & {\hspace{0pt}2} \\
\addlinespace[3pt]
{\hspace{0pt}2} & {\hspace{0pt}-0.4} & {\hspace{0pt}1.0} & {\hspace{0pt}0.6} & {\hspace{0pt}1} \\
\addlinespace[3pt]
{\hspace{0pt}3} & {\hspace{0pt}-0.2} & {\hspace{0pt}1.0} & {\hspace{0pt}0.6} & {\hspace{0pt}1} \\
\addlinespace[3pt]
{\hspace{0pt}12} & {\hspace{0pt}0.4} & {\hspace{0pt}0.9} & {\hspace{0pt}0.6} & {\hspace{0pt}1} \\
{\hspace{0pt}12} & {\hspace{0pt}-0.2} & {\hspace{0pt}1.0} & {\hspace{0pt}0.7} & {\hspace{0pt}1} \\
\addlinespace[3pt]
{\hspace{0pt}5} & {\hspace{0pt}-0.2} & {\hspace{0pt}1.0} & {\hspace{0pt}0.7} & {\hspace{0pt}1} \\
{\hspace{0pt}5} & {\hspace{0pt}-0.2} & {\hspace{0pt}1.0} & {\hspace{0pt}0.6} & {\hspace{0pt}1} \\
\addlinespace[3pt]
{\hspace{0pt}3} & {\hspace{0pt}0.2} & {\hspace{0pt}0.9} & {\hspace{0pt}0.6} & {\hspace{0pt}1} \\
\addlinespace[3pt]
{\hspace{0pt}5} & {\hspace{0pt}0.2} & {\hspace{0pt}0.9} & {\hspace{0pt}0.6} & {\hspace{0pt}1} \\
\bottomrule
\end{tabular}
\caption{Selection frequencies of Quick parameters and Standard weights. Each row is a combination jointly selected in a split; frequencies sum to 30.}
\end{table*}

At test time, Quick consistently uses the first valid auxiliary output and is evaluated against five independent target scores. Standard enumerates ten partitions per paper, first averaging over partitions, then over that paper's test-set appearances, and finally equally across the 198 papers. Each paper appears in the test set 8–23 times.

\subsection{Squared Distribution Risk with Independent Targets}

For a predicted distribution $q$ and independent target scores $Z_1,\ldots,Z_m$, define

\[
\begin{aligned}\widehat R(q)={}&\sum_s q(s)^2-\frac2m\sum_{j=1}^m q(Z_j)\\ &+\frac{1}{m(m-1)}\sum_{j\ne\ell}\mathbf1[Z_j=Z_\ell].\end{aligned}
\]

Conditional on $q$, the expectation of the second term is $-2\sum_s q(s)p_i(s)$ and that of the final term is $\sum_s p_i(s)^2$. Therefore,

\[
\mathbb E[\widehat R(q)\mid q]=\sum_s(q(s)-p_i(s))^2.
\]

The final term estimates the squared probability sum through collisions between independent target scores, removing variation due to a finite target sample. Individual paper-level risk estimates may be negative; these values are retained when averaging.

For a frequency estimator based on $n$ direct scores, expected risk is

\[
\mathbb E\|\hat p_{i,n}-p_i\|_2^2=\frac{1-\sum_s p_i(s)^2}{n}.
\]

Direct-sampling baselines estimate the numerator from the collision statistic of the five existing scores and divide by $n$. Risks for 1, 2, 3, and 5 reviews are estimates of expected risk under this sampling law, without treating the five-score histogram as a known target distribution.

For an auxiliary prediction conditionally independent of two reasoning scores, and a frequency estimator unbiased for $p_i$, Standard's risk decomposes as

\[
\begin{aligned}&\mathbb E\|q_{S,i}-p_i\|_2^2\\ &=\frac{w^2}{2}\bigl(1-\|p_i\|_2^2\bigr)+(1-w)^2\mathbb E\|q_{Q,i}-p_i\|_2^2.\end{aligned}
\]

The cross term vanishes because the reasoning-frequency error has mean zero and is conditionally independent of the auxiliary prediction. Weights are selected on training papers, and the combined predictor's risk is evaluated on test papers.

\subsection{Pooled and Year-Specific Risks}

\begin{table*}[tp]\centering
\normalsize
\setlength{\tabcolsep}{4pt}
\renewcommand{\arraystretch}{1.12}
\renewcommand{\thetable}{S22}
\begin{tabular}{@{}>{\raggedleft\arraybackslash}p{76.868pt}>{\raggedleft\arraybackslash}p{86.042pt}>{\raggedright\arraybackslash}p{108.350pt}>{\raggedleft\arraybackslash}p{159.984pt}@{}}
\toprule
{\textbf{\hspace{0pt}Year}} & {\textbf{\hspace{0pt}Papers}} & {\textbf{\hspace{0pt}Method}} & {\textbf{\hspace{0pt}Risk and 95\% paper-bootstrap CI}} \\
\midrule
{\hspace{0pt}All} & {\hspace{0pt}198} & {\hspace{0pt}Quick} & {\hspace{0pt}0.2427 [0.2029, 0.2860]} \\
{\hspace{0pt}All} & {\hspace{0pt}198} & {\hspace{0pt}Standard} & {\hspace{0pt}0.0829 [0.0750, 0.0911]} \\
{\hspace{0pt}All} & {\hspace{0pt}198} & {\hspace{0pt}1 Reasoning} & {\hspace{0pt}0.2434 [0.2051, 0.2828]} \\
{\hspace{0pt}All} & {\hspace{0pt}198} & {\hspace{0pt}2 Reasoning} & {\hspace{0pt}0.1217 [0.1025, 0.1414]} \\
{\hspace{0pt}All} & {\hspace{0pt}198} & {\hspace{0pt}3 Reasoning} & {\hspace{0pt}0.0811 [0.0684, 0.0943]} \\
{\hspace{0pt}All} & {\hspace{0pt}198} & {\hspace{0pt}5 Reasoning} & {\hspace{0pt}0.0487 [0.0410, 0.0566]} \\
\addlinespace[3pt]
{\hspace{0pt}2025} & {\hspace{0pt}99} & {\hspace{0pt}Quick} & {\hspace{0pt}0.2141 [0.1634, 0.2709]} \\
{\hspace{0pt}2025} & {\hspace{0pt}99} & {\hspace{0pt}Standard} & {\hspace{0pt}0.0932 [0.0821, 0.1041]} \\
{\hspace{0pt}2025} & {\hspace{0pt}99} & {\hspace{0pt}1 Reasoning} & {\hspace{0pt}0.3040 [0.2495, 0.3586]} \\
{\hspace{0pt}2025} & {\hspace{0pt}99} & {\hspace{0pt}2 Reasoning} & {\hspace{0pt}0.1520 [0.1247, 0.1793]} \\
{\hspace{0pt}2025} & {\hspace{0pt}99} & {\hspace{0pt}3 Reasoning} & {\hspace{0pt}0.1013 [0.0832, 0.1195]} \\
{\hspace{0pt}2025} & {\hspace{0pt}99} & {\hspace{0pt}5 Reasoning} & {\hspace{0pt}0.0608 [0.0499, 0.0717]} \\
\addlinespace[3pt]
{\hspace{0pt}2026} & {\hspace{0pt}99} & {\hspace{0pt}Quick} & {\hspace{0pt}0.2713 [0.2129, 0.3350]} \\
{\hspace{0pt}2026} & {\hspace{0pt}99} & {\hspace{0pt}Standard} & {\hspace{0pt}0.0727 [0.0620, 0.0839]} \\
{\hspace{0pt}2026} & {\hspace{0pt}99} & {\hspace{0pt}1 Reasoning} & {\hspace{0pt}0.1828 [0.1333, 0.2354]} \\
{\hspace{0pt}2026} & {\hspace{0pt}99} & {\hspace{0pt}2 Reasoning} & {\hspace{0pt}0.0914 [0.0667, 0.1177]} \\
{\hspace{0pt}2026} & {\hspace{0pt}99} & {\hspace{0pt}3 Reasoning} & {\hspace{0pt}0.0609 [0.0444, 0.0785]} \\
{\hspace{0pt}2026} & {\hspace{0pt}99} & {\hspace{0pt}5 Reasoning} & {\hspace{0pt}0.0366 [0.0267, 0.0471]} \\
\bottomrule
\end{tabular}
\caption{Risks from the same held-out analysis, aggregated by year. Parameters remain those fitted on the original mixed-year training sets; test results are grouped by paper year without year-specific refitting.}
\end{table*}

Overall Standard risk is 0.0829. Year-specific risks are 0.0932 in 2025 and 0.0727 in 2026, with 99 papers per year; two direct reasoning reviews yield 0.1520 and 0.0914. All intervals hold the original parameters and splits fixed while resampling papers, preserving the same within-paper aggregation order.

\subsection{Bootstrap with Parameter Refitting}

For each paper-level refit bootstrap, we draw 198 papers with replacement from the 198-paper set, record each paper's multiplicity, and use it as a weight within the original 30 train–test splits. The training side searches all 234 Quick parameter combinations and eight Standard weights again. The test side continues to exclude training papers for that split. Refit results retain the original within-paper averaging over test appearances and are then aggregated using the bootstrap paper weights.

This procedure preserves the original split design while propagating uncertainty from paper resampling and parameter selection. We run 1,000 resamples using frozen scores and probabilities, without new reviews. The original, unresampled fit and paper-level risks match the saved main-paper results to a tolerance of $10^{-12}$.

\begin{table*}[tp]\centering
\normalsize
\setlength{\tabcolsep}{4pt}
\renewcommand{\arraystretch}{1.12}
\renewcommand{\thetable}{S23}
\begin{tabular}{@{}>{\raggedright\arraybackslash}p{103.081pt}>{\raggedleft\arraybackslash}p{78.639pt}>{\raggedleft\arraybackslash}p{127.809pt}>{\raggedleft\arraybackslash}p{121.715pt}@{}}
\toprule
{\textbf{\hspace{0pt}Method}} & {\textbf{\hspace{0pt}Risk}} & {\textbf{\hspace{0pt}Fixed-parameter 95\% CI (10,000 resamples)}} & {\textbf{\hspace{0pt}Refit 95\% CI (1,000 resamples)}} \\
\midrule
{\hspace{0pt}Quick} & {\hspace{0pt}0.2427} & {\hspace{0pt}[0.2029, 0.2860]} & {\hspace{0pt}[0.2020, 0.2934]} \\
{\hspace{0pt}Standard} & {\hspace{0pt}0.0829} & {\hspace{0pt}[0.0750, 0.0911]} & {\hspace{0pt}[0.0746, 0.0924]} \\
{\hspace{0pt}1 Reasoning} & {\hspace{0pt}0.2434} & {\hspace{0pt}[0.2051, 0.2828]} & {\hspace{0pt}[0.2040, 0.2823]} \\
{\hspace{0pt}2 Reasoning} & {\hspace{0pt}0.1217} & {\hspace{0pt}[0.1025, 0.1414]} & {\hspace{0pt}[0.1020, 0.1412]} \\
{\hspace{0pt}3 Reasoning} & {\hspace{0pt}0.0811} & {\hspace{0pt}[0.0684, 0.0943]} & {\hspace{0pt}[0.0680, 0.0941]} \\
{\hspace{0pt}5 Reasoning} & {\hspace{0pt}0.0487} & {\hspace{0pt}[0.0410, 0.0566]} & {\hspace{0pt}[0.0408, 0.0565]} \\
\bottomrule
\end{tabular}
\caption{Fixed-parameter and refit intervals. Direct-sampling baselines have no fitted parameters; their two columns use different bootstrap resampling settings.}
\end{table*}

With refitting, the risk difference between Standard and two reasoning reviews is $-$0.0388, with a 95\% interval of [$-$0.0525, $-$0.0259]. Relative to three reasoning reviews, it is +0.0018, with an interval of [$-$0.0062, +0.0091]. The finding that Standard reduces risk relative to two reasoning reviews and approaches the risk of three persists after parameter reselection.

\subsection{Acceptance AUC Under the Same Held-Out Design}

All methods use the same 198 papers, parameters, and test splits as the risk analysis. Predictions are distributional expected scores; parameters are not fitted to acceptance labels. Direct-reasoning baselines enumerate all equal-sized subsets of the five scores, while Standard enumerates the ten two-input combinations. AUC is computed on test papers for each combination, then averaged over combinations and the 30 splits.

\begin{table*}[tp]\centering
\normalsize
\setlength{\tabcolsep}{4pt}
\renewcommand{\arraystretch}{1.12}
\renewcommand{\thetable}{S24}
\begin{tabular}{@{}>{\raggedright\arraybackslash}p{88.310pt}>{\raggedleft\arraybackslash}p{47.676pt}>{\raggedleft\arraybackslash}p{72.734pt}>{\raggedleft\arraybackslash}p{88.618pt}>{\raggedleft\arraybackslash}p{125.908pt}@{}}
\toprule
{\textbf{\hspace{0pt}Method}} & {\textbf{\hspace{0pt}R/\allowbreak{}Q}} & {\textbf{\hspace{0pt}Mean held-out AUC: all 198}} & {\textbf{\hspace{0pt}Excl. withdrawn: 150 papers}} & {\textbf{\hspace{0pt}Independent-subset AUC (198 papers)}} \\
\midrule
{\hspace{0pt}Quick} & {\hspace{0pt}0/\allowbreak{}1} & {\hspace{0pt}0.7092} & {\hspace{0pt}0.7039} & {\hspace{0pt}0.7092} \\
{\hspace{0pt}Standard} & {\hspace{0pt}2/\allowbreak{}1} & {\hspace{0pt}0.7194} & {\hspace{0pt}0.7187} & {\hspace{0pt}0.7194} \\
{\hspace{0pt}1 Reasoning} & {\hspace{0pt}1/\allowbreak{}0} & {\hspace{0pt}0.6895} & {\hspace{0pt}0.6898} & {\hspace{0pt}0.6899} \\
{\hspace{0pt}2 Reasoning} & {\hspace{0pt}2/\allowbreak{}0} & {\hspace{0pt}0.7063} & {\hspace{0pt}0.7072} & {\hspace{0pt}0.7066} \\
{\hspace{0pt}3 Reasoning} & {\hspace{0pt}3/\allowbreak{}0} & {\hspace{0pt}0.7124} & {\hspace{0pt}0.7135} & {\hspace{0pt}0.7127} \\
{\hspace{0pt}5 Reasoning} & {\hspace{0pt}5/\allowbreak{}0} & {\hspace{0pt}0.7162} & {\hspace{0pt}0.7176} & {\hspace{0pt}0.7162} \\
\bottomrule
\end{tabular}
\caption{Acceptance discrimination under a common held-out design. The last column additionally selects score subsets independently for each paper; the main columns use the same round-index subset across papers in each comparison.}
\end{table*}

On all 198 papers, Quick exceeds one reasoning review in AUC by +0.0197, and Standard exceeds two reasoning reviews by +0.0131. Excluding withdrawals gives differences of +0.0141 and +0.0115. This table reports point estimates; intervals for squared risk concern a different statistic and are not AUC intervals.

\subsection{Independent-Target Validation on Ten Papers}

The existing independent-target validation sampled five papers per year, excluding the long-input paper \texttt{2026-\allowbreak{}paper\_\allowbreak{}067} before sampling. Each paper received ten additional GPT-5.6-Sol medium target scores. Predictions use the original five scores. Quick and Standard are fitted on the other 188 papers, yielding $T=8,a=0.2,b=0.9,w=0.7$.

For Quick, every available auxiliary output for a paper forms a separate one-call prediction, and losses are averaged without first combining the outputs into a single prediction. Standard uses the first auxiliary output and all two-score combinations from the original five scores. Direct-reasoning methods enumerate all subsets of the corresponding size. All methods first average over prediction combinations within each paper, then equally over the ten papers.

\begin{table*}[tp]\centering
\normalsize
\setlength{\tabcolsep}{4pt}
\renewcommand{\arraystretch}{1.12}
\renewcommand{\thetable}{S25}
\begin{tabular}{@{}>{\raggedright\arraybackslash}p{79.758pt}>{\raggedleft\arraybackslash}p{50.113pt}>{\raggedleft\arraybackslash}p{61.124pt}>{\raggedleft\arraybackslash}p{67.218pt}>{\raggedleft\arraybackslash}p{82.486pt}>{\raggedleft\arraybackslash}p{74.544pt}@{}}
\toprule
{\textbf{\hspace{0pt}Method}} & {\textbf{\hspace{0pt}Brier}} & {\textbf{\hspace{0pt}W1 (points)}} & {\textbf{\hspace{0pt}Modal accuracy}} & {\textbf{\hspace{0pt}Absolute target-mean error (points)}} & {\textbf{\hspace{0pt}Absolute error in P(score$\geq$6)}} \\
\midrule
{\hspace{0pt}Quick} & {\hspace{0pt}0.497} & {\hspace{0pt}0.363} & {\hspace{0pt}60.0\%} & {\hspace{0pt}0.315} & {\hspace{0pt}0.141} \\
{\hspace{0pt}Standard} & {\hspace{0pt}0.353} & {\hspace{0pt}0.194} & {\hspace{0pt}69.0\%} & {\hspace{0pt}0.163} & {\hspace{0pt}0.113} \\
{\hspace{0pt}1 Reasoning} & {\hspace{0pt}0.552} & {\hspace{0pt}0.276} & {\hspace{0pt}72.0\%} & {\hspace{0pt}0.276} & {\hspace{0pt}0.132} \\
{\hspace{0pt}2 Reasoning} & {\hspace{0pt}0.422} & {\hspace{0pt}0.194} & {\hspace{0pt}75.0\%} & {\hspace{0pt}0.194} & {\hspace{0pt}0.108} \\
{\hspace{0pt}3 Reasoning} & {\hspace{0pt}0.379} & {\hspace{0pt}0.178} & {\hspace{0pt}77.0\%} & {\hspace{0pt}0.178} & {\hspace{0pt}0.104} \\
{\hspace{0pt}5 Reasoning} & {\hspace{0pt}0.344} & {\hspace{0pt}0.140} & {\hspace{0pt}80.0\%} & {\hspace{0pt}0.140} & {\hspace{0pt}0.100} \\
\bottomrule
\end{tabular}
\caption{Complete metrics for the existing ten-paper independent-target validation. The two-reasoning row is computed from existing scores; all five fixed baselines are retained.}
\end{table*}

Brier loss is computed against each of the ten one-hot targets and then averaged, including randomness in the target scores themselves. W1 compares the predicted distribution with the empirical CDF of the ten target scores, in score points. Modal accuracy compares the modes of the prediction and the ten target scores, choosing the smallest score in a tie. Mean-score error and $P(Y\ge6)$ error compare the respective means and tail probabilities.

In this independent-target set, Standard has Brier loss 0.353, W1 of 0.194 points, and modal accuracy of 69\%; three reasoning reviews yield 0.379, 0.178 points, and 77\%. These metrics measure the full distribution, cumulative distribution, and most likely score, respectively, and are reported together.

\subsection{Historical Call Costs}

The existing cost audit uses the original script's historical pretax USD rates per million tokens: 4 for ordinary input, 5 for cache writes, 0.4 for cached input, and 20 for output. Call cost is computed from usage fields as

\[
\mathrm{cost}=\frac{4(I-C-W)+0.4C+5W+20O}{10^6},
\]

Here $I$ denotes input tokens, $C$ cached-input tokens, $W$ cache-write tokens, and $O$ output tokens.

\begin{table*}[tp]\centering
\normalsize
\setlength{\tabcolsep}{4pt}
\renewcommand{\arraystretch}{1.12}
\renewcommand{\thetable}{S26}
\begin{tabular}{@{}>{\raggedright\arraybackslash}p{134.987pt}>{\raggedleft\arraybackslash}p{50.067pt}>{\raggedleft\arraybackslash}p{61.661pt}>{\raggedleft\arraybackslash}p{53.125pt}>{\raggedleft\arraybackslash}p{57.701pt}>{\raggedleft\arraybackslash}p{57.701pt}@{}}
\toprule
{\textbf{\hspace{0pt}Call type}} & {\textbf{\hspace{0pt}Shared papers}} & {\textbf{\hspace{0pt}Recorded calls}} & {\textbf{\hspace{0pt}Missing usage}} & {\textbf{\hspace{0pt}Mean cost, all calls (USD)}} & {\textbf{\hspace{0pt}Mean first-call cost per paper (USD)}} \\
\midrule
{\hspace{0pt}Auxiliary probability calls} & {\hspace{0pt}198} & {\hspace{0pt}578} & {\hspace{0pt}0} & {\hspace{0pt}0.067784} & {\hspace{0pt}0.153340} \\
{\hspace{0pt}medium reasoning reviews} & {\hspace{0pt}198} & {\hspace{0pt}990} & {\hspace{0pt}0} & {\hspace{0pt}0.063410} & {\hspace{0pt}0.171158} \\
\bottomrule
\end{tabular}
\caption{Historical per-call costs computed from recorded usage in the existing shared sample.}
\end{table*}

First and subsequent calls differ in cache state, so the table reports both historical averages. The main paper's R/\allowbreak{}Q columns describe call composition; this table gives observed costs for those calls without combining different cache states into a single deployment price. The frozen summaries do not contain consistent end-to-end latency measurements for comparing methods under matched conditions, so no latency estimate is reported.

\section{Revision Crossover Experiment}

\subsection{Seven-Cell Design and Paired Scoring}

Three letters denote Writer, Selector, and Scorer, in that order, with \texttt{L} for Luna and \texttt{H} for Haiku. Luna candidates are processed by two Selectors and evaluated by two Scorers, yielding \texttt{LLL}, \texttt{LHL}, \texttt{LLH}, and \texttt{LHH}. Haiku candidates yield \texttt{HLL}, \texttt{HHL}, and \texttt{HHH}; \texttt{HLH} was not collected. All cells use the same 199 papers.

Section B.5 describes the Writers' feedback sources. Each Writer's candidate set is held fixed across Selectors, and both Scorers read the same final manuscript for a given Writer/\allowbreak{}Selector combination. Each Scorer independently evaluates the original and revised manuscripts three times each. A Scorer's original-manuscript baseline is reused across revision conditions. Paper-level gain is defined as

\[
\hat\Delta_{r,i}=\frac13\sum_{t=1}^{3}Y'_{r,it}-\frac13\sum_{t=1}^{3}Y_{r,it}.
\]

Positive, zero, and negative gains are determined by the difference between the two sums of three scores, avoiding floating-point changes to direction. Overall gains and cross-Scorer differences use paired paper-level resampling.

\begin{table*}[tp]\centering
\small
\setlength{\tabcolsep}{4pt}
\renewcommand{\arraystretch}{1.12}
\renewcommand{\thetable}{S27}
\begin{tabular}{@{}>{\raggedright\arraybackslash}p{52.733pt}>{\raggedleft\arraybackslash}p{39.403pt}>{\raggedleft\arraybackslash}p{33.303pt}>{\raggedleft\arraybackslash}p{54.423pt}>{\raggedleft\arraybackslash}p{51.633pt}>{\raggedleft\arraybackslash}p{116.883pt}>{\raggedleft\arraybackslash}p{58.863pt}@{}}
\toprule
{\textbf{\hspace{0pt}Writer/\allowbreak{}Selector/\allowbreak{}Scorer}} & {\textbf{\hspace{0pt}Year}} & {\textbf{\hspace{0pt}N}} & {\textbf{\hspace{0pt}Original mean}} & {\textbf{\hspace{0pt}Revised mean}} & {\textbf{\hspace{0pt}Gain and 95\% CI}} & {\textbf{\hspace{0pt}Positive/\allowbreak{}zero/\allowbreak{}negative}} \\
\midrule
{\hspace{0pt}LLL} & {\hspace{0pt}All} & {\hspace{0pt}199} & {\hspace{0pt}3.441} & {\hspace{0pt}3.915} & {\hspace{0pt}+0.474 [+0.367, +0.578]} & {\hspace{0pt}120/\allowbreak{}54/\allowbreak{}25} \\
{\hspace{0pt}LLL} & {\hspace{0pt}2025} & {\hspace{0pt}100} & {\hspace{0pt}3.797} & {\hspace{0pt}4.077} & {\hspace{0pt}+0.280 [+0.137, +0.417]} & {\hspace{0pt}50/\allowbreak{}35/\allowbreak{}15} \\
{\hspace{0pt}LLL} & {\hspace{0pt}2026} & {\hspace{0pt}99} & {\hspace{0pt}3.081} & {\hspace{0pt}3.751} & {\hspace{0pt}+0.670 [+0.512, +0.818]} & {\hspace{0pt}70/\allowbreak{}19/\allowbreak{}10} \\
\addlinespace[3pt]
{\hspace{0pt}LHL} & {\hspace{0pt}All} & {\hspace{0pt}199} & {\hspace{0pt}3.441} & {\hspace{0pt}3.913} & {\hspace{0pt}+0.472 [+0.362, +0.580]} & {\hspace{0pt}129/\allowbreak{}41/\allowbreak{}29} \\
{\hspace{0pt}LHL} & {\hspace{0pt}2025} & {\hspace{0pt}100} & {\hspace{0pt}3.797} & {\hspace{0pt}4.103} & {\hspace{0pt}+0.307 [+0.153, +0.453]} & {\hspace{0pt}60/\allowbreak{}23/\allowbreak{}17} \\
{\hspace{0pt}LHL} & {\hspace{0pt}2026} & {\hspace{0pt}99} & {\hspace{0pt}3.081} & {\hspace{0pt}3.721} & {\hspace{0pt}+0.640 [+0.481, +0.791]} & {\hspace{0pt}69/\allowbreak{}18/\allowbreak{}12} \\
\addlinespace[3pt]
{\hspace{0pt}LLH} & {\hspace{0pt}All} & {\hspace{0pt}199} & {\hspace{0pt}4.241} & {\hspace{0pt}3.893} & {\hspace{0pt}-0.348 [-0.432, -0.265]} & {\hspace{0pt}32/\allowbreak{}63/\allowbreak{}104} \\
{\hspace{0pt}LLH} & {\hspace{0pt}2025} & {\hspace{0pt}100} & {\hspace{0pt}4.473} & {\hspace{0pt}4.057} & {\hspace{0pt}-0.417 [-0.540, -0.293]} & {\hspace{0pt}15/\allowbreak{}28/\allowbreak{}57} \\
{\hspace{0pt}LLH} & {\hspace{0pt}2026} & {\hspace{0pt}99} & {\hspace{0pt}4.007} & {\hspace{0pt}3.727} & {\hspace{0pt}-0.279 [-0.387, -0.175]} & {\hspace{0pt}17/\allowbreak{}35/\allowbreak{}47} \\
\addlinespace[3pt]
{\hspace{0pt}LHH} & {\hspace{0pt}All} & {\hspace{0pt}199} & {\hspace{0pt}4.241} & {\hspace{0pt}3.945} & {\hspace{0pt}-0.296 [-0.379, -0.214]} & {\hspace{0pt}31/\allowbreak{}63/\allowbreak{}105} \\
{\hspace{0pt}LHH} & {\hspace{0pt}2025} & {\hspace{0pt}100} & {\hspace{0pt}4.473} & {\hspace{0pt}4.117} & {\hspace{0pt}-0.357 [-0.477, -0.233]} & {\hspace{0pt}17/\allowbreak{}23/\allowbreak{}60} \\
{\hspace{0pt}LHH} & {\hspace{0pt}2026} & {\hspace{0pt}99} & {\hspace{0pt}4.007} & {\hspace{0pt}3.771} & {\hspace{0pt}-0.236 [-0.340, -0.135]} & {\hspace{0pt}14/\allowbreak{}40/\allowbreak{}45} \\
\addlinespace[3pt]
{\hspace{0pt}HLL} & {\hspace{0pt}All} & {\hspace{0pt}199} & {\hspace{0pt}3.441} & {\hspace{0pt}3.742} & {\hspace{0pt}+0.302 [+0.238, +0.367]} & {\hspace{0pt}108/\allowbreak{}70/\allowbreak{}21} \\
{\hspace{0pt}HLL} & {\hspace{0pt}2025} & {\hspace{0pt}100} & {\hspace{0pt}3.797} & {\hspace{0pt}4.003} & {\hspace{0pt}+0.207 [+0.120, +0.293]} & {\hspace{0pt}47/\allowbreak{}38/\allowbreak{}15} \\
{\hspace{0pt}HLL} & {\hspace{0pt}2026} & {\hspace{0pt}99} & {\hspace{0pt}3.081} & {\hspace{0pt}3.478} & {\hspace{0pt}+0.397 [+0.303, +0.495]} & {\hspace{0pt}61/\allowbreak{}32/\allowbreak{}6} \\
\addlinespace[3pt]
{\hspace{0pt}HHL} & {\hspace{0pt}All} & {\hspace{0pt}199} & {\hspace{0pt}3.441} & {\hspace{0pt}3.722} & {\hspace{0pt}+0.281 [+0.216, +0.347]} & {\hspace{0pt}96/\allowbreak{}84/\allowbreak{}19} \\
{\hspace{0pt}HHL} & {\hspace{0pt}2025} & {\hspace{0pt}100} & {\hspace{0pt}3.797} & {\hspace{0pt}3.993} & {\hspace{0pt}+0.197 [+0.117, +0.280]} & {\hspace{0pt}40/\allowbreak{}49/\allowbreak{}11} \\
{\hspace{0pt}HHL} & {\hspace{0pt}2026} & {\hspace{0pt}99} & {\hspace{0pt}3.081} & {\hspace{0pt}3.448} & {\hspace{0pt}+0.367 [+0.273, +0.468]} & {\hspace{0pt}56/\allowbreak{}35/\allowbreak{}8} \\
\addlinespace[3pt]
{\hspace{0pt}HHH} & {\hspace{0pt}All} & {\hspace{0pt}199} & {\hspace{0pt}4.241} & {\hspace{0pt}4.291} & {\hspace{0pt}+0.050 [+0.000, +0.104]} & {\hspace{0pt}57/\allowbreak{}102/\allowbreak{}40} \\
{\hspace{0pt}HHH} & {\hspace{0pt}2025} & {\hspace{0pt}100} & {\hspace{0pt}4.473} & {\hspace{0pt}4.500} & {\hspace{0pt}+0.027 [-0.050, +0.107]} & {\hspace{0pt}27/\allowbreak{}50/\allowbreak{}23} \\
{\hspace{0pt}HHH} & {\hspace{0pt}2026} & {\hspace{0pt}99} & {\hspace{0pt}4.007} & {\hspace{0pt}4.081} & {\hspace{0pt}+0.074 [+0.010, +0.138]} & {\hspace{0pt}30/\allowbreak{}52/\allowbreak{}17} \\
\bottomrule
\end{tabular}
\caption{Complete means, gains, and year-specific results for the seven-cell revision experiment. Overall intervals retain the main paper's original 10,000-resample bootstrap outputs; year-specific intervals use 10,000 paper-level resamples.}
\end{table*}

Both years retain opposite average gains for Luna revisions under the two Scorers. For example, year-specific gains for \texttt{LLL} are +0.280 and +0.670, whereas those for \texttt{LLH} are $-$0.417 and $-$0.279. The year-specific results for \texttt{LHL} and \texttt{LHH} likewise remain positive and negative, respectively.

\subsection{Candidate Edits and Selector Agreement}

The two Selectors independently accept or reject each candidate edit. For accepted sets $A_i,B_i$, Jaccard similarity is $J_i=|A_i\cap B_i|/|A_i\cup B_i|$, defined as 1 when both sets are empty. The table reports candidate counts and acceptance intersections for each Writer.

\begin{table*}[tp]\centering
\small
\setlength{\tabcolsep}{4pt}
\renewcommand{\arraystretch}{1.12}
\renewcommand{\thetable}{S28}
\begin{tabular}{@{}>{\raggedright\arraybackslash}p{46.283pt}>{\raggedleft\arraybackslash}p{65.193pt}>{\raggedleft\arraybackslash}p{47.953pt}>{\raggedleft\arraybackslash}p{47.953pt}>{\raggedleft\arraybackslash}p{44.063pt}>{\raggedleft\arraybackslash}p{39.643pt}>{\raggedleft\arraybackslash}p{43.533pt}>{\raggedleft\arraybackslash}p{64.623pt}@{}}
\toprule
{\textbf{\hspace{0pt}Writer}} & {\textbf{\hspace{0pt}Candidates}} & {\textbf{\hspace{0pt}Luna accepts}} & {\textbf{\hspace{0pt}Haiku accepts}} & {\textbf{\hspace{0pt}Both accept}} & {\textbf{\hspace{0pt}Luna only}} & {\textbf{\hspace{0pt}Haiku only}} & {\textbf{\hspace{0pt}Mean paper-level Jaccard}} \\
\midrule
{\hspace{0pt}Luna} & {\hspace{0pt}9327} & {\hspace{0pt}8857} & {\hspace{0pt}8722} & {\hspace{0pt}8459} & {\hspace{0pt}398} & {\hspace{0pt}263} & {\hspace{0pt}0.9188} \\
{\hspace{0pt}Haiku} & {\hspace{0pt}2587} & {\hspace{0pt}2442} & {\hspace{0pt}2450} & {\hspace{0pt}2354} & {\hspace{0pt}88} & {\hspace{0pt}96} & {\hspace{0pt}0.9198} \\
\bottomrule
\end{tabular}
\caption{Complete comparison of the two Selectors' accepted sets with the Writer fixed. Jaccard similarity is computed per paper and then averaged over 199 papers.}
\end{table*}

\begin{table*}[tp]\centering
\normalsize
\setlength{\tabcolsep}{4pt}
\renewcommand{\arraystretch}{1.12}
\renewcommand{\thetable}{S29}
\begin{tabular}{@{}>{\raggedright\arraybackslash}p{257.216pt}>{\raggedleft\arraybackslash}p{190.028pt}@{}}
\toprule
{\textbf{\hspace{0pt}Metric}} & {\textbf{\hspace{0pt}Value}} \\
\midrule
{\hspace{0pt}candidates} & {\hspace{0pt}9327} \\
{\hspace{0pt}accepted\_luna} & {\hspace{0pt}8857} \\
{\hspace{0pt}accepted\_haiku} & {\hspace{0pt}8722} \\
{\hspace{0pt}both} & {\hspace{0pt}8459} \\
{\hspace{0pt}luna\_only} & {\hspace{0pt}398} \\
{\hspace{0pt}haiku\_only} & {\hspace{0pt}263} \\
{\hspace{0pt}decision\_disagreements} & {\hspace{0pt}661} \\
{\hspace{0pt}Mean paper-level Jaccard} & {\hspace{0pt}0.9188} \\
{\hspace{0pt}Paper-level Jaccard Q25/\allowbreak{}median/\allowbreak{}Q75} & {\hspace{0pt}0.8817 /\allowbreak{} 0.9400 /\allowbreak{} 0.9694} \\
{\hspace{0pt}Candidates per paper: Q25/\allowbreak{}median/\allowbreak{}Q75} & {\hspace{0pt}33.0 /\allowbreak{} 41.0 /\allowbreak{} 53.5} \\
{\hspace{0pt}Identical decision sets} & {\hspace{0pt}17} \\
{\hspace{0pt}Identical final text} & {\hspace{0pt}17} \\
\bottomrule
\end{tabular}
\caption{Paper-level selection statistics for Luna candidates. Candidate totals, acceptance intersections, and one-sided acceptances count diffs; quantiles and identical-decision counts are paper-level quantities.}
\end{table*}

Of Luna's 9,327 candidates, 8,459 are accepted by both Selectors and 661 receive different decisions. Decisions are identical for 17 papers; the other 182 have different final texts. Candidate acceptance and final scoring occur at different stages: a large overlap in accepted candidates coexists with opposite gains from the two Scorers on fixed final manuscripts.

Haiku generates 2,587 candidates. Luna and Haiku accept 2,442 and 2,450, respectively, with 2,354 accepted by both. Mean paper-level Jaccard similarity is 0.9198. The Writers produce different numbers of edits, while selection agreement within their respective candidate sets is similar.

\subsection{Paper-Level Gain Directions and Cross-Model Differences}

Beyond overall gains, we classify each paper's difference between three-review means as positive, zero, or negative. The table gives complete direction contingency tables for the three shared-manuscript conditions.

\begin{table*}[tp]\centering
\normalsize
\setlength{\tabcolsep}{4pt}
\renewcommand{\arraystretch}{1.12}
\renewcommand{\thetable}{S30}
\begin{tabular}{@{}>{\raggedright\arraybackslash}p{85.612pt}>{\raggedright\arraybackslash}p{89.902pt}>{\raggedleft\arraybackslash}p{83.797pt}>{\raggedleft\arraybackslash}p{77.087pt}>{\raggedleft\arraybackslash}p{86.844pt}@{}}
\toprule
{\textbf{\hspace{0pt}Writer/\allowbreak{}Selector}} & {\textbf{\hspace{0pt}Luna direction}} & {\textbf{\hspace{0pt}Haiku positive}} & {\textbf{\hspace{0pt}Haiku zero}} & {\textbf{\hspace{0pt}Haiku negative}} \\
\midrule
{\hspace{0pt}LL} & {\hspace{0pt}Positive} & {\hspace{0pt}27} & {\hspace{0pt}47} & {\hspace{0pt}46} \\
{\hspace{0pt}LL} & {\hspace{0pt}Zero} & {\hspace{0pt}5} & {\hspace{0pt}10} & {\hspace{0pt}39} \\
{\hspace{0pt}LL} & {\hspace{0pt}Negative} & {\hspace{0pt}0} & {\hspace{0pt}6} & {\hspace{0pt}19} \\
\addlinespace[3pt]
{\hspace{0pt}LH} & {\hspace{0pt}Positive} & {\hspace{0pt}27} & {\hspace{0pt}44} & {\hspace{0pt}58} \\
{\hspace{0pt}LH} & {\hspace{0pt}Zero} & {\hspace{0pt}1} & {\hspace{0pt}13} & {\hspace{0pt}27} \\
{\hspace{0pt}LH} & {\hspace{0pt}Negative} & {\hspace{0pt}3} & {\hspace{0pt}6} & {\hspace{0pt}20} \\
\addlinespace[3pt]
{\hspace{0pt}HH} & {\hspace{0pt}Positive} & {\hspace{0pt}31} & {\hspace{0pt}47} & {\hspace{0pt}18} \\
{\hspace{0pt}HH} & {\hspace{0pt}Zero} & {\hspace{0pt}21} & {\hspace{0pt}46} & {\hspace{0pt}17} \\
{\hspace{0pt}HH} & {\hspace{0pt}Negative} & {\hspace{0pt}5} & {\hspace{0pt}9} & {\hspace{0pt}5} \\
\bottomrule
\end{tabular}
\caption{Paper-level gain directions under Luna and Haiku. The nine counts for each Writer/\allowbreak{}Selector condition sum to 199.}
\end{table*}

For Luna-written, Luna-selected manuscripts, 46 papers have positive Luna gains and negative Haiku gains, and none show the reverse combination. Both gains are positive for 27 papers, negative for 19, and zero for 10. The two strict reversal combinations total 61 papers for Luna-written, Haiku-selected manuscripts and 23 for Haiku-written, Haiku-selected manuscripts.

\begin{figure*}[tp]\centering
\includegraphics[width=\textwidth]{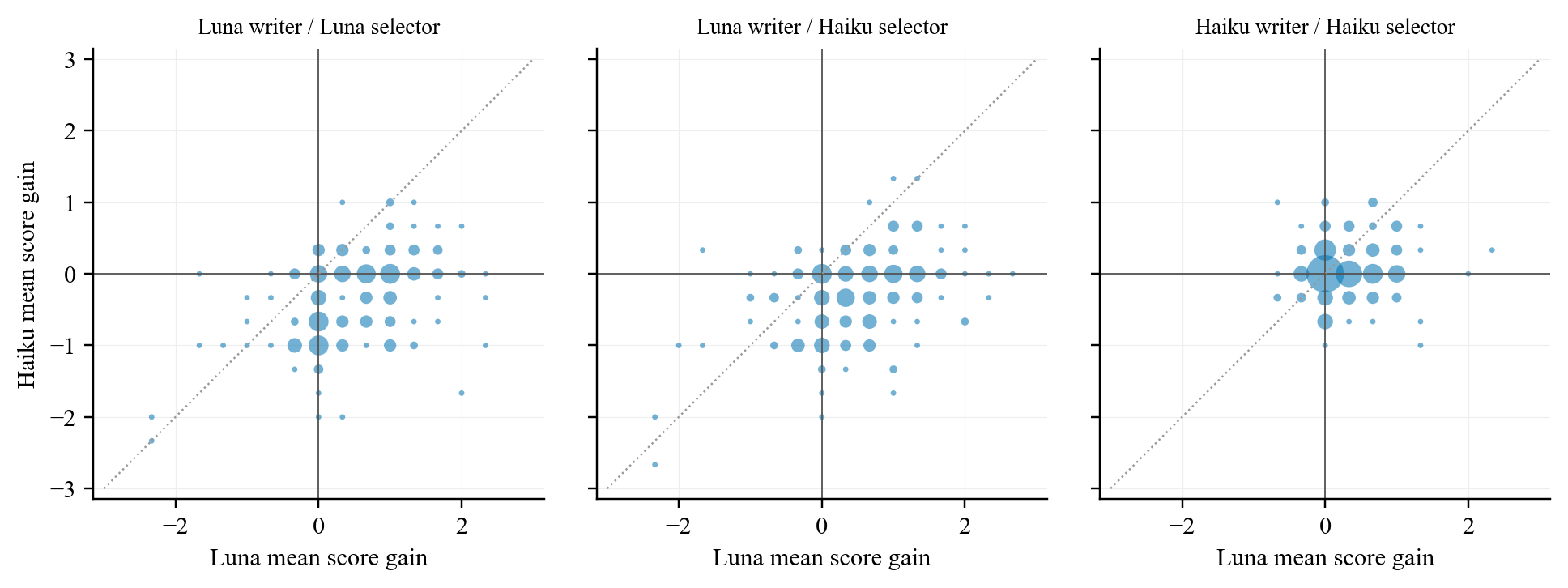}
\renewcommand{\thefigure}{F1}
\caption{Each panel contains gains for 199 papers, with Luna on the horizontal axis and Haiku on the vertical axis. Coincident coordinates are combined into bubbles whose area is proportional to the paper count. Dashed lines mark equal gains under the two Scorers. Coordinates are not jittered.}
\end{figure*}

\begin{table*}[tp]\centering
\normalsize
\setlength{\tabcolsep}{4pt}
\renewcommand{\arraystretch}{1.12}
\renewcommand{\thetable}{S31}
\begin{tabular}{@{}>{\raggedright\arraybackslash}p{84.576pt}>{\raggedleft\arraybackslash}p{84.587pt}>{\raggedleft\arraybackslash}p{94.388pt}>{\raggedleft\arraybackslash}p{84.576pt}>{\raggedleft\arraybackslash}p{75.116pt}@{}}
\toprule
{\textbf{\hspace{0pt}Writer/\allowbreak{}Selector}} & {\textbf{\hspace{0pt}Gain Pearson}} & {\textbf{\hspace{0pt}Same direction (including both zero)}} & {\textbf{\hspace{0pt}Strict reversal}} & {\textbf{\hspace{0pt}At least one zero}} \\
\midrule
{\hspace{0pt}LL} & {\hspace{0pt}0.415} & {\hspace{0pt}28.1\%} & {\hspace{0pt}23.1\%} & {\hspace{0pt}53.8\%} \\
{\hspace{0pt}LH} & {\hspace{0pt}0.407} & {\hspace{0pt}30.2\%} & {\hspace{0pt}30.7\%} & {\hspace{0pt}45.7\%} \\
{\hspace{0pt}HH} & {\hspace{0pt}0.085} & {\hspace{0pt}41.2\%} & {\hspace{0pt}11.6\%} & {\hspace{0pt}70.4\%} \\
\bottomrule
\end{tabular}
\par\medskip
\begin{tabular}{@{}>{\raggedright\arraybackslash}p{85.269pt}>{\raggedleft\arraybackslash}p{155.834pt}>{\raggedleft\arraybackslash}p{95.070pt}>{\raggedleft\arraybackslash}p{95.070pt}@{}}
\toprule
{\textbf{\hspace{0pt}Writer/\allowbreak{}Selector}} & {\textbf{\hspace{0pt}Luna$-$Haiku and 95\% CI}} & {\textbf{\hspace{0pt}Luna mean gain-noise SD}} & {\textbf{\hspace{0pt}Haiku mean gain-noise SD}} \\
\midrule
{\hspace{0pt}LL} & {\hspace{0pt}+0.822 [+0.717, +0.928]} & {\hspace{0pt}0.237} & {\hspace{0pt}0.210} \\
{\hspace{0pt}LH} & {\hspace{0pt}+0.769 [+0.662, +0.879]} & {\hspace{0pt}0.243} & {\hspace{0pt}0.232} \\
{\hspace{0pt}HH} & {\hspace{0pt}+0.231 [+0.154, +0.310]} & {\hspace{0pt}0.245} & {\hspace{0pt}0.237} \\
\bottomrule
\end{tabular}
\caption{Paper-level gain associations, direction distributions, and mean cross-Scorer differences. Gain-noise SD is estimated from the sample variances of three original and three revised scores, taking the square root within each paper before averaging.}
\end{table*}

At least one Scorer assigns zero gain in 53.8\%, 45.7\%, and 70.4\% of papers across the three conditions. Direction disagreement can therefore be separated into strict reversals and cases involving zero gain; the full contingency tables preserve this distinction.

\subsection{Scoring Noise and Correlation Adjustment}

Under a model with fixed manuscripts and independent, zero-mean scoring errors, the noise variance of the three-review original mean is $e_{o,i}=S_{o,i}^2/3$, and the noise variance of the gain estimate is

\[
e_{\Delta,i}=\frac{S_{o,i}^2+S_{v,i}^2}{3}.
\]

Let $v_x,v_y$ be the observed across-paper variances for two Scorers, $e_x,e_y$ their mean noise variances, and $c$ their observed covariance. Keeping the covariance unchanged and subtracting diagonal noise terms gives

\[
\rho_{\mathrm{adj}}=\frac{c}{\sqrt{(v_x-e_x)(v_y-e_y)}}.
\]

\begin{table*}[tp]\centering
\normalsize
\setlength{\tabcolsep}{4pt}
\renewcommand{\arraystretch}{1.12}
\renewcommand{\thetable}{S32}
\begin{tabular}{@{}>{\raggedright\arraybackslash}p{119.509pt}>{\raggedleft\arraybackslash}p{95.489pt}>{\raggedleft\arraybackslash}p{95.489pt}>{\raggedleft\arraybackslash}p{49.047pt}>{\raggedleft\arraybackslash}p{63.710pt}@{}}
\toprule
{\textbf{\hspace{0pt}Target}} & {\textbf{\hspace{0pt}Observed correlation and 95\% CI}} & {\textbf{\hspace{0pt}Adjusted correlation and 95\% CI}} & {\textbf{\hspace{0pt}Valid adjusted resamples}} & {\textbf{\hspace{0pt}Out-of-range resamples}} \\
\midrule
{\hspace{0pt}Original} & {\hspace{0pt}0.804 [0.741, 0.854]} & {\hspace{0pt}0.843 [0.783, 0.890]} & {\hspace{0pt}10000} & {\hspace{0pt}0} \\
{\hspace{0pt}Luna Writer/\allowbreak{}Luna Selector gain} & {\hspace{0pt}0.415 [0.261, 0.551]} & {\hspace{0pt}0.521 [0.336, 0.686]} & {\hspace{0pt}10000} & {\hspace{0pt}0} \\
{\hspace{0pt}Luna Writer/\allowbreak{}Haiku Selector gain} & {\hspace{0pt}0.407 [0.255, 0.541]} & {\hspace{0pt}0.518 [0.337, 0.673]} & {\hspace{0pt}10000} & {\hspace{0pt}0} \\
{\hspace{0pt}Haiku Writer/\allowbreak{}Haiku Selector gain} & {\hspace{0pt}0.085 [-0.071, 0.237]} & {\hspace{0pt}0.230 [-0.200, 0.902]} & {\hspace{0pt}9767} & {\hspace{0pt}189} \\
\bottomrule
\end{tabular}
\caption{Correlations before and after noise adjustment. Intervals retain the original analysis's 10,000 paper-level bootstrap resamples. Out-of-range adjusted correlations are not clipped.}
\end{table*}

The original-manuscript correlation is 0.804 before adjustment and 0.843 after it. Observed gain correlations for the two Luna revision conditions are 0.415 and 0.407, rising to 0.521 and 0.518 after adjustment. For Haiku revisions, the observed correlation is 0.085 and the adjusted value is approximately 0.230, with a wide interval. Some bootstrap samples have nonpositive noise-adjusted variances or correlations outside the usual range. The table retains valid-resample counts and out-of-range cases.

The adjustment assumes independent errors across models and between original and revised manuscripts. Writers use existing first reviews to generate revisions; Section B.5 specifies their actual relationship to the original-score baselines. We use noise subtraction to diagnose the contribution of scoring error. Mean cross-Scorer differences for fixed final manuscripts are computed directly from paired scores and do not depend on this adjustment.

\subsection{Structural Checks of Manuscript Packages}

We compare original and final packages paper by paper for the four Writer/\allowbreak{}Selector conditions, checking record counts and order, fields other than \texttt{text}, leading instructions and page-marker text, image indices and stored hashes, and the existence of referenced image files.

\begin{table*}[tp]\centering
\normalsize
\setlength{\tabcolsep}{4pt}
\renewcommand{\arraystretch}{1.12}
\renewcommand{\thetable}{S33}
\begin{tabular}{@{}>{\raggedright\arraybackslash}p{84.394pt}>{\raggedleft\arraybackslash}p{78.905pt}>{\raggedleft\arraybackslash}p{77.695pt}>{\raggedleft\arraybackslash}p{78.289pt}>{\raggedright\arraybackslash}p{103.963pt}@{}}
\toprule
{\textbf{\hspace{0pt}Writer/\allowbreak{}Selector}} & {\textbf{\hspace{0pt}Papers}} & {\textbf{\hspace{0pt}Edited text blocks}} & {\textbf{\hspace{0pt}Image entries}} & {\textbf{\hspace{0pt}Nontext fields /\allowbreak{} protected records /\allowbreak{} image-index checks}} \\
\midrule
{\hspace{0pt}LL} & {\hspace{0pt}199} & {\hspace{0pt}8857} & {\hspace{0pt}2360} & {\hspace{0pt}All passed} \\
{\hspace{0pt}LH} & {\hspace{0pt}199} & {\hspace{0pt}8722} & {\hspace{0pt}2360} & {\hspace{0pt}All passed} \\
{\hspace{0pt}HH} & {\hspace{0pt}199} & {\hspace{0pt}2450} & {\hspace{0pt}2360} & {\hspace{0pt}All passed} \\
{\hspace{0pt}HL} & {\hspace{0pt}199} & {\hspace{0pt}2442} & {\hspace{0pt}2360} & {\hspace{0pt}All passed} \\
\bottomrule
\end{tabular}
\caption{Structural checks for the four final-text conditions. Image counts refer to index entries; each condition uses the same 199 original papers.}
\end{table*}

All four conditions pass these checks. The Luna-written conditions accept 8,857 and 8,722 text blocks; the Haiku-written conditions accept 2,442 and 2,450. Recomputing SHA-256 for all 9,440 image files across the four final-package conditions gives exact matches to the original image-index hashes. This audit covers input-package structure, text-replacement rules, and image files. The study did not separately collect blinded expert labels for factual preservation or quality changes in the revisions.

\subsection{Real Revision Examples Selected by a Fixed Rule}

Cases are selected from five direction strata in \texttt{LL}: Luna positive/\allowbreak{}Haiku negative, both positive, both negative, Luna negative/\allowbreak{}Haiku positive, and at least one zero. Papers in each stratum are sorted by a fixed hash rule, and the first is selected. The Luna-negative/\allowbreak{}Haiku-positive stratum is empty. One additional case is selected by the same rule from Luna-positive/\allowbreak{}Haiku-zero papers in the Haiku-written, Haiku-selected condition.

Each case shows the first accepted block of suitable display length after sorting edits by page number and text-block ID. Both original and revised text must contain 70–750 characters after whitespace normalization; if no block qualifies, the first accepted block is used. Scores always refer to the complete revised manuscript. The displayed block illustrates the edit and is not assigned responsibility for the whole-paper gain.

\subsubsection{Case 1: Luna Positive /\allowbreak{} Haiku Negative}

Paper \texttt{2025-\allowbreak{}paper\_\allowbreak{}092}; Luna Writer/\allowbreak{}Luna Selector; 46 papers in this direction stratum. The displayed edit is on PDF page 3, text block \texttt{27}, diff \texttt{d0003}, and was accepted by this condition's Selector.

\begin{table*}[tp]\centering
\normalsize
\setlength{\tabcolsep}{4pt}
\renewcommand{\arraystretch}{1.12}
\renewcommand{\thetable}{S34}
\begin{tabular}{@{}>{\raggedright\arraybackslash}p{98.175pt}>{\raggedleft\arraybackslash}p{103.697pt}>{\raggedleft\arraybackslash}p{100.628pt}>{\raggedleft\arraybackslash}p{128.744pt}@{}}
\toprule
{\textbf{\hspace{0pt}Scorer}} & {\textbf{\hspace{0pt}Three original scores}} & {\textbf{\hspace{0pt}Three revised scores}} & {\textbf{\hspace{0pt}Whole-paper mean gain}} \\
\midrule
{\hspace{0pt}Luna} & {\hspace{0pt}3, 4, 4} & {\hspace{0pt}5, 5, 4} & {\hspace{0pt}+1.000} \\
{\hspace{0pt}Haiku} & {\hspace{0pt}5, 6, 5} & {\hspace{0pt}5, 5, 5} & {\hspace{0pt}-0.333} \\
\bottomrule
\end{tabular}
\caption{Original and complete revised-manuscript scores for Case 1.}
\end{table*}

Original:

\begin{quote}3. We demonstrate theoretically how a) SGFT can learn highly performant policies, despite the bias introduced by the SGFT objective and b) how this enables SGFT to overcome compounding errors which plague standard MBRL approaches.\end{quote}

Revised:

\begin{quote}3. We analyze how value-transfer bias and model error affect SGFT policies. The resulting guarantee is conditional on uniform closeness of Vsim to the optimal real-world value and improvability under real-world dynamics; neither condition is empirically verified in our experiments.\end{quote}

\subsubsection{Case 2: Both Positive}

Paper \texttt{2026-\allowbreak{}paper\_\allowbreak{}005}; Luna Writer/\allowbreak{}Luna Selector; 27 papers in this direction stratum. The displayed edit is on PDF page 1, text block \texttt{10}, diff \texttt{d0002}, and was accepted by this condition's Selector.

\begin{table*}[tp]\centering
\normalsize
\setlength{\tabcolsep}{4pt}
\renewcommand{\arraystretch}{1.12}
\renewcommand{\thetable}{S35}
\begin{tabular}{@{}>{\raggedright\arraybackslash}p{98.175pt}>{\raggedleft\arraybackslash}p{103.697pt}>{\raggedleft\arraybackslash}p{100.628pt}>{\raggedleft\arraybackslash}p{128.744pt}@{}}
\toprule
{\textbf{\hspace{0pt}Scorer}} & {\textbf{\hspace{0pt}Three original scores}} & {\textbf{\hspace{0pt}Three revised scores}} & {\textbf{\hspace{0pt}Whole-paper mean gain}} \\
\midrule
{\hspace{0pt}Luna} & {\hspace{0pt}3, 4, 4} & {\hspace{0pt}5, 5, 5} & {\hspace{0pt}+1.333} \\
{\hspace{0pt}Haiku} & {\hspace{0pt}4, 4, 4} & {\hspace{0pt}5, 5, 5} & {\hspace{0pt}+1.000} \\
\bottomrule
\end{tabular}
\caption{Original and complete revised-manuscript scores for Case 2.}
\end{table*}

Original:

\begin{quote}Despite rapid progress, two gaps remain. First, no research has scaled EFMs beyond the levels nec- essary to fully exploit this potential—previous efforts have been limited by computational resources and dataset availability. Second, no systematic, quantitative analysis of EFM scaling behavior has been conducted, leaving fundamental questions unanswered about optimal resource allocation and scaling strategies for neural data—in particular, how to allocate a fixed compute budget across model size and number of training epochs.\end{quote}

Revised:

\begin{quote}Despite rapid progress, two gaps remain. First, previous EFM efforts have explored a narrower range of model and data scales than those considered here, in part because of computational resources and dataset availability. Second, systematic quantitative analysis of EFM scaling behavior remains limited, leaving open questions about resource allocation and how to divide a fixed compute budget between model size and training duration.\end{quote}

\subsubsection{Case 3: Both Negative}

Paper \texttt{2025-\allowbreak{}paper\_\allowbreak{}020}; Luna Writer/\allowbreak{}Luna Selector; 19 papers in this direction stratum. The displayed edit is on PDF page 2, text block \texttt{15}, diff \texttt{d0004}, and was accepted by this condition's Selector.

\begin{table*}[tp]\centering
\normalsize
\setlength{\tabcolsep}{4pt}
\renewcommand{\arraystretch}{1.12}
\renewcommand{\thetable}{S36}
\begin{tabular}{@{}>{\raggedright\arraybackslash}p{98.175pt}>{\raggedleft\arraybackslash}p{103.697pt}>{\raggedleft\arraybackslash}p{100.628pt}>{\raggedleft\arraybackslash}p{128.744pt}@{}}
\toprule
{\textbf{\hspace{0pt}Scorer}} & {\textbf{\hspace{0pt}Three original scores}} & {\textbf{\hspace{0pt}Three revised scores}} & {\textbf{\hspace{0pt}Whole-paper mean gain}} \\
\midrule
{\hspace{0pt}Luna} & {\hspace{0pt}3, 4, 4} & {\hspace{0pt}3, 3, 3} & {\hspace{0pt}-0.667} \\
{\hspace{0pt}Haiku} & {\hspace{0pt}4, 4, 4} & {\hspace{0pt}3, 3, 3} & {\hspace{0pt}-1.000} \\
\bottomrule
\end{tabular}
\caption{Original and complete revised-manuscript scores for Case 3.}
\end{table*}

Original:

\begin{quote}We show that each Koopman operator is represented by the exponential of a matrix constructed with Toeplitz matrices and diagonal matrices. This allows us to apply Krylov subspace methods (Gallopou- los \& Saad, 1992; Güttel, 2013; Hashimoto \& Nodera, 2016) to compute the estimation of Koopman operators with low computational costs. By virtue of the universal property of Toeplitz matrices (Ye \& Lim, 2016), we can show the universality of the proposed model with a linear algebraic approach. We also show a generalization bound of the proposed model using a reproducing kernel Hilbert space (RKHS) associated with the Fourier functions. We can analyze both the universality and generalization error with the same framework.\end{quote}

Revised:

\begin{quote}We represent each layer by the exponential of a matrix constructed from Toeplitz and diagonal matrices. This structure allows Krylov subspace methods to compute matrix-exponential actions. Using the universal property of Toeplitz matrices (Ye \& Lim, 2016), we establish a finite-dimensional function-approximation result. We also analyze generalization using an RKHS associated with Fourier functions, with the bound depending on operator norms and requiring appropriate controls on the model class.\end{quote}

\subsubsection{Case 4: At Least One Zero}

Paper \texttt{2026-\allowbreak{}paper\_\allowbreak{}066}; Luna Writer/\allowbreak{}Luna Selector; 107 papers in this direction stratum. The displayed edit is on PDF page 1, text block \texttt{9}, diff \texttt{d0003}, and was accepted by this condition's Selector.

\begin{table*}[tp]\centering
\normalsize
\setlength{\tabcolsep}{4pt}
\renewcommand{\arraystretch}{1.12}
\renewcommand{\thetable}{S37}
\begin{tabular}{@{}>{\raggedright\arraybackslash}p{98.175pt}>{\raggedleft\arraybackslash}p{103.697pt}>{\raggedleft\arraybackslash}p{100.628pt}>{\raggedleft\arraybackslash}p{128.744pt}@{}}
\toprule
{\textbf{\hspace{0pt}Scorer}} & {\textbf{\hspace{0pt}Three original scores}} & {\textbf{\hspace{0pt}Three revised scores}} & {\textbf{\hspace{0pt}Whole-paper mean gain}} \\
\midrule
{\hspace{0pt}Luna} & {\hspace{0pt}1, 1, 1} & {\hspace{0pt}1, 1, 1} & {\hspace{0pt}+0.000} \\
{\hspace{0pt}Haiku} & {\hspace{0pt}2, 2, 3} & {\hspace{0pt}2, 1, 1} & {\hspace{0pt}-1.000} \\
\bottomrule
\end{tabular}
\caption{Original and complete revised-manuscript scores for Case 4.}
\end{table*}

Original:

\begin{quote}Large Language models (LLMs) such as Transformer (Vaswani et al., 2017), BERT (Devlin et al., 2019a), PaLM (Chowdhery et al., 2023), and GPT-4o (Hurst et al., 2024) have demonstrated re- markable capabilities in natural language understanding and generation, which helps to achieve a wide range of tasks such as language translation, sentiment analysis, and question answering. Simi- larly, video Transformers such as TimeSformer (Bertasius et al., 2021), ViViT (Arnab et al., 2021), and Video Swin Transformer (Liu et al., 2022) have illustrated remarkable progress in capturing spatio-temporal dynamics, greatly advancing applications such as video understanding and editing.\end{quote}

Revised:

\begin{quote}Video Transformers, including TimeSformer (Bertasius et al., 2021), ViViT (Arnab et al., 2021), and Video Swin Transformer (Liu et al., 2022), model spatio-temporal dependencies. This manuscript examines the computational cost of exact block attention and a proposed dynamic approach. The review identified fundamental correctness and dimensionality errors in that approach; the results below therefore define the target computation, not a validated efficient algorithm.\end{quote}

\subsubsection{Case 5: Haiku Revision, Luna Positive /\allowbreak{} Haiku Zero}

Paper \texttt{2025-\allowbreak{}paper\_\allowbreak{}093}; Haiku Writer/\allowbreak{}Haiku Selector; 47 papers in this direction stratum. The displayed edit is on PDF page 1, text block \texttt{20}, diff \texttt{d0002}, and was accepted by this condition's Selector.

\begin{table*}[tp]\centering
\normalsize
\setlength{\tabcolsep}{4pt}
\renewcommand{\arraystretch}{1.12}
\renewcommand{\thetable}{S38}
\begin{tabular}{@{}>{\raggedright\arraybackslash}p{98.175pt}>{\raggedleft\arraybackslash}p{103.697pt}>{\raggedleft\arraybackslash}p{100.628pt}>{\raggedleft\arraybackslash}p{128.744pt}@{}}
\toprule
{\textbf{\hspace{0pt}Scorer}} & {\textbf{\hspace{0pt}Three original scores}} & {\textbf{\hspace{0pt}Three revised scores}} & {\textbf{\hspace{0pt}Whole-paper mean gain}} \\
\midrule
{\hspace{0pt}Luna} & {\hspace{0pt}3, 4, 4} & {\hspace{0pt}4, 4, 4} & {\hspace{0pt}+0.333} \\
{\hspace{0pt}Haiku} & {\hspace{0pt}4, 5, 5} & {\hspace{0pt}5, 5, 4} & {\hspace{0pt}+0.000} \\
\bottomrule
\end{tabular}
\caption{Original and complete revised-manuscript scores for Case 5.}
\end{table*}

Original:

\begin{quote}To address this gap, in this paper, we curate a comprehensive ex- trapolation benchmark consisting of seven datasets sourced from Matminer (Ward et al., 2018), accompanied by well-defined train/\allowbreak{}validation/\allowbreak{}test splits based on real-\end{quote}

Revised:

\begin{quote}To address this gap, in this paper, we curate a comprehensive extrapolation benchmark consisting of seven tasks derived from four datasets sourced from Matminer (Ward et al., 2018), accompanied by well-defined train/\allowbreak{}validation/\allowbreak{}test splits based on real-\end{quote}

\section{Reproduction Resources}

The code package includes all data, code, configurations, and external files needed to reproduce the analyses and figures.

\section{Ethical Considerations and Responsible Use}

\textbf{Data Sources and Licensing}. This study uses publicly available ICLR 2025 and 2026 submissions and associated review metadata obtained through OpenReview. We selected 100 papers from each year for academic analysis of AI-based peer review. The original manuscripts and their contents remain subject to the copyrights and applicable licenses of their respective authors. Our reported findings primarily consist of independently computed statistics and analyses. Any redistribution of source materials or derived artifacts is subject to the applicable licenses and terms of use.

\textbf{Use of AI Models.} Large language models are the primary research objects in this study. They are used to evaluate manuscripts, generate and select candidate revisions, and assess scoring consistency. These model-generated reviews and revisions constitute experimental outputs rather than independent human assessments. The experimental procedures, statistical analyses, and interpretation of the results remain the responsibility of the authors.

\textbf{Ethical Risks.} The study analyzes publicly available research manuscripts and review information without recruiting human participants. Manuscript inputs are anonymized for AI evaluation. Our findings should not be interpreted as evidence that AI-generated reviews can replace human peer reviewers or reliably determine publication decisions. Automated manuscript revisions may also introduce inaccuracies, and their scientific validity requires independent verification.

%% file: references.bib
@inproceedings{liu-etal-2023-g,
    title = "{G}-Eval: {NLG} Evaluation using {G}pt-4 with Better Human Alignment",
    author = "Liu, Yang  and
      Iter, Dan  and
      Xu, Yichong  and
      Wang, Shuohang  and
      Xu, Ruochen  and
      Zhu, Chenguang",
    editor = "Bouamor, Houda  and
      Pino, Juan  and
      Bali, Kalika",
    booktitle = "Proceedings of the 2023 Conference on Empirical Methods in Natural Language Processing",
    month = dec,
    year = "2023",
    address = "Singapore",
    publisher = "Association for Computational Linguistics",
    url = "https://aclanthology.org/2023.emnlp-main.153/",
    doi = "10.18653/v1/2023.emnlp-main.153",
    pages = "2511--2522"
}

@inproceedings{chiang-lee-2023-closer,
    title = "A Closer Look into Using Large Language Models for Automatic Evaluation",
    author = "Chiang, Cheng-Han  and
      Lee, Hung-yi",
    editor = "Bouamor, Houda  and
      Pino, Juan  and
      Bali, Kalika",
    booktitle = "Findings of the {A}ssociation for {C}omputational {L}inguistics: {EMNLP} 2023",
    month = dec,
    year = "2023",
    address = "Singapore",
    publisher = "Association for Computational Linguistics",
    url = "https://aclanthology.org/2023.findings-emnlp.599/",
    doi = "10.18653/v1/2023.findings-emnlp.599",
    pages = "8928--8942"
}

@inproceedings{du-2025-titletrap,
    title = "{T}itle{T}rap: Probing Presentation Bias in {LLM}-Based Scientific Reviewing",
    author = "Du, Shurui",
    editor = "Akter, Mousumi  and
      Chowdhury, Tahiya  and
      Eger, Steffen  and
      Leiter, Christoph  and
      Opitz, Juri  and
      {\c{C}}ano, Erion",
    booktitle = "Proceedings of the 5th Workshop on Evaluation and Comparison of {NLP} Systems",
    month = dec,
    year = "2025",
    address = "Mumbai, India",
    publisher = "Association for Computational Linguistics",
    url = "https://aclanthology.org/2025.eval4nlp-1.10/",
    doi = "10.18653/v1/2025.eval4nlp-1.10",
    pages = "119--125",
    ISBN = "979-8-89176-305-0"
}

@inproceedings{garg-etal-2025-revieweval,
    title = "{R}eview{E}val: An Evaluation Framework for {AI}-Generated Reviews",
    author = "Garg, Madhav Krishan  and
      Prasad, Tejash  and
      Singhal, Tanmay  and
      Kirtani, Chhavi  and
      Mandal, Murari  and
      Kumar, Dhruv",
    editor = "Christodoulopoulos, Christos  and
      Chakraborty, Tanmoy  and
      Rose, Carolyn  and
      Peng, Violet",
    booktitle = "Findings of the {A}ssociation for {C}omputational {L}inguistics: {EMNLP} 2025",
    month = nov,
    year = "2025",
    address = "Suzhou, China",
    publisher = "Association for Computational Linguistics",
    url = "https://aclanthology.org/2025.findings-emnlp.1120/",
    doi = "10.18653/v1/2025.findings-emnlp.1120",
    pages = "20542--20564",
    ISBN = "979-8-89176-335-7"
}

@inproceedings{homan-etal-2026-many,
    title = "How Many Ratings per Item are Necessary for Reliable Significance Testing?",
    author = "Homan, Christopher M.  and
      Korn, Flip  and
      Pandita, Deepak  and
      Welty, Chris",
    editor = "Demberg, Vera  and
      Inui, Kentaro  and
      Marquez, Llu{\'i}s",
    booktitle = "Findings of the {A}ssociation for {C}omputational {L}inguistics: {EACL} 2026",
    month = mar,
    year = "2026",
    address = "Rabat, Morocco",
    publisher = "Association for Computational Linguistics",
    url = "https://aclanthology.org/2026.findings-eacl.223/",
    doi = "10.18653/v1/2026.findings-eacl.223",
    pages = "4258--4273",
    ISBN = "979-8-89176-386-9"
}

@misc{stureborg2024inconsistent,
  title = {Large Language Models are Inconsistent and Biased Evaluators},
  author = {Stureborg, Rickard and Alikaniotis, Dimitris and Suhara, Yoshi},
  year = {2024},
  eprint = {2405.01724},
  archivePrefix = {arXiv},
  primaryClass = {cs.CL},
  doi = {10.48550/arXiv.2405.01724},
  url = {https://arxiv.org/abs/2405.01724}
}

@misc{li2026rhetoric,
  title = {How Can Rhetoric Reward-Hack {AI} Reviewers? Dissecting Rhetorical Sensitivity in {AI}-Based Peer Review},
  author = {Li, Ming and Wang, Chenguang and Li, Xirui and Zeng, Xinyue and Li, Dianqi and Shi, Peng and Zhou, Dawei and Zhou, Tianyi},
  year = {2026},
  eprint = {2608.08975},
  archivePrefix = {arXiv},
  primaryClass = {cs.CL},
  doi = {10.48550/arXiv.2608.08975},
  url = {https://arxiv.org/abs/2608.08975}
}

@misc{li2026llmreviewer,
  title = {{LLM}-as-a-Reviewer: Benchmarking Their Ability, Divergence, and Prompt Injection Resistance as Paper Reviewers},
  author = {Li, Lingyao and Xiong, Junjie and Zhu, Changjia and Yu, Runlong and Chen, Chen and Wang, Junyu and Ma, Renkai and Lu, Zhicong},
  year = {2026},
  eprint = {2605.25415},
  archivePrefix = {arXiv},
  primaryClass = {cs.CL},
  doi = {10.48550/arXiv.2605.25415},
  url = {https://arxiv.org/abs/2605.25415}
}

@article{hoeffding1963probability,
  title = {Probability Inequalities for Sums of Bounded Random Variables},
  author = {Hoeffding, Wassily},
  journal = {Journal of the American Statistical Association},
  year = {1963},
  volume = {58},
  number = {301},
  pages = {13--30},
  doi = {10.1080/01621459.1963.10500830},
  url = {https://doi.org/10.1080/01621459.1963.10500830}
}

@misc{zou2025leveraging,
  author = {Zou, James and Thakkar, Nitya},
  title = {Leveraging {LLM} Feedback to Enhance Review Quality},
  year = {2025},
  month = apr,
  howpublished = {ICLR Blog},
  url = {https://blog.iclr.cc/2025/04/15/leveraging-llm-feedback-to-enhance-review-quality/}
}

@misc{aaai2025launches,
  author = {{AAAI}},
  title = {{AAAI} Launches {AI}-Powered Peer Review Assessment System},
  year = {2025},
  howpublished = {Association for the Advancement of Artificial Intelligence},
  url = {https://aaai.org/aaai-launches-ai-powered-peer-review-assessment-system/}
}
